\documentclass[pdflatex,sn-basic]{sn-jnl}

\usepackage{bm} 

\jyear{2021}%

\theoremstyle{thmstyleone}%
\theoremstyle{thmstyletwo}%

\theoremstyle{thmstylethree}%

\begin{document}

\title[Self-supervised Extraction of Human Motion Structure via ...]{Self-supervised Extraction of Human Motion Structures via Frame-wise Discrete Features}


\author[1,2]{\fnm{Tetsuya} \sur{Abe}}\email{s211948v@st.go.tuat.ac.jp}

\author*[1,2]{\fnm{Ryusuke} \sur{Sagawa}}\email{ryusuke.sagawa@aist.go.jp}

\author[1,2]{\fnm{Ko} \sur{Ayusawa}}\email{k.ayusawa@aist.go.jp}

\author[3]{\fnm{Wataru} \sur{Takano}}\email{takano@sigmath.es.osaka-u.ac.jp}

\affil*[1]{\orgname{Tokyo University of Agriculture and Technology}, \orgaddress{\city{Koganei}, \state{Tokyo}, \country{Japan}}}

\affil*[2]{\orgname{National Institute of Advanced Industrial Science and Technology}, \orgaddress{\city{Tsukuba}, \state{Ibaraki}, \country{Japan}}}

\affil[3]{\orgname{Osaka University}, \orgaddress{\city{Toyonaka}, \state{Osaka}, \country{Japan}}}

\abstract{The present paper proposes an encoder-decoder model for extracting the
structures of human motions represented by frame-wise discrete features 
in a self-supervised manner. In the proposed method, features are extracted as codes 
in a motion codebook without the use of human knowledge, and the 
relationship between these codes can be visualized on a graph. Since the codes 
are expected to be temporally sparse compared to the captured frame rate and can 
be shared by multiple sequences, the proposed network model also addresses the 
need for training constraints. Specifically, the model consists of 
self-attention layers and a vector clustering block. The attention layers 
contribute to finding sparse keyframes and discrete features as motion 
codes, which are then extracted by vector clustering. The constraints are 
realized as training losses so that the same motion codes can be as contiguous as 
possible and can be shared by multiple sequences. In addition, we propose the use 
of causal self-attention as a method by which to calculate attention 
for long sequences consisting of numerous frames. In our experiments, the 
sparse structures of motion codes were used to compile a graph that facilitates 
visualization of the relationship between the codes and the differences between 
sequences. We then evaluated the effectiveness of the extracted motion codes by 
applying them to multiple recognition tasks and found that performance levels 
comparable to task-optimized methods could be achieved by linear probing. }




\maketitle

\section{Introduction}\label{sec:Introduction}

Improved recognition of human behaviors will provide important progress toward
realizing advanced technologies, such as human-robot interactions. However, one
of the difficulties with action recognition can be traced to the continuity of
human motions because even just a few seconds of motion will contain several
smoothly connected actions. In order to minimize the effects of this complexity, most
existing human action recognition methods use motion data with annotations added
for specific motion segments. In efforts aimed at learning segmented motion data, recent research has successfully extracted motion features that are useful for action recognition \cite{su2020predict,lin2020ms2l,thoker2021skeleton}. These methods, which assume that the given motions have the same level of granularity, seek to identify recurring or similar motion sequence patterns in latent space. The extracted features are usually convoluted with the whole-motion sequence. However, the effectiveness of such feature representations strongly depends on the granularity of the motion annotations that are applied during the training
phase. For example, when a sequence of walking data is given, the entire
sequence is converted to a feature that can represent ``walking", but its meaning
cannot be subdivided in latent space into ``one step with the left foot" or ``one
step with the right foot". Since this fixed-granularity problem is found in the
human motion features of all of these methods, their applicability to multiple
applications is limited to generalizations that require different levels of
representation granularity. 

Human motions consist of multiple stages, each of which is influenced by the
characteristics of the individual human or the contexts of specific motions.
Therefore, it is necessary to extract representations of multiple spatial and
temporal components in order to effectively recognize these motions. Taken
further, if it were possible to extract the unique representations of a specific
individual, this would be very useful for understanding the characteristics of his
or her human motions. For example, this could help to explain the differences in the
motions between beginners and experts. However, since each individual possesses
unique representations that are not shared among others, the actions of one person are not expected to completely match the shared knowledge of others
in a manner similar to natural language. Therefore, first among the three
primary issues to be resolved is finding a way to extract a unique
representation without using preexisting knowledge.

Typically, human motions are expressed as multi-dimensional continuous
quantities, such as joint angles at each sampling moment. However, when we
consider the case of generating a new motion, specifying all
joint angles at each moment is difficult. 
Therefore, in order to make it
feasible to specify such values, representations which are a finite
number of components or parameters are required. This is the second representation issue that
must be addressed.

Since the components that make up a motion have temporal relationships with each
other, the recognition of the motion is equivalent to identifying those
relationships. In addition, since any human motion will have a relationship
with motions that took place several seconds or more previously, the temporal
receptive field for the recognition should be wider than the dependency.
Furthermore, since it is necessary for the recognition of human motions, the
receptive field should be wider than several hundred frames. Therefore, the
third issue is determining a recognition method that can be used with a wide
receptive field.

With the above motivations in mind, our goal is to extract identifiable features
that can be used to represent human motions without using preexisting knowledge (annotated motion
data).
The contributions of the present study are as follows. We show how an intermediate
representation of human motion can be generated in latent space based on an
encoder-decoder model without using preexisting knowledge (annotated motion
data). The representation consists of a finite number of components that are
obtained by discretizing the latent space, and the proposed method realizes a
wide temporal receptive field by using an attention-based network to extract
relationships in a long sequence.

\section{Related work}\label{sec:Related work}

The process of understanding human behavior is typically categorized into
several tasks, such as action recognition and action segmentation. Action
recognition is defined as the task of finding areas of correspondence between
input data and action labels. In this task, the system basically identifies
answers with actions that match the input data. Hence, in supervised approaches,
action labels are assigned, and the target actions to be detected are fixed. As
examples, previously methods proposed \cite{girdhar2017actionvlad,wang2021oadtr,sigurdsson2017asynchronous} determine
actions based on supervised learning directly from video footage used as input
data. 

Another approach by which to determine actions is to use skeletal information acquired by
detecting human poses or by using a motion capture
system~\cite{shi2021adasgn,duan2021revisiting,yan2018spatial}. Some methods
based on skeletal
information~\cite{takano2008integrating,takano2015statistical,plappert2018learning,ahn2018text2action,yamada2018paired,ahuja2019language2pose}
define the problem as a translation between motion and language.
However, since the cost of obtaining input data, such as videos with action
labels, is high, unsupervised approaches, such as pre-training, have been used
to learn representations from video footage~\cite{misra2016shuffle}. In these
methods, latent spaces are learned from skeletal information based on
self-supervised approaches~\cite{su2021self,su2020predict}. 

Action segmentation is the task of segmenting a temporal sequence of input data
into multiple actions. A number of methods
\cite{farha2019ms,huang2020improving,chen2020action} determine multiple action
types and their boundaries in examined video frames based on supervised
learning. Unsupervised approaches to action segmentation that work by using
clustering to find the same actions~\cite{li2021action,del2015articulated,sarfraz2021temporally} and aligning
phases~\cite{dwibedi2019temporal} have also been proposed. 

Separately, since numerous actions are said to have hierarchical structures,
which means multi-level action labels can be defined, other previous methods
based on supervised
learning~\cite{lan2015action,pirsiavash2014parsing,lea2017temporal,lea2016segmental,richard2017weakly}
have proposed the use of fine-grained action labels to explain coarse-level
actions. For example, a method proposed in a previous
study~\cite{sener2018unsupervised} learns the latent space of sub-actions by
means of an unsupervised approach based on clustering. Using skeletal information to learn the representations of human motion based on
unsupervised learning is a method that can also be used for other tasks. For
example, when examining video footage, the joint angles of the next frame have
been predicted using previous frame data
~\cite{jain2016structural,guo2020action2motion,ghosh2017learning,butepage2017deep}.

One of the major approaches used to explain human actions is to find
relationships between the motions and the words of natural languages. Since
words are common knowledge, the parts of an action can be readily understood by
users. Hence, various methods of facilitating mutual translations between
motions and language have been studied. These approaches can achieve a useful
intermediate representation of motion by finding the relationship between a
motion and a language. In one example, a motion language model encoded by a
hidden Markov model (HMM) was used as an intermediate representation to express
the relationships between motions and language~\cite{takano2015statistical}. 

Separately, a context vector encoded by a recurrent neural network (RNN) was
used to provide an intermediate representation of
motion~\cite{plappert2018learning}, which allowed human motions to be generated
from the input text by learning the intermediate representation as the output of
an RNN based on a generative adversarial network (GAN) in
\cite{ahn2018text2action}, while an autoencoder for motion and language that
calculates two latent vectors has also been proposed~\cite{yamada2018paired}. In
the latter method, the difference between the latent vectors is minimized in
training and then used as the intermediate representation. Joint embedding of
the latent vectors has been used for an autoencoder \cite{ahuja2019language2pose}
, which reproduces the motion from the joint vector, and the latent vector
of the language encoder is only used in inference time. However, in all of these
approaches, the intermediate representations of the motion are supervised by the
annotations using language labels, which means representing
unique characteristics that cannot be expressed by language is difficult.

Furthermore, while these supervised or self-supervised methods with annotated
(segmented) motion data can extract the motion features for action recognition,
the latent space obtained in this manner is mediated by human knowledge that
segments the motion data, which means that these methods require motion data with an
appropriate level of granularity. In order to avoid this problem, we propose a method
that can extract feature representations without preexisting human knowledge
(segmented motion data with annotation label). More specifically, the proposed method is based on a variational autoencoder
that generates a discrete latent space, such as a vector quantized-variational
autoencoder (VQ-VAE) ~\cite{DBLP:journals/corr/abs-1711-00937} or a dynamical
variational autoencoder  (dVAE)~\cite{DBLP:journals/corr/abs-2102-12092}.
Temporal relationships between frames are found based on self-attention, as
proposed in the Transformer~\cite{vaswani2017attention}. The great benefit of
discrete representation is its ability to find the discontinuous points of
motion data and thus help in detecting actions in non-segmented motion data.

\section{Proposed method}\label{sec:Proposed method}

The proposed method extracts the discrete motion feature of each frame in a long
sequence. Once extracted, the motion feature is regarded as a motion code in a
motion codebook, which is a set of components used to explain various motions.
The purpose of such frame-wise feature extraction is to obtain motion codes
independently of existing knowledge. However, since the proposed method does not
use segmented motion data, the limited receptive field of a convolutional
network is unsuitable. Accordingly, the proposed method uses a self-attention
architecture, and a vector-quantized
framework~\cite{DBLP:journals/corr/abs-1711-00937} to find discrete
representations among motion data by considering temporal relationships over a
wide range of frames. The temporal dependency of human motions is unknown but is considered to be more
than several seconds. For example, the receptive field of the network used along
the time axis should be several hundred frames wide if the motion is captured at
100 Hz. 

\subsection{Encoder-decoder with discrete latent space}\label{sec:Encoder-decoder with discrete latent space}

\begin{figure}[h]
\centering
\includegraphics[width=1\textwidth]{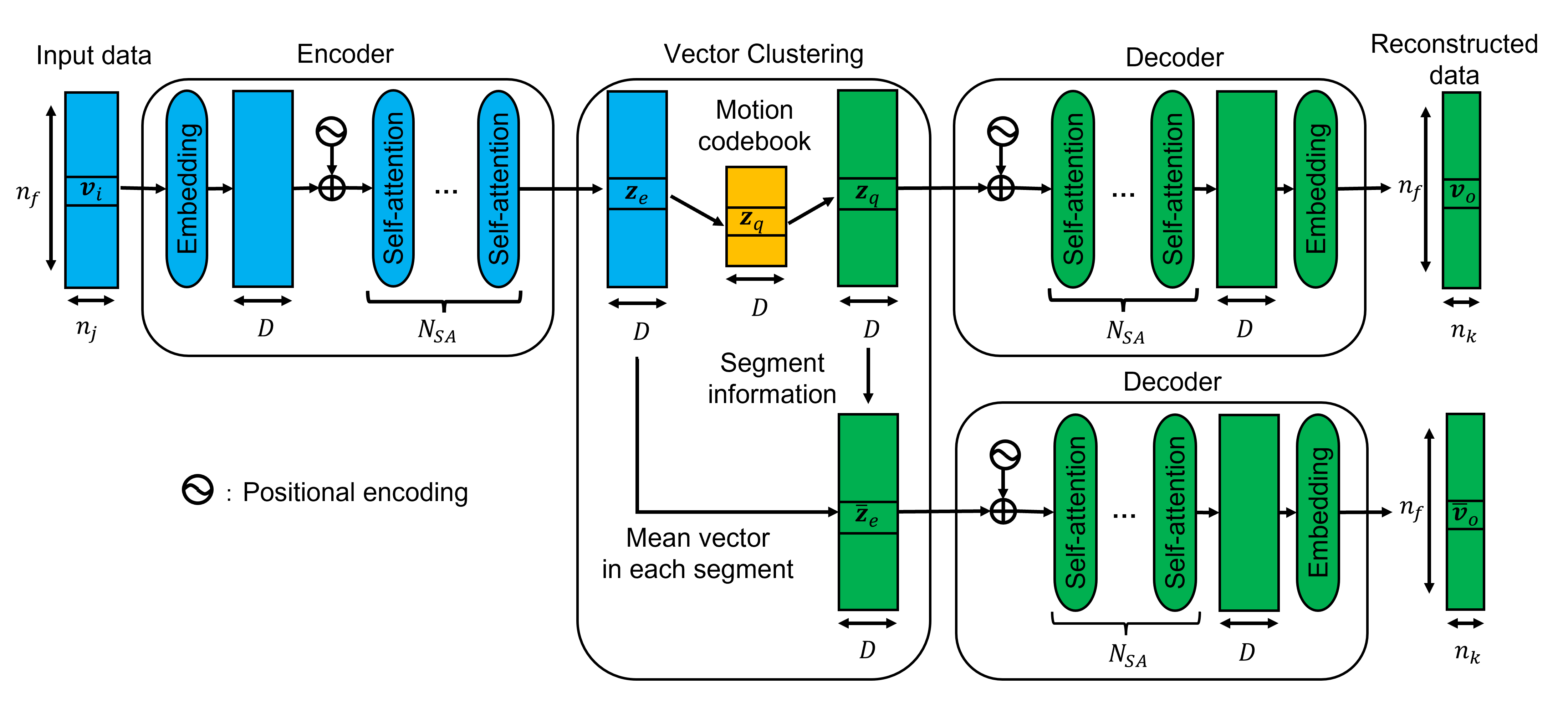}
\caption{Proposed model consistings of an encoder, a decoder, and vector clustering, which has two outputs: $\bm{z}_q$ and $\bar{\bm{z}}_e$. The former is a motion code replaced from the encoder output, and the latter is the mean vector of a segment of the same motion code}\label{fig:Network architecture}
\end{figure}

The proposed method generates a discrete latent space that describes the
structure of a human motion using a network consisting of an encoder and a
decoder combined with a block of clustering latent vectors that are used to
extract discrete motion codes. Vector clustering is realized as a quantization
process that maps the encoder output to the nearest embedding vector in a motion
codebook. In the proposed study, this quantization is implemented based on
VQ-VAE~\cite{DBLP:journals/corr/abs-1711-00937}, and the encoder and decoder are
realized by self-attention layers~\cite{vaswani2017attention} in order to find
relationships between frames.

The architecture of the proposed model is shown in Fig. \ref{fig:Network
architecture}. The input data of a motion consist of $n_f$ frames of
$n_j$-dimensional vectors at each frame. and the encoder converts the input
vector $\bm{v}_i$ at each frame to a feature vector $\bm{z}_e$. The vector
clustering has two outputs: one output replaces $\bm{z}_e$ with the vector
$\bm{z}_q$, which is the nearest neighbor in the codebook based on Euclidean
distance in the latent space. Note that the codebook itself consists of 512
kinds of embedding vectors $\bm{z}_q$ in latent space. The other output replaces
an encoded vector with the mean vector $\bar{\bm{z}}_e$ of each segment, which
consists of consecutive frames of the same motion code $\bm{z}_q$. The
difference between $\bm{z}_q$ and $\bar{\bm{z}}_e$ is considered to be the
parameter of a segment that represents variations in the same cluster. Since
$\bm{z}_q$ does not contain this parameter, we refer to it as the default
parameter. The decoder reconstructs the outputs $\bm{v}_o$ and $\bar{\bm{v}}_o$
using $\bm{z}_q$ and $\bar{\bm{z}}_e$, respectively.

The outputs are framewise human motions that correspond to the input. If the
input is motion, then the model is designed as an autoencoder. However, the input can
also be other modalities, such as first-person video input by the subject. In
the present paper, a high-dimensional video input is assumed to have already been
encoded frame by frame for use as a feature vector and that the temporal
relationship has already been extracted by the proposed model.

\subsection{Layered causal self-attention for a long sequence}\label{sec:Layered causal self-attention for a long sequence}

The input sequence length can be more than a thousand frames if, for example, it
is captured for more than one minute at 30 frames per second (fps). Because
identifying every combination of frames by self-attention due to
time and memory limitations is not feasible, the attention matrix is only calculated for a
portion of a sequence even if the multiple self-attention layers
proposed in \cite{vaswani2017attention} are used.
Therefore, for example, the attention is calculated for $M (< N)$ frames as the attention width, even though the sequence has a total of $N$ frames.
In this section, the use of
layered causal self-attention is proposed to overcome this limitation.

In the human motion extraction task, it can be assumed that the output motion is
causal only for the past input and output. However, since simply masking future
frames will not decrease computational costs, the proposed approach only
calculates the attention between each frame and its preceding $M-1$ frames. The
output $\bm{z}$ of self-attention is calculated using the following equation: 

\begin{align}
   \bm{z} = \mbox{softmax}(\frac{\bm{Q}\bm{K}^T}{\sqrt{D}}\bm{V}) \label{eq1}
\end{align}
where $\bm{Q}, \bm{K}$, and $\bm{V}$ are the query, key, and value vectors, and
$D$ is the dimension of the feature vector.

\begin{figure}[h]
\centering
\includegraphics[width=1\textwidth]{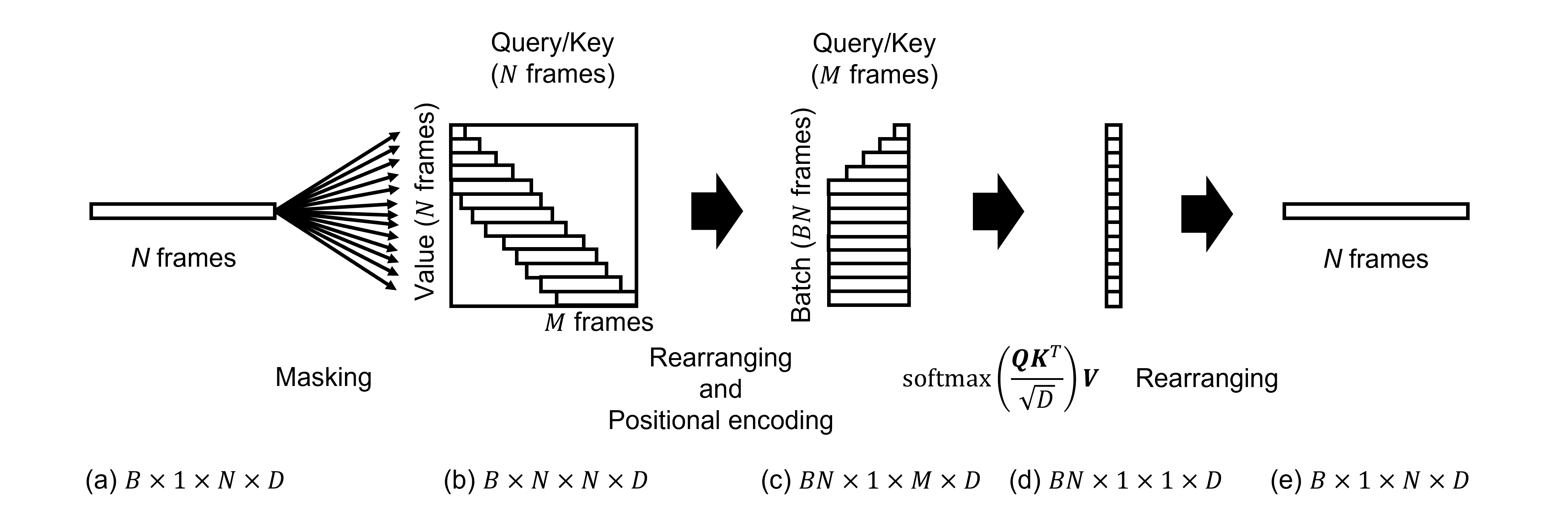}
\caption{Attention is calculated only for preceding frames. By
   rearranging the vectors, a wide receptive field is realized with a narrow
   attention width for each layer. The sizes indicate the number of vectors or
   vector dimensions: $\mbox{Batch} \times \mbox{Value} \times \mbox{Query/Key}
   \times \mbox{Feature}$ }\label{fig:causal_attention}
\end{figure}

Fig. \ref{fig:causal_attention}(a) is an input sequence and Fig. \ref{fig:causal_attention}(b) is the attention mask
for each frame. The dimensions are the batch size $B$, the number of value
vectors, and the number of query/key vectors. The vectors are rearranged so that
the attention is calculated between a single value vector and vectors of the
preceding $M-1$ frames, as shown in Fig. \ref{fig:causal_attention}(c). The value vector is the feature of the
last frame of each row in Fig. \ref{fig:causal_attention}(c), and output $\bm{z}$ is obtained with attention 
width $M$ for every frame except the beginning of the sequence. If the
number of attention layers is $N_{SA}$, then the total receptive field size is
$N_{SA}M$ frames for every frame. For preceding frames that are less than $M$,
the query and key vectors are padded by zero vectors. Since the attention is
calculated for the last frame, the positional encoding is given as a position
relative to the value vector frame. Therefore, the positional encoding values
are the same in each column in Fig. \ref{fig:causal_attention}(c). In the present study, the method used to calculate
the position is the same as that used in the original Transformer. The output is
rearranged again, as shown in Fig. \ref{fig:causal_attention}(e), and then used as input for the feed-forward
network in the same manner as proposed in the Transformer.

\subsection{Loss for extracting motion codes shared by sequences}\label{sec:Loss for extracting motion codes shared by sequences}

If the network is optimized to reconstruct the output $\bm{v}_o$ without using
any constraint to share motion codes, then each sequence will be reconstructed using
unique codes. However, if the latent space forms a structure of human motions, a
latent vector extracted by the proposed method is regarded as a motion code that
is expected to be shared by multiple sequences. Hence, the following loss $L$ is
proposed to encourage motion code sharing:

\begin{align}
   L = \sum^{n_f}_{k} \alpha L_{\mbox{reconst}} + L_{\mbox{latent}} \label{eq2}
\end{align}
where $L_{\mbox{reconst}}$ and $L_{\mbox{latent}}$ are the losses of
reconstruction and latent space, respectively, for the $k$-th frame of a motion
sequence of $n_f$ frames, and $\alpha$ is a user-defined weight of the
reconstruction loss, as are $\beta$ and $\gamma$ in equations presented later.

\subsubsection{Reconstruction loss}\label{sec:Reconstruction loss}

The reconstruction loss is calculated with the parameter of each
segment and the default parameter. In the proposed method, it is assumed that
the default parameter of a motion code indicates temporally local motion. The
loss with the constraint of local motion is defined as follows:

\begin{align}
   L_{\mbox{reconst}} & = L_{\mbox{p}} + L_{\mbox{v}} \label{eq3}
\end{align}

\begin{align}
   L_{\mbox{p}} = \parallel \bm{v}_{ik} - \bar{\bm{v}}_{ok} \parallel^2 \label{eq4}
\end{align}

\begin{align}
   L_{\mbox{v}} = \parallel \dot{\bm{v}}_{ik} - \dot{\bm{v}}_{ok} \parallel^2 \label{eq5}
\end{align}
where $L_{\mbox{p}}$ is the reconstruction loss with the parameter of each
segment, and $L_{\mbox{v}}$ is that of a local motion. In addition,
$L_{\mbox{p}}$ is calculated as the difference of the output vectors, and
$L_{\mbox{v}}$ is calculated as the difference of the temporal derivative of
these output vectors. If the output vectors refer to the position, the default
parameter constraint is imposed on the velocity of the vectors.

\subsubsection{Latent space loss}\label{sec:Latent space loss}

The second part of loss $L$ is the latent space loss, which is defined as follows:

\begin{align}
   L_{\mbox{latent}} = L_{\mbox{vc}} + L_{\mbox{tv}} \label{eq6}
\end{align}

\begin{align}
   L_{\mbox{vc}} = \parallel \mbox{sg}[\bm{z}_{qk}] - \bm{z}_{ek} \parallel^2
    + \beta \parallel \bm{z}_{qk} - \mbox{sg}[\bm{z}_{ek}] \parallel^2 \label{eq7}
\end{align}

\begin{align}
   L_{\mbox{tv}} = \gamma \parallel \dot{\bm{z}}_{qk} \parallel_1 \label{eq8}
\end{align}
where $L_{\mbox{vc}}$ is the loss of vector clustering, and $L_{\mbox{tv}}$ is that
of the temporal change of quantized latent vectors.
Here, $L_{\mbox{vc}}$ is based on
the vector quantization~\cite{DBLP:journals/corr/abs-1711-00937} used to make
the encoded vector $\bm{z}_{ek}$ easy to quantize, and $\bm{z}_{qk}$ is an entry
in the motion codebook. The function sg is the stop-gradient operator that is
defined as an identity at the forward computation time and has zero derivatives.
Moreover, $L_{\mbox{tv}}$ is the constraint that ensures the same motion code continues for as
long as possible and is realized by minimizing the total variation defined by
$L^1$ norm.

\subsubsection{Restricting motion codes}\label{sec:Restricting motion codes}

Let $S$ be a subset of input sequences that includes all types of motion in the dataset, and let $Z_q$
be the subset of motion codes used to encode the sequences in $S$. In this case,
all sequences in the input dataset must be reconstructed using the subset
$Z_q$. Since different subsets $Z_{qj} (j=1,\ldots,J)$ can be taken from the input
dataset with $J$ sequences, the loss is minimized by restricting the codebook to a subset $Z_{qj}$
that is randomly chosen. In the first epoch, all codes are used for encoding,
and one of the subsets is used to encode each sequence after the second epoch.
Restricting the set of motion codes ensures that the codes are shared. 

\subsection{Visualizing the structure of motion codes}\label{sec:Visualizing the structure of motion codes}

The decoder generates human motion from motion codes. Therefore, the attention
weight for a frame indicates the frames referenced during generation. By
considering multiple layers of attention and residual connections, the weight
matrix $\bm{W}$ is calculated from the attention weight $\bm{W}_i (i=1,\ldots,N_{SA})$ of the $i$-th decoder attention layer as follows:

\begin{align}
    \bm{W} = (\bm{I} + \bm{W}_{N_{SA}})\cdots(\bm{I} + \bm{W}_2)(\bm{I} + \bm{W}_1) \label{eq9}
\end{align}
where $\bm{I}$ is the identity matrix. Since the length $N$ of input sequences is
longer than the number of frames $M$ for the attention layers, $\bm{W}_i$ is
calculated by concatenating the attention weights of partial sequences.

\begin{figure}[h]
   \centering
   \begin{tabular}{cc}
   \includegraphics[width=0.45\textwidth]{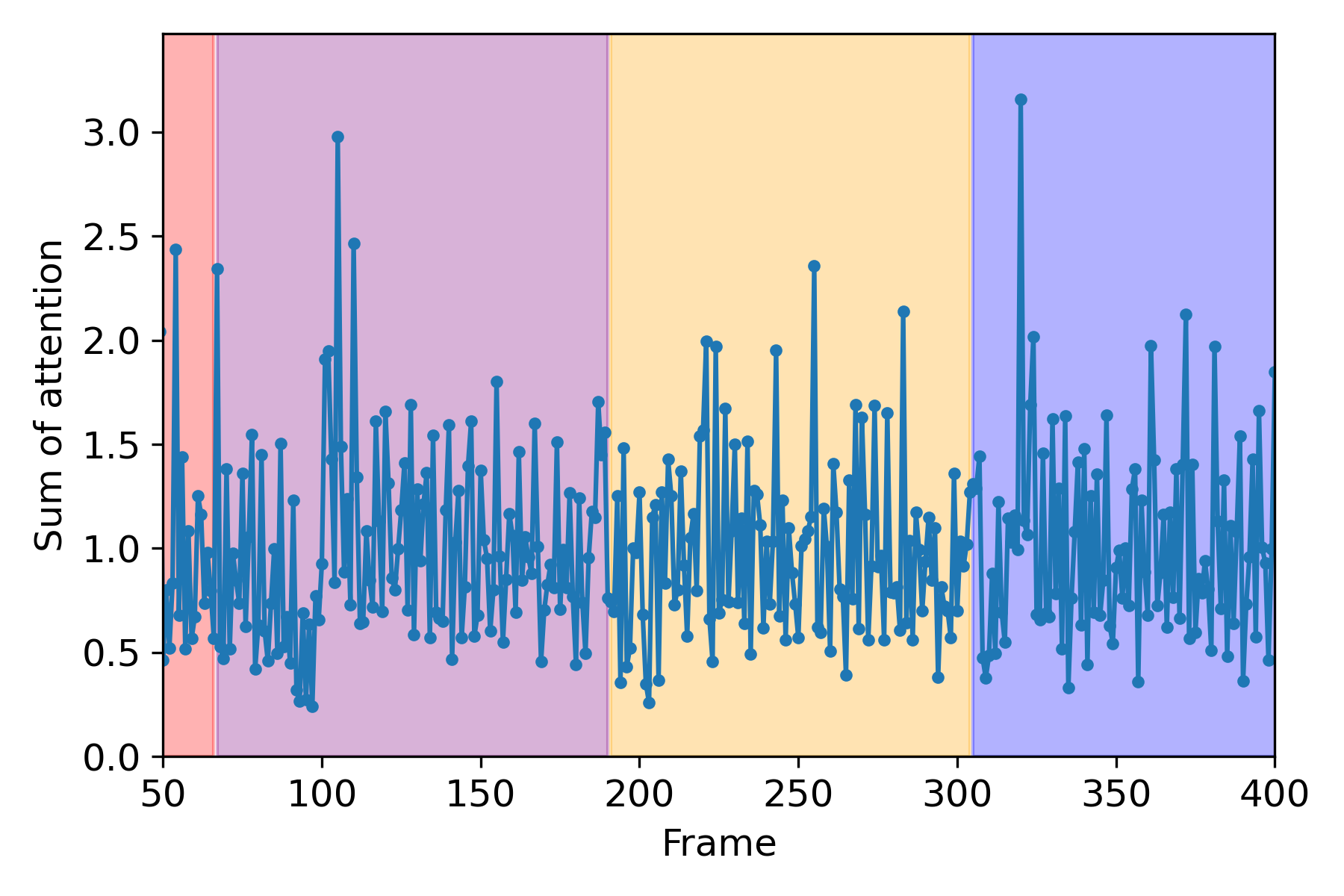} &
   \includegraphics[width=0.45\textwidth]{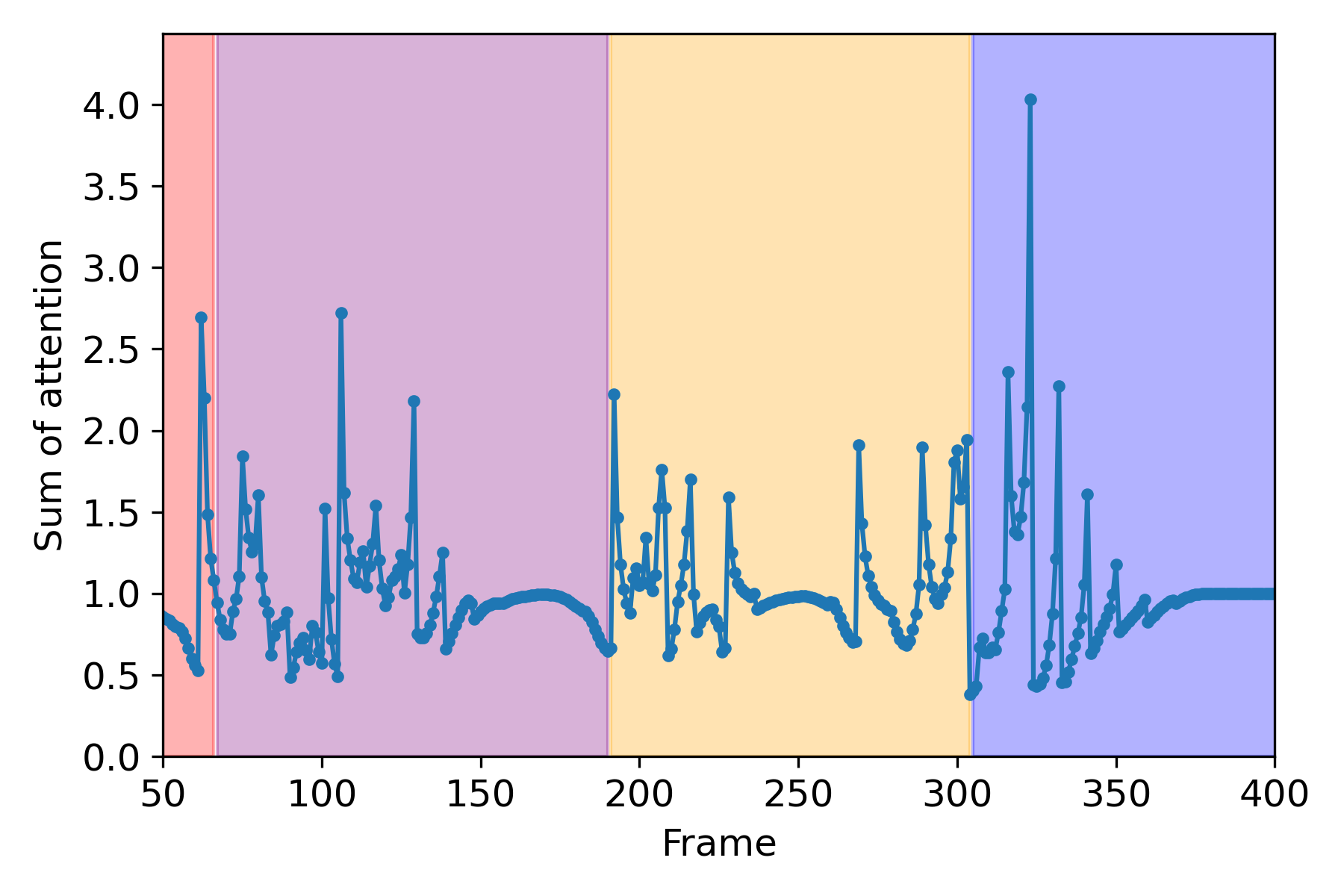} 
   \\ \small{(a)} &  \small{(b)}
   \end{tabular}
   \caption{Sum of attention weights for each query/keyframe: (a) basic variational
autoencoder (VAE) model, (b) proposed model}
    \label{fig:sum_attention}
\end{figure}

High-weight frames, which are important for reconstructing motion, can be
considered as keyframes. We have observed that the quantized latent space defined
using motion codes generates high attention weights more sparsely, as compared to the
continuous latent space defined by the basic use of a variational autoencoder (VAE)~\cite{kingma2013auto}.
Fig. \ref{fig:sum_attention} shows an example of the sum of attention weights for each
query/key frame. These two graphs show the same part of a sequence decoded by a basic VAE model (Fig. \ref{fig:sum_attention}(a)) and by the proposed model (Fig. \ref{fig:sum_attention}(b)). This example is one of the
sequences in the JHU-ISI gesture and skill assessment working set (JIGSAWS)~\cite{gao2014jhu}, and the background color
indicates the action labels given in the dataset. Although the distribution of
the weights is dense in Fig. \ref{fig:sum_attention}(a), the proposed model generates
a sparse distribution in Fig. \ref{fig:sum_attention}(b). Since high weights can be observed for the frames
close to the label boundaries, the weights appear to have some semantic meaning.

\begin{figure}[h]
   \centering
   \begin{tabular}{cc}
      \includegraphics[width=0.35\textwidth]{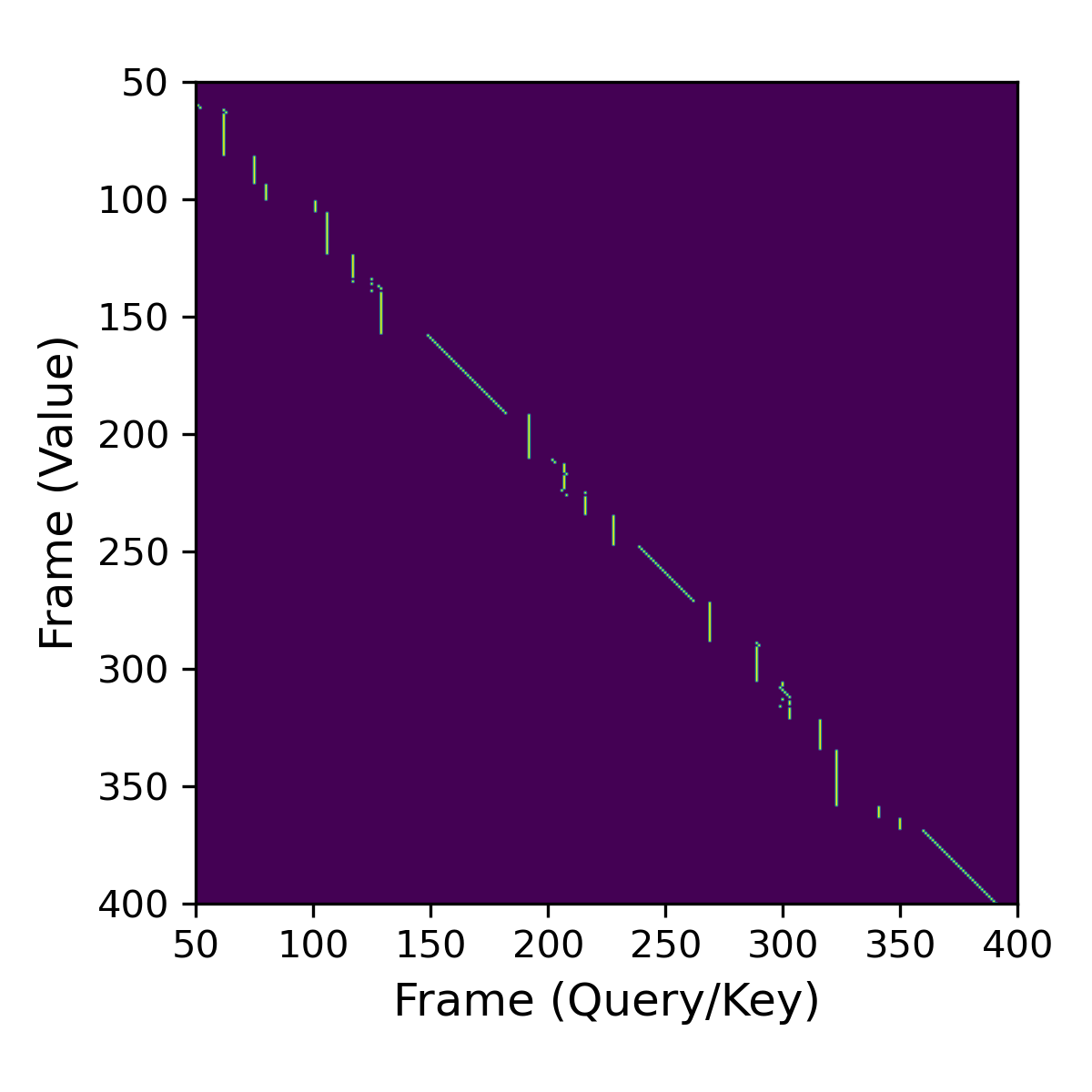} &
      \includegraphics[width=0.5\textwidth]{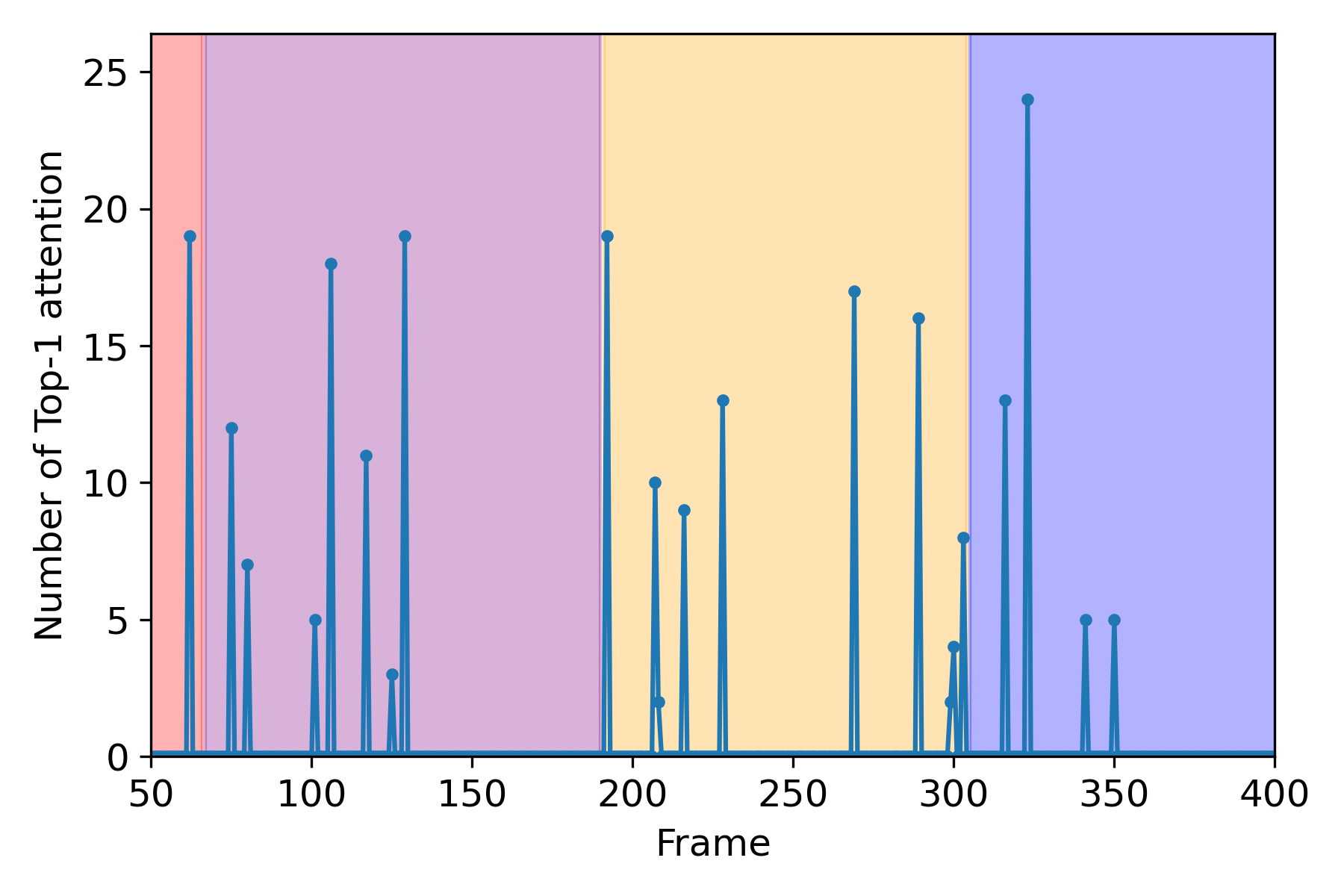}
      \\ \small{(a)} &  \small{(b)} 
   \end{tabular}
   \includegraphics[width=0.5\textwidth]{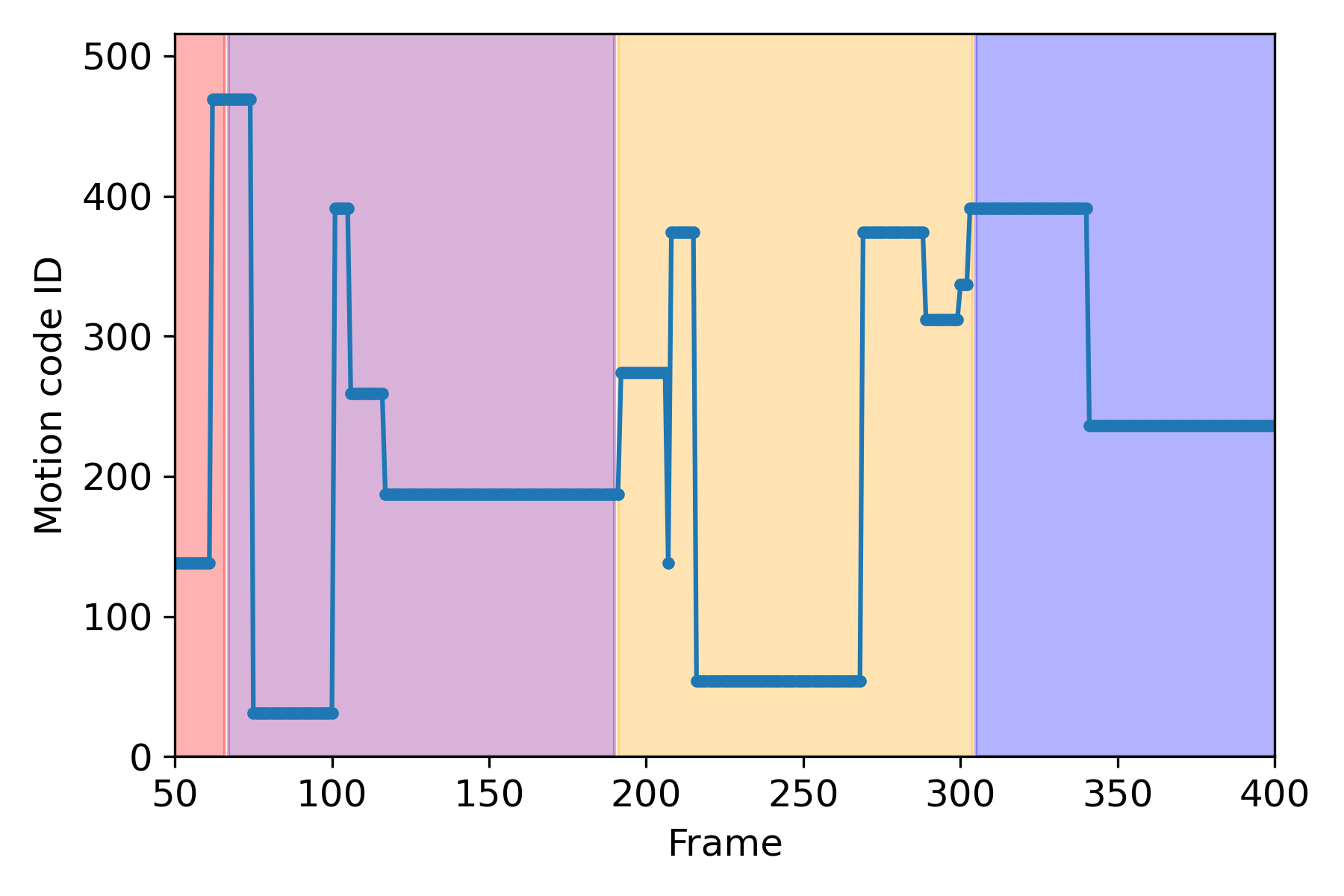}
   \\ \small{(c)}
   \caption{(a) Top-1 frames for each value vector, (b) Number of
    times counted as top-1 attention frames, which is used to extract keyframes, (c) Motion
    code transitions assigned to the keyframes}
    \label{fig:top1_attention}
\end{figure}

\begin{figure}[h]
   \centering
   \includegraphics[width=1\textwidth]{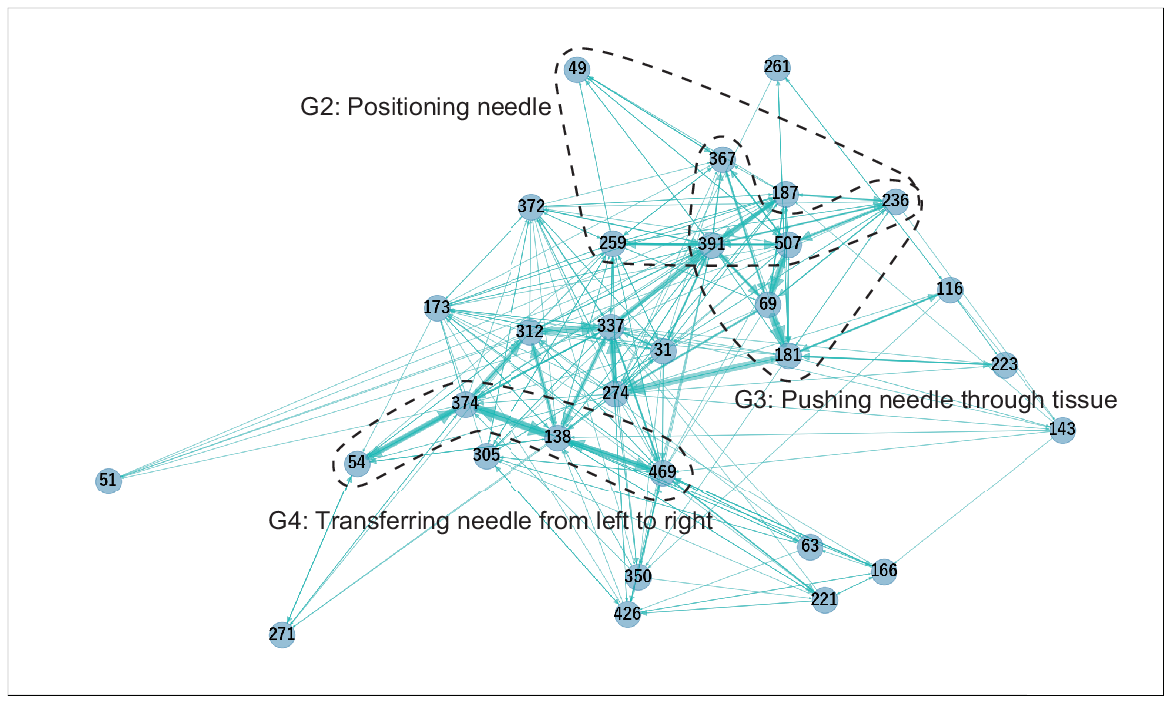}
    \caption{Graph of keyframe motion codes is constructed by the
           transition. The numbers indicate the IDs of motion codes. 
           The relationship between the annotation labels is
           visualized in the graph}
    \label{fig:transition_JIGSAWS_label}
\end{figure}

In order to extract sparse keyframes, we propose counting the top-1 frame weight for each
frame. Fig. \ref{fig:top1_attention}(a) shows the top-1 frame weight for each value
vector, and Fig. \ref{fig:top1_attention}(b) is the number of top-1 frames larger than one, which
are used as keyframes. The motion code transitions assigned to the keyframes are
shown in Fig. \ref{fig:top1_attention}(c). By counting these transitions, the relationships between them can
be used to form a graph, as shown in Fig. \ref{fig:transition_JIGSAWS_label}. The
positions of the motion codes within the graph are calculated by the
Fruchterman-Reingold force-directed algorithm~\cite{fruchterman1991graph}
implemented in NetworkX~\cite{SciPyProceedings_11}. The motion codes that correspond to
three of the annotation labels are enclosed by dotted lines. The labels
that occur subsequently (G2 and G3) share the motion codes, while the labels
that occur at the different phases of the motion (G4) use separate motion codes.

\section{Experiments}\label{sec:Experiments}

As a dataset that contains multiple components in each sequence of human motion,
the JIGSAWS dataset~\cite{gao2014jhu}, which contains video and kinematic data
for robotic surgical tasks performed by operators with different skill levels,
is first used in our experiments. In the present study, the suturing task, with 39
trials produced using eight subjects, is evaluated. For annotation purposes, 10
labels are used to describe actions at each frame, and the skill level (novice,
intermediate, or expert) is given for each subject. The split of training and
test sequences is the same as that of the setting in \cite{ismail2018evaluating} for
cross-validation.

In the kinematic data provided in JIGSAWS, six-dimensional poses for the
grippers and the gripper angles of the two robot arms at each frame are used as
the input/output data, which means that 14 variables are calculated for each frame.
Each recorded sequence is between one and three minutes long and is captured at
30 fps at $640 \times 480$ pixels. First, the kinematic data are used as the
input of the proposed method. The hyperparameters of the model in the present paper are
as follows. The number of attention layers is $N_{SA}=6$ for both the encoder and
decoder. The motion codebook has 512 vectors of $D(=256)$
dimensions, and the attention widths are $M=100$ and $M=10$ for the encoder and decoder, respectively.

\begin{figure}[h]
   \centering
   \begin{tabular}{cc}
   \includegraphics[width=0.45\textwidth]{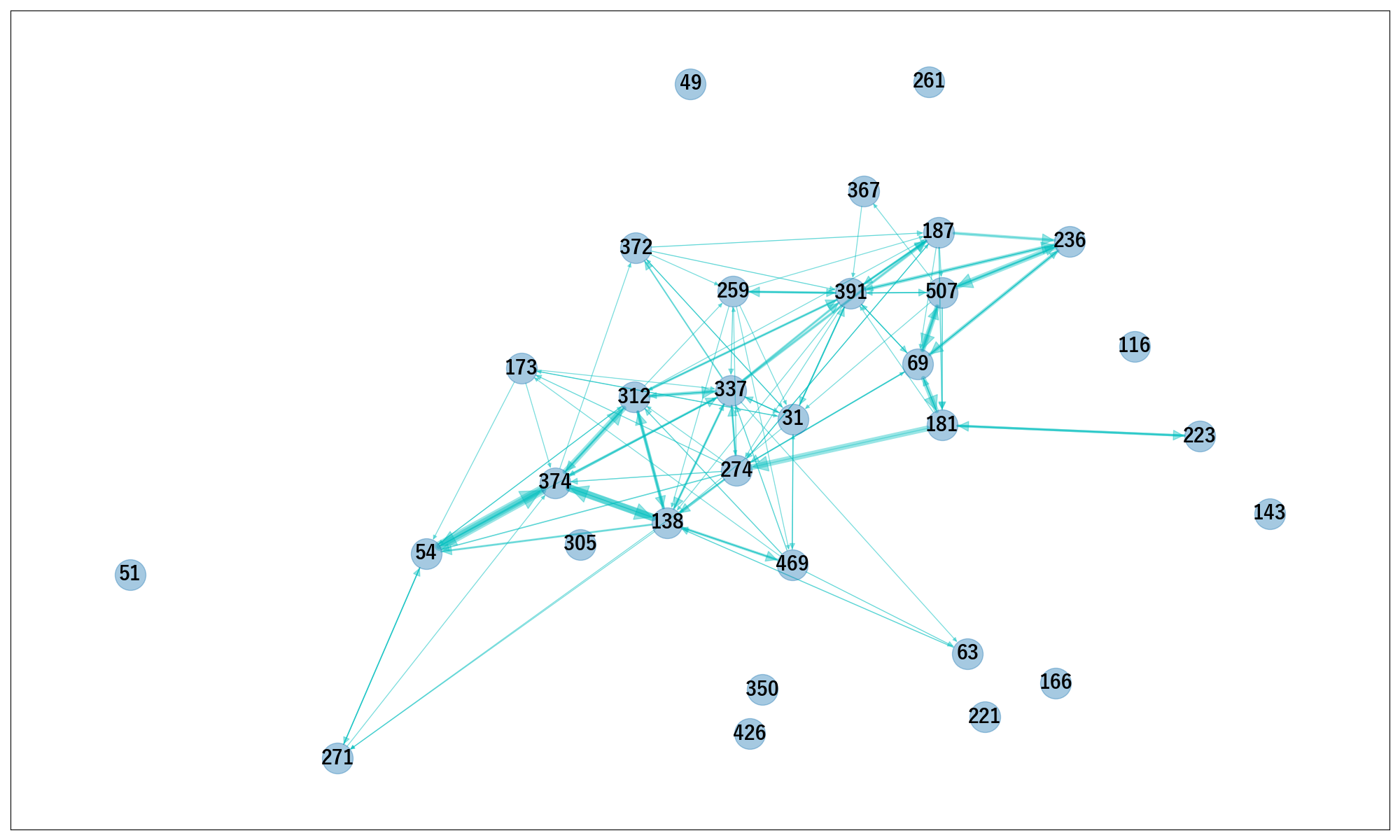} &
   \includegraphics[width=0.45\textwidth]{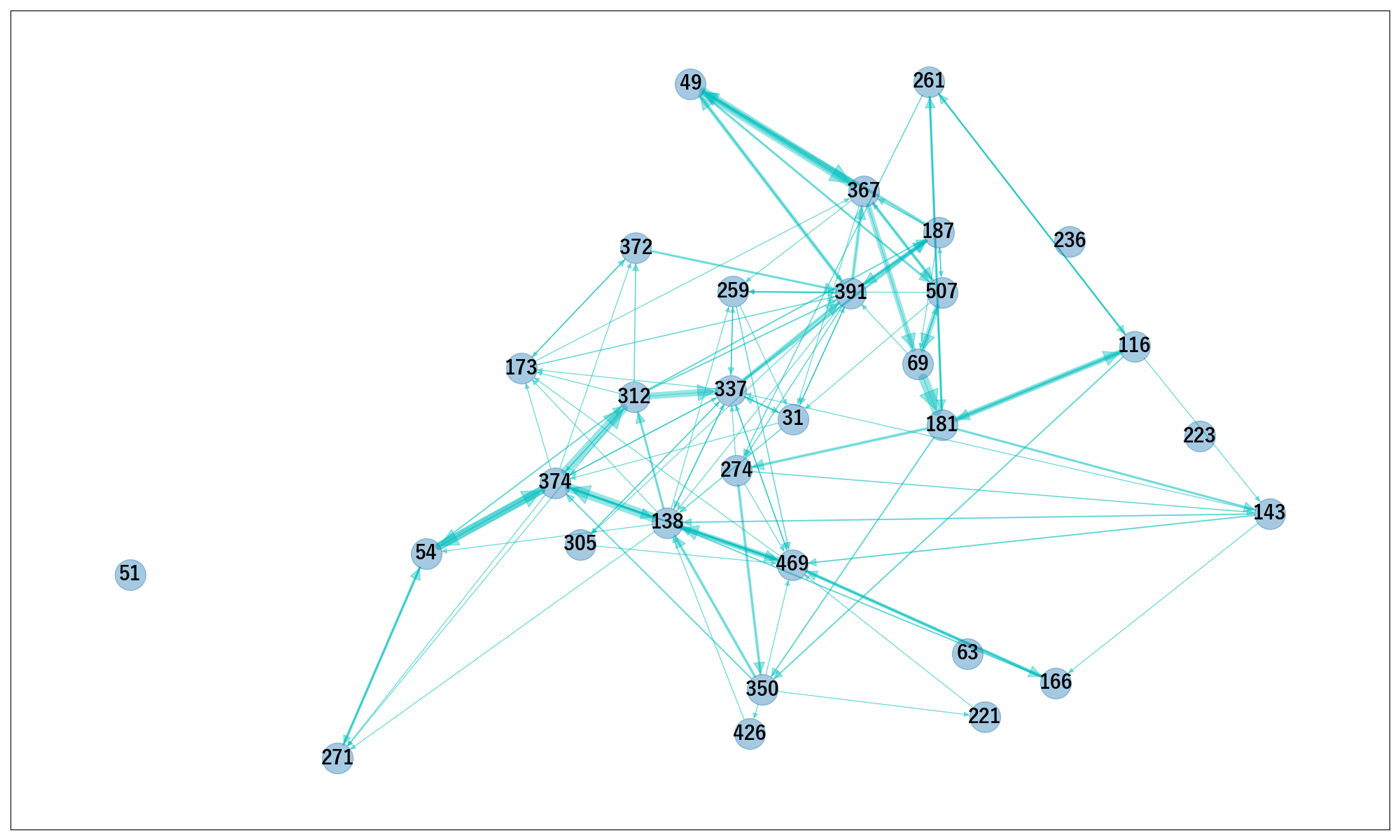} \\
    \small{(a) Subject C} &  \small{(b) Subject F}
   \end{tabular}
    \caption{Keyframe motion code graphs for two subjects for different sets
    of motion codes}
    \label{fig:transition_JIGSAWS_subjects}
\end{figure}

Differences between the subjects can be visualized by comparing their motion
code structures, which are shown in Fig. \ref{fig:transition_JIGSAWS_subjects}. Here, half of the motion codes are shared by both subjects, and the
others are only used by one of the subjects, which indicates that the motion
codes can express both the similarities and uniqueness of the subjects. 

\subsection{Evaluating the latent space by linear probing of multiple tasks}\label{sec:Evaluating the latent space by linear probing of multiple tasks}

For quantitative evaluation of the latent space generated by the proposed
method, the task of recognizing annotated labels was tested by using the motion
codes as the input. The purpose was to evaluate whether the generated motion codes,
which are trained by reconstructing the output vectors, contain useful
information. Since fine-tuning the trained network to a specific task is not an
appropriate method by which to evaluate the usefulness of motion codes, linear probing with
a simple linear layer, in which the backbone network to generate latent space is
fixed, was applied.

Since the motion codes have sufficient structure to allow them to reconstruct
human motion, the generated motion codes are expected to be applicable to
multiple tasks without optimization to specific tasks. In the present study, two
tasks, action segmentation and skill classification, are tested with the JIGSAWS
dataset. The former is a task that assigns action
labels for each frame, and the latter involves classifying each sequence to
one of the skill levels. The head blocks are trained for these tasks by a single
linear layer. 

\begin{figure}[h]
   \centering
   \includegraphics[width=1\textwidth]{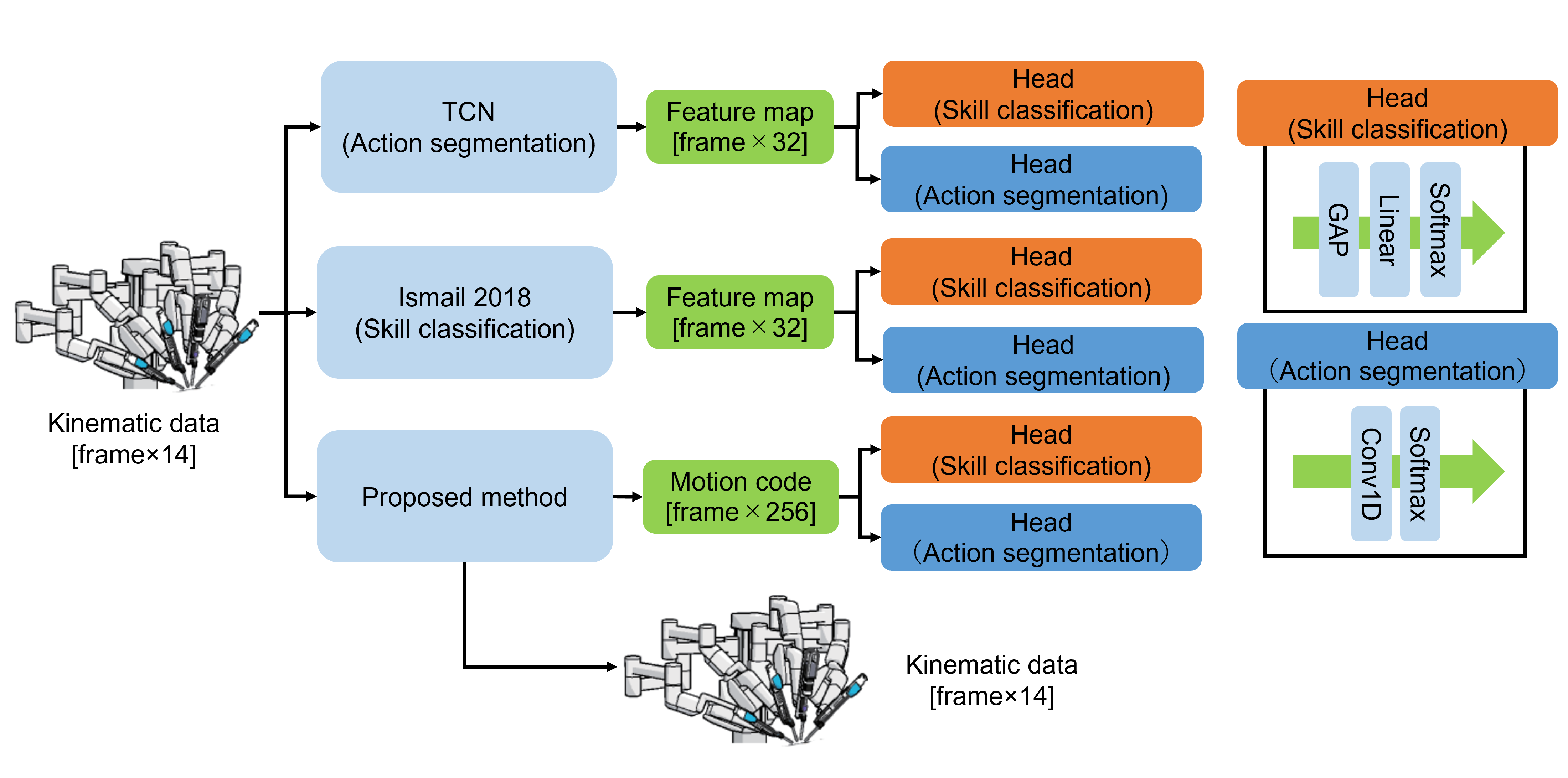}
    \caption{Evaluation of motion codes by linear probing with JIGSAWS kinematic
    inputs}
    \label{fig:linear_probe}
\end{figure}

As the baseline methods, the network for skill classification by Ismail
et al.~\cite{ismail2018evaluating} and temporal convolutional networks
(TCNs)~\cite{lea2016temporal} for action segmentation are compared. As shown in
Fig. \ref{fig:linear_probe}, the backbone and head blocks for the designed tasks are
trained end-to-end for each method, and the head for the other task is trained
as linear probing. Except for the input dimension, the head block architectures
are the same for the three methods. The input vector of the linear layers is the
quantized feature vector of the motion codes. The linear layer of action segmentation uses a part of the sequence as the input and is implemented as a 1D
convolution. The 1D convolution of the head for action segmentation uses 500
frames around each frame.

Tables \ref{tab:JIGSAWS_score_gesture} and \ref{tab:JIGSAWS_score_skill} show the
quantitative results of action segmentation and skill classification,
respectively. The accuracy is the percentage of correctly labeled frames, and
the edit score is the segmental edit distance~\cite{lea2016temporal} to measure the
correctness of the temporal ordering of actions. The micro average accuracy and
the macro average recall~\cite{ahmidi2017dataset} are computed as the average of
total correct predictions across all classes and the mean of true positive rates
for each class, respectively. The compared methods that are optimized to one of
the two tasks show good results for the optimized tasks, but the results for the
other task by linear probing are degraded. Although the proposed method is not
optimized for each task, the results are comparable to the method optimized for
each task. This proves that the motion codes extract effective information to
understand the temporal structure of motion and to explain the static
characteristics between motions.

\begin{table}[h]
   \centering
   \caption{Results of action segmentation for JIGSAWS kinematic inputs}
   \label{tab:JIGSAWS_score_gesture}
   \begin{tabular}{c|cc}
      \hline
      & Accuracy & Edit score \\
      \hline
      Ismail 2018\cite{ismail2018evaluating} & 64.9 & 55.9 \\
      TCN\cite{lea2016temporal} & 80.3 & 85.6 \\
      MotionCode (Proposed) & 82.6 & 65.7 \\
      \hline
   \end{tabular}
\end{table}

\begin{table}[h]
   \centering
   \caption{Results of skill classification for JIGSAWS kinematic inputs}
   \label{tab:JIGSAWS_score_skill}
   \begin{tabular}{c|cc}
      \hline
      & \begin{tabular}{c} Micro average\\ accuracy \end{tabular}
      & \begin{tabular}{c} Macro average\\ recall \end{tabular} \\
      \hline
      Ismail 2018\cite{ismail2018evaluating} & 99.4 & 99.6 \\
      TCN\cite{lea2016temporal} & 59.0 & 46.7 \\
      MotionCode (Proposed) & 94.9 & 94.9 \\
      \hline
   \end{tabular}
\end{table}

\subsection{Evaluation with video/3D skeleton inputs}\label{sec:Evaluation with video/3D skeleton inputs}

\subsubsection{Extracting motion codes from video}\label{sec:Extracting motion codes from video}

\begin{figure}[h]
   \centering
   \includegraphics[width=1\textwidth]{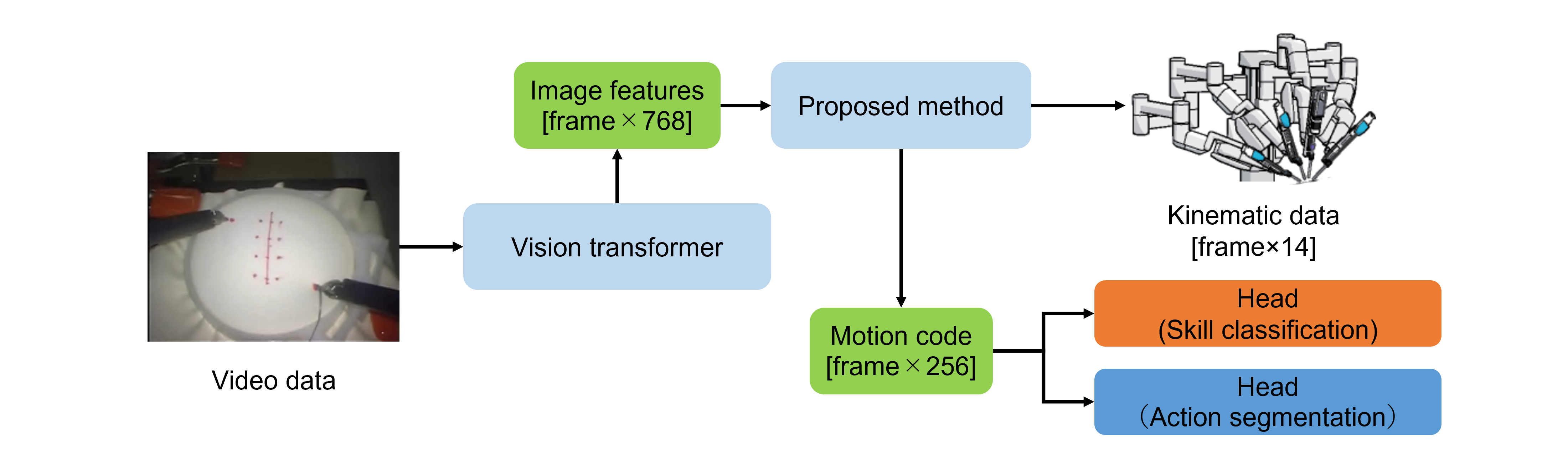}
    \caption{Evaluation of motion codes by linear probing with JIGSAWS video
    inputs}
    \label{fig:linear_probe_2_video}
\end{figure}

The next experiment is extracting motion codes from the video in the JIGSAWS
dataset, which means that the output modality is different from that of the input. In
this experiment, each frame in a video is encoded as a feature frame by frame using an image encoding block, and the feature is used as the input of the proposed
encoder-decoder model, as shown in Fig. \ref{fig:linear_probe_2_video}. The image encoding is implemented by
Vision transformer~\cite{dosovitskiy2020image}, and the parameters are fine-tuned to predict
the kinematic data of each frame. The dimension of the feature vector used as
input is 768. Since no implementation is available to test the tasks of action segmentation and skill classification from video, the proposed method is
compared with the methods tested under the same condition. The results obtained by MsM-CRF~\cite{ahmidi2017dataset} and 3D ConvNet~\cite{funke2019video}
are shown for comparison in Tables \ref{tab:JIGSAWS_score_gesture_i2m} and \ref{tab:JIGSAWS_score_skill_i2m} for action segmentation and skill
classification, respectively. Although the recognition by the proposed method is
linear probing without fine-tuning, the results of the proposed method are
comparable with those of the methods that are optimized for the respective
tasks. 

\begin{table}[h]
   \centering
   \caption{Results of action segmentation for JIGSAWS video inputs}
   \label{tab:JIGSAWS_score_gesture_i2m}
   \begin{tabular}{c|cc}
      \hline
      & \begin{tabular}{c} Micro average\\ accuracy \end{tabular}
      & \begin{tabular}{c} Macro average\\ recall \end{tabular} \\
      \hline
      MsM-CRF~\cite{ahmidi2017dataset} & 84.4 & 71.8 \\
      MotionCode (Proposed) & 82.5 & 70.9 \\
      \hline
   \end{tabular}
\end{table}

\begin{table}[h]
   \centering
   \caption{Results of skill classification for JIGSAWS video inputs}
   \label{tab:JIGSAWS_score_skill_i2m}
   \begin{tabular}{c|cc}
      \hline
      & \begin{tabular}{c} Micro average\\ accuracy \end{tabular}
      & \begin{tabular}{c} Macro average\\ recall \end{tabular} \\
      \hline
      3D ConvNet~\cite{funke2019video} & 100 & 100 \\
      MotionCode (Proposed) & 94.9 & 96.5 \\
      \hline
   \end{tabular}
\end{table}

\subsubsection{Extracting motion codes from a 3D skeleton}\label{sec:Extracting motion codes from a 3D skeleton}

\begin{figure}[h]
   \centering
   \includegraphics[width=1\textwidth]{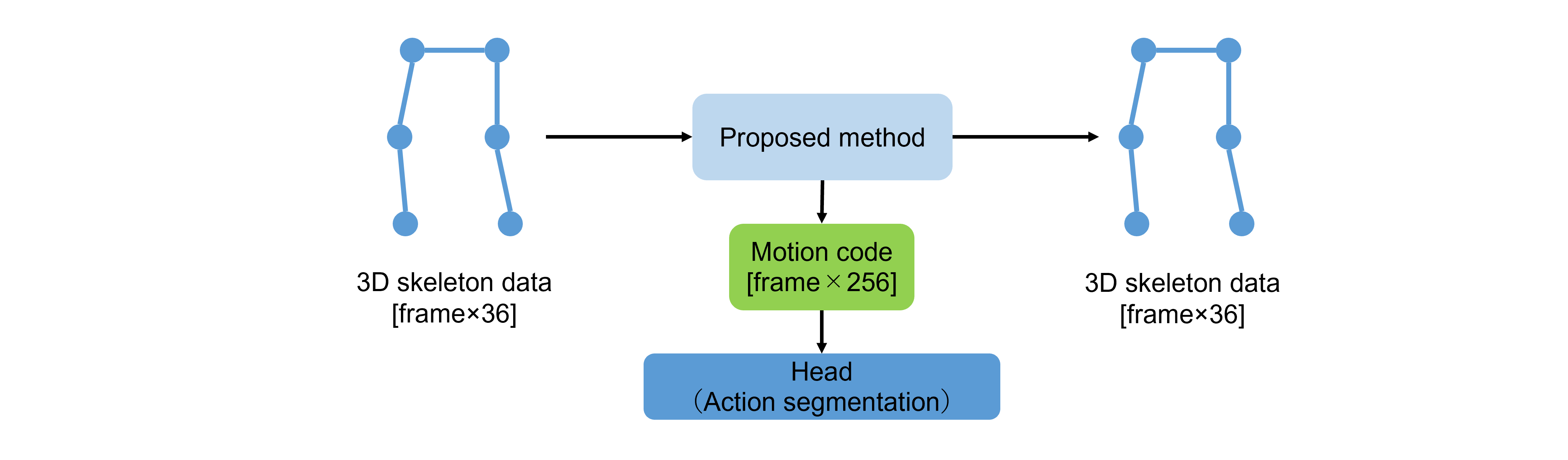}
    \caption{Evaluation of motion codes by linear probing with 3D skeleton inputs}
    \label{fig:linear_probe_2_HuGaDB}
\end{figure}

The next experiment is extracting the motion codes from a 3D skeleton dataset
(HuGaDB)~\cite{chereshnev2017hugadb}, as shown in Fig. \ref{fig:linear_probe_2_HuGaDB}. The dataset
contains the motions of the lower half of the body, such as walking, taking
stairs up or down, and sitting down, with segmentation and annotation. The
input skeleton consists of six joints with a three-axis accelerometer and three-axis
gyroscope data at each joint. The motions are captured at 60Hz for 18 subjects,
and their lengths are 300 frames to 12,000 frames. The split of training and test sequences
is the same as the setting in \cite{filtjens2022skeleton}. The results are
evaluated by accuracy and F1@50. The former is sample-wise
accuracy, as described above, and the latter is the F1-score of
classifying segments by 50\% intersection over union (IoU) overlap with respect
to the corresponding expert annotation~\cite{lea2017temporal}.
Table \ref{tab:HuGaDB_segmentation} shows the results compared with the methods tested
in \cite{filtjens2022skeleton}. The accuracy of the proposed method is
comparable to those of the other methods that optimized for the segmentation task,
although the F1@50 score is lower than those of the others, which is the case because
no device is provided to avoid over-segmentation in linear probing,
as is the edit score in Table \ref{tab:JIGSAWS_score_gesture}.

\begin{table}[h]
   \centering
   \caption{Results of action segmentation for the 3D skeleton dataset (HuGaDB)}
   \label{tab:HuGaDB_segmentation}
   \begin{tabular}{c|cc}
      \hline
      & Accuracy & F1@50 \\
      \hline
      Bi-LSTM	& 86.1	& 81.5 \\
      TCN	& 88.3	& 56.8 \\
      ST-GCN	& 88.7	& 67.7 \\
      MS-TCN	& 86.8	& 89.9 \\
      MS-GCN	& 90.4	& 93.0 \\
      MotionCode (Proposed) & 87.5 & 58.5 \\
      \hline
   \end{tabular}
\end{table}

\subsection{Ablation study}\label{sec:Ablation study}

\subsubsection{Generating motion codes without restriction}\label{sec:Generating motion codes without restriction}

\begin{figure}[h]
  \centering
  \begin{tabular}{cc}
   \includegraphics[width=0.45\textwidth]{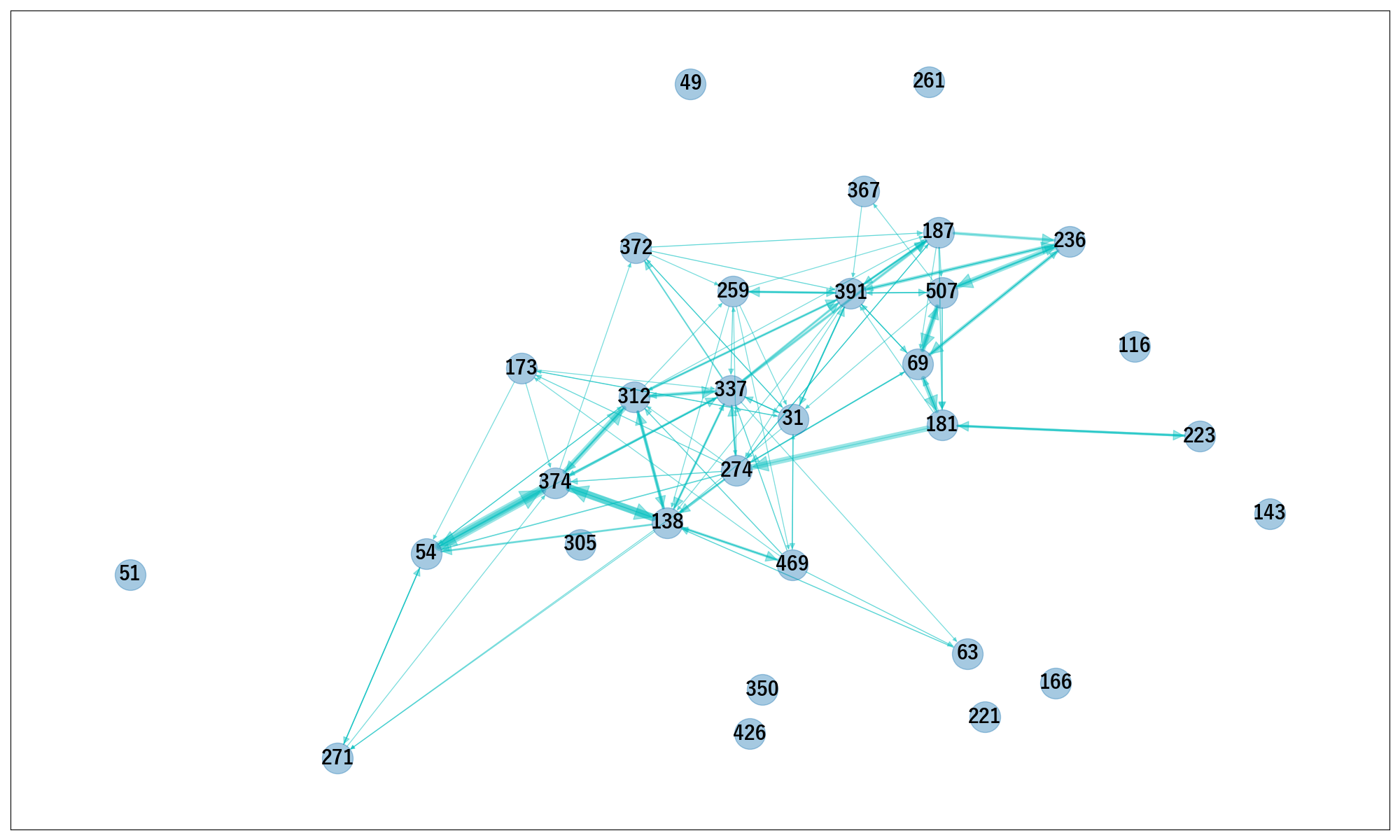} &
   \includegraphics[width=0.45\textwidth]{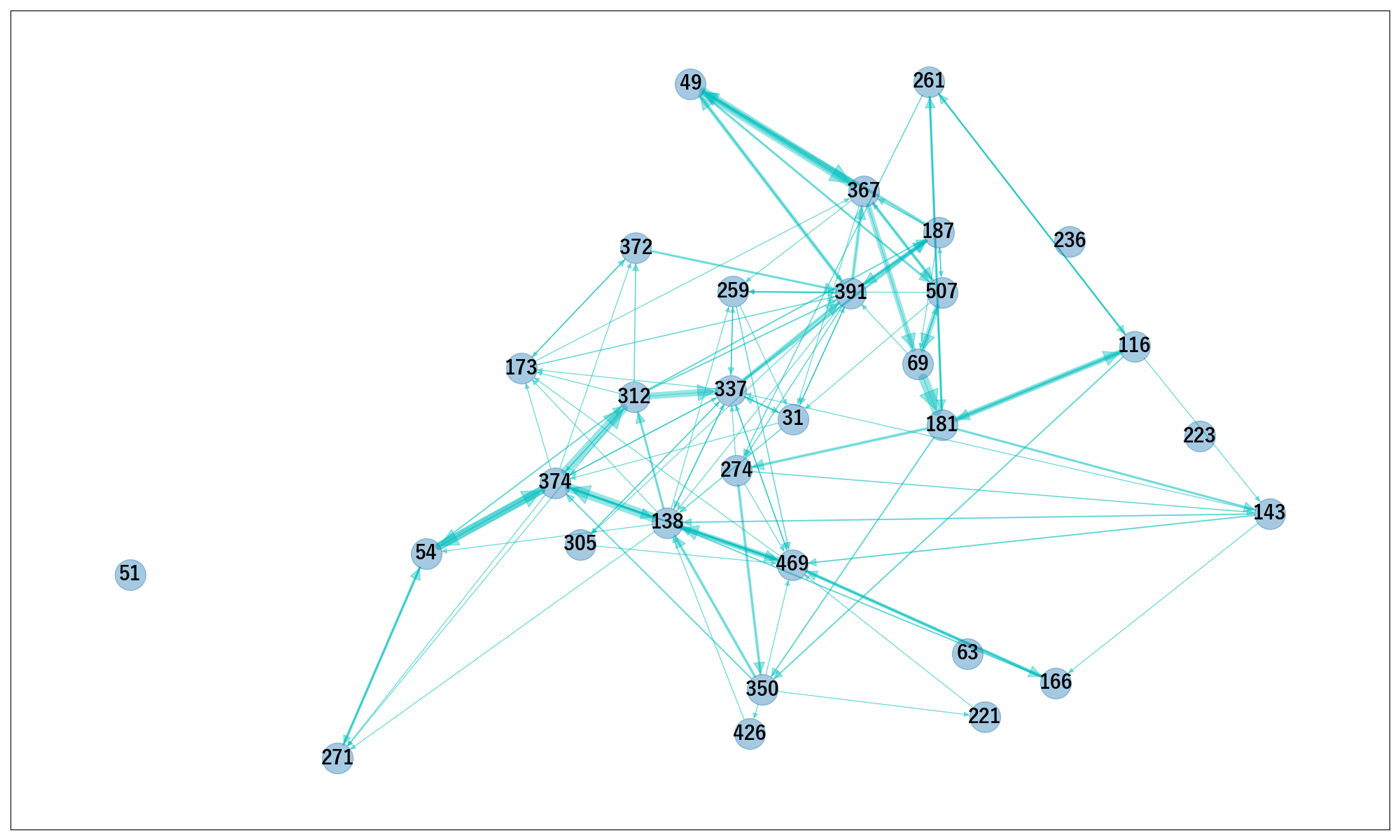} \\
  \small{(a) Subject C (with restriction)} & \small{(b) Subject F (with restriction)} \\
   \includegraphics[width=0.45\textwidth]{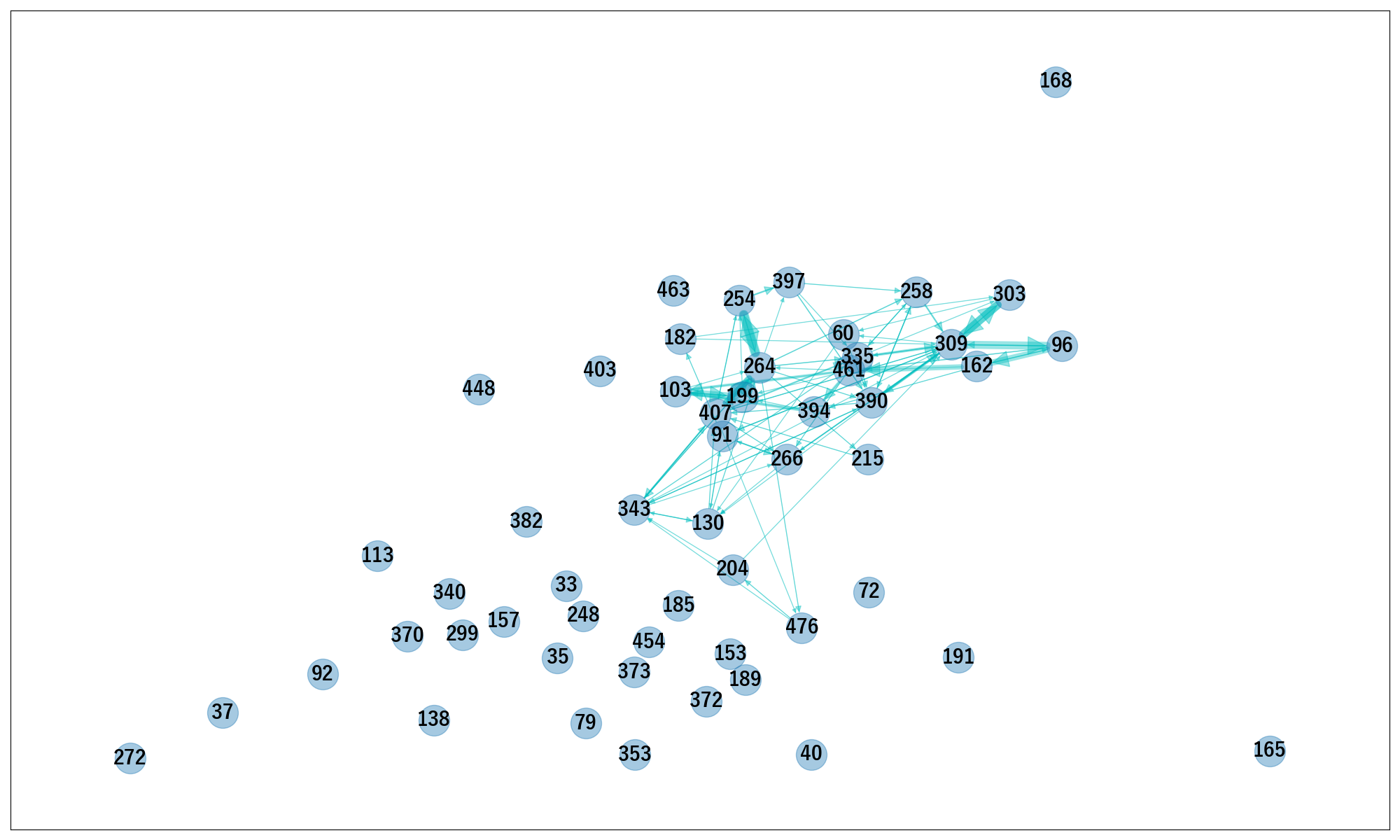} &
   \includegraphics[width=0.45\textwidth]{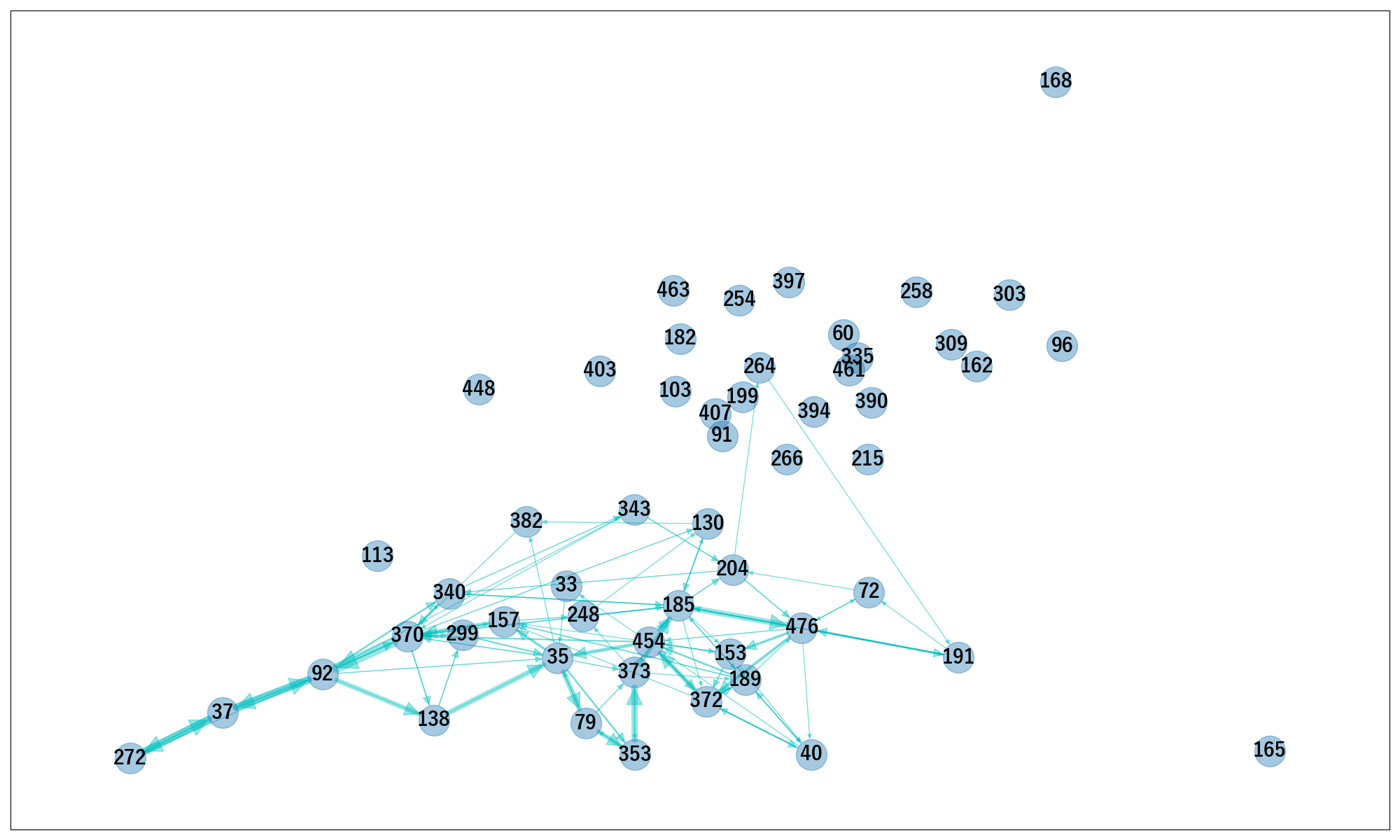} \\
   \small{(c) Subject C (without restriction)} &  \small{(d) Subject F (without restriction)} \\
  \end{tabular}
    \caption{Motion code graphs for the results trained with / without restricting
      motion codes}
    \label{fig:comparison_transition_JIGSAWS_restriction}
\end{figure}

Restricting motion codes in the training introduced in Section \ref{sec:Restricting motion codes} is expected
to encourage motion code sharing. 
Fig. \ref{fig:comparison_transition_JIGSAWS_restriction} shows a comparison of the results of training with and without restriction.
In the restricted case, most of the motion codes used by subjects C and F are shared, which indicates that
the motions are translatable to each other between the subjects.
On the other hand, in the non-restricted case, most of the motion codes are not shared between subjects C and F. The reconstruction loss
decreases more easily by using different motion codes than by sharing motion codes, however, using split codes is not desirable for translatability.
The result shows the need for restrictions to share motion codes between subjects.

\subsubsection{Effect of the attention width of the decoder}\label{sec:Effect of the attention width of the decoder}

\begin{figure}[h]
   \centering
   \begin{tabular}{cc}
   \includegraphics[width=0.45\textwidth]{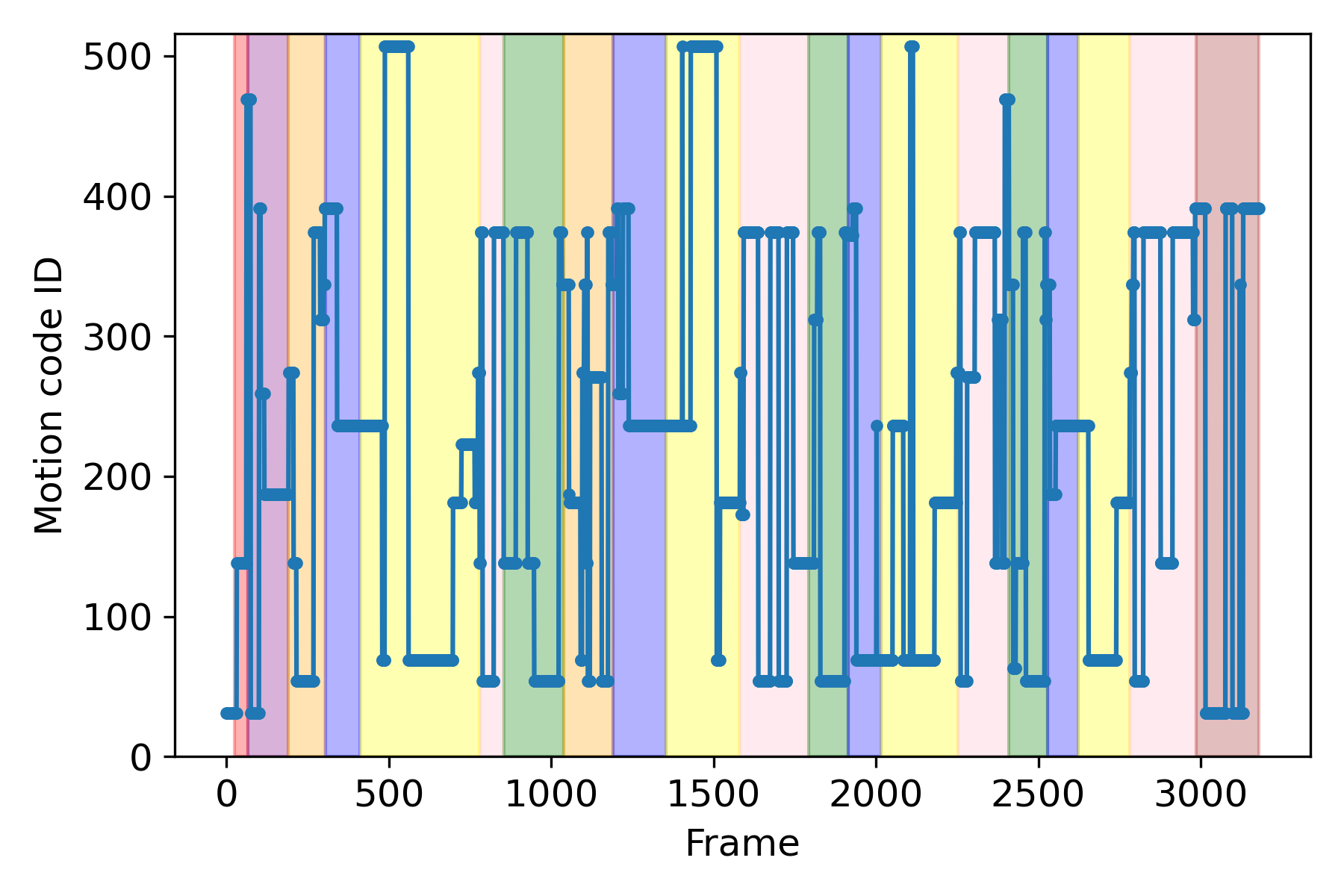} &
   \includegraphics[width=0.45\textwidth]{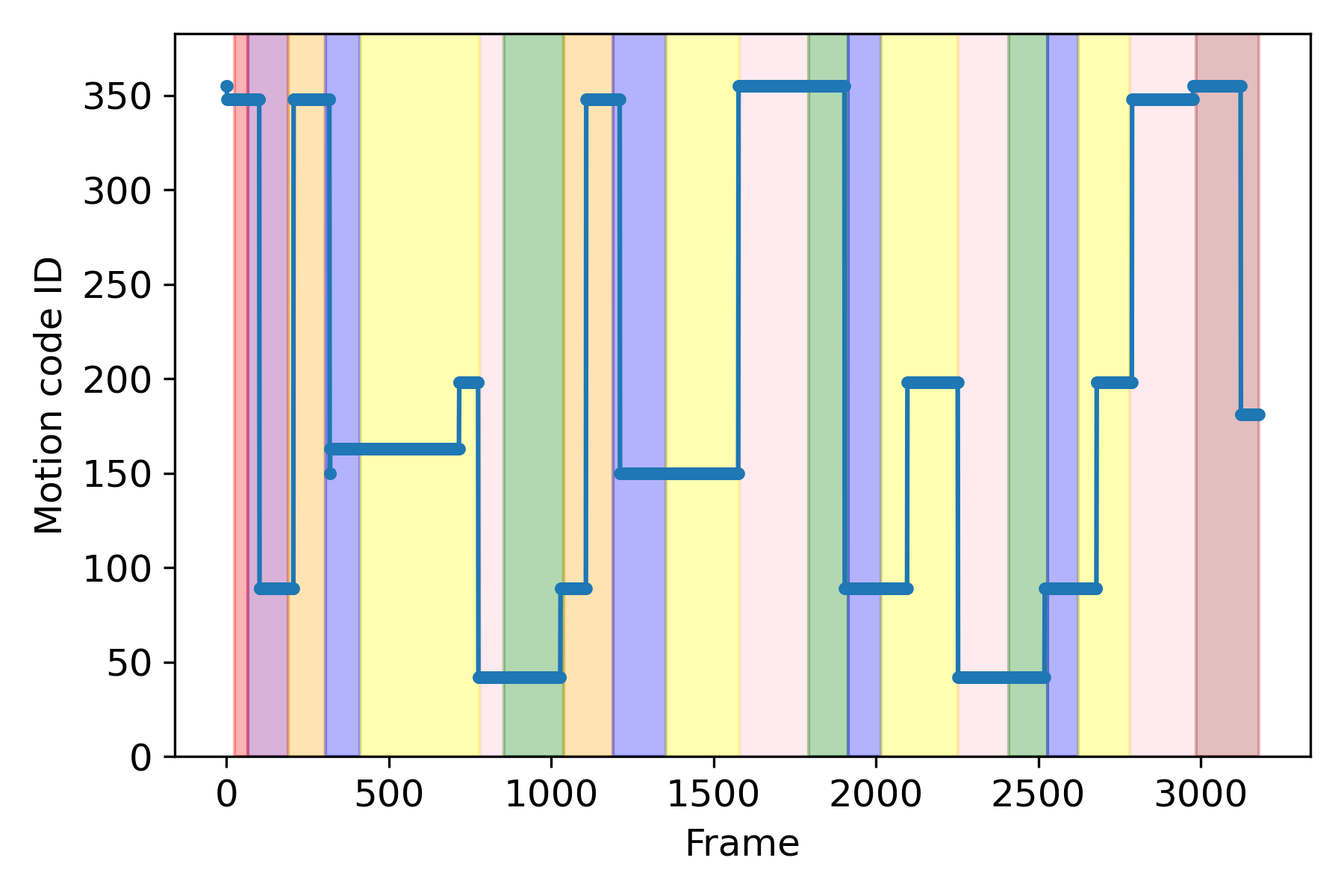} \\
    \small{(a)} &  \small{(b)}
   \end{tabular}
    \caption{Transition of keyframe motion codes with (a) $M=10$ and (b) $M=100$}
    \label{fig:dec100_id}
\end{figure}

\begin{figure}[h]
  \centering
  \begin{tabular}{cc}
   \includegraphics[width=0.45\textwidth]{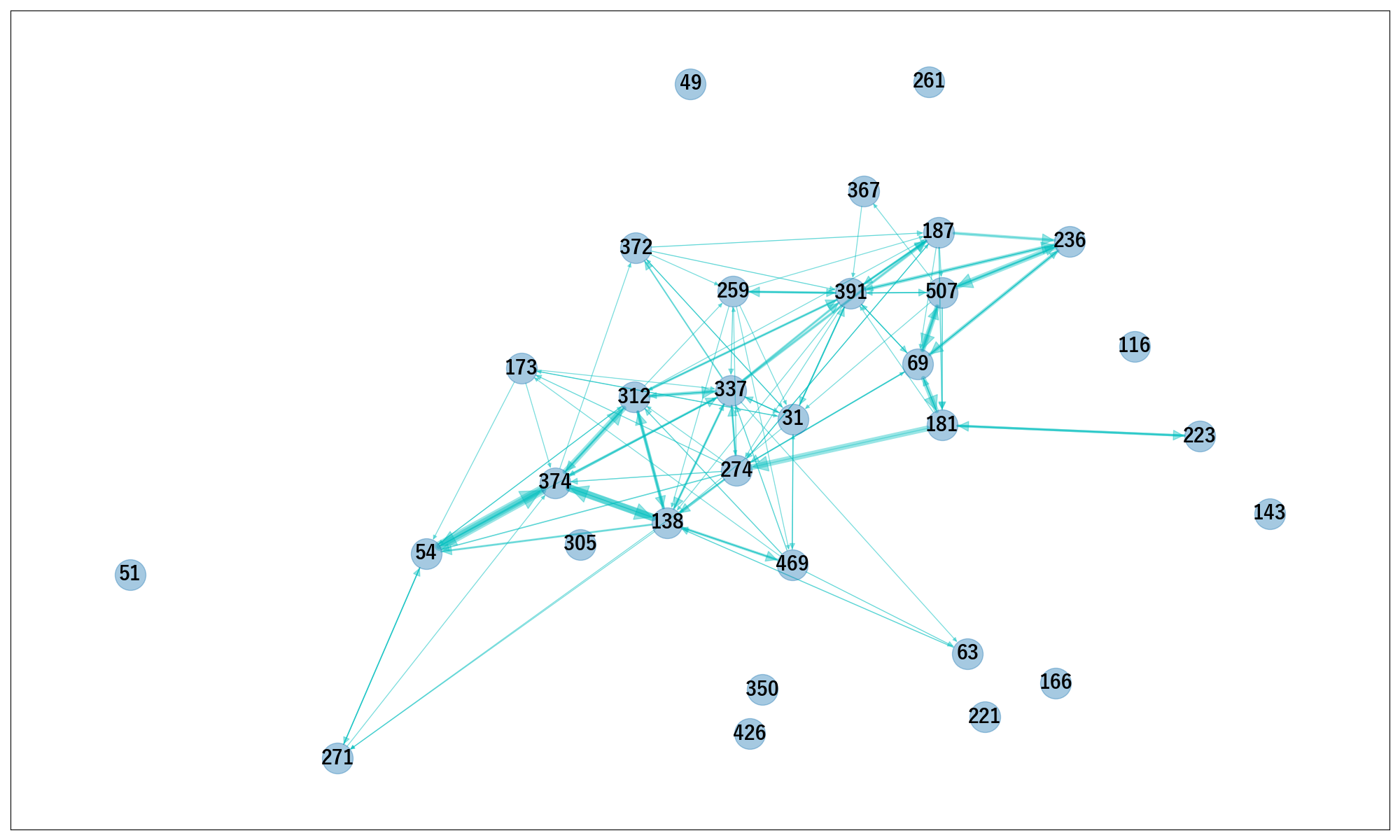} &
   \includegraphics[width=0.45\textwidth]{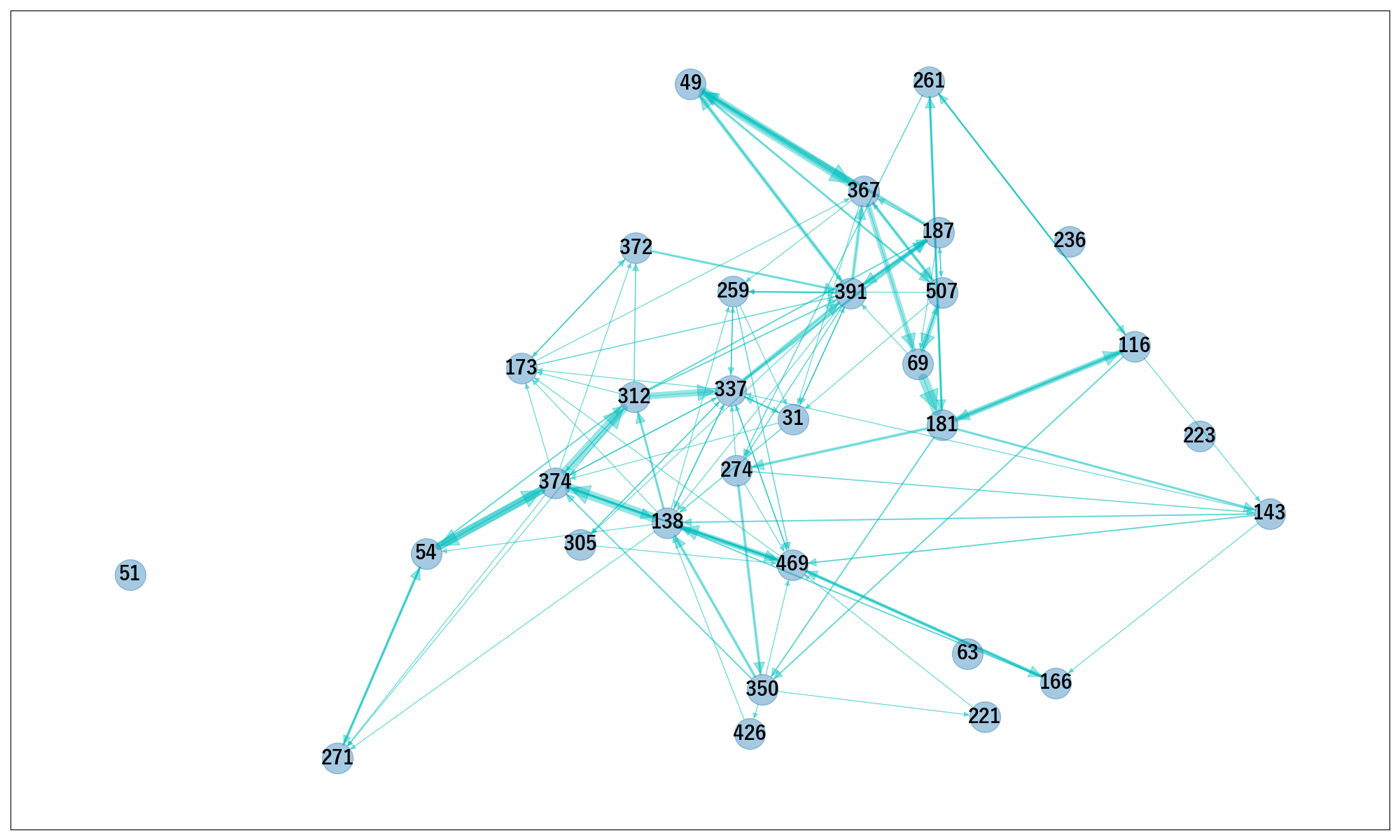} \\
   \small{(a) Subject C ($M=10$)} &  \small{(b) Subject F ($M=10$)} \\
   \includegraphics[width=0.45\textwidth]{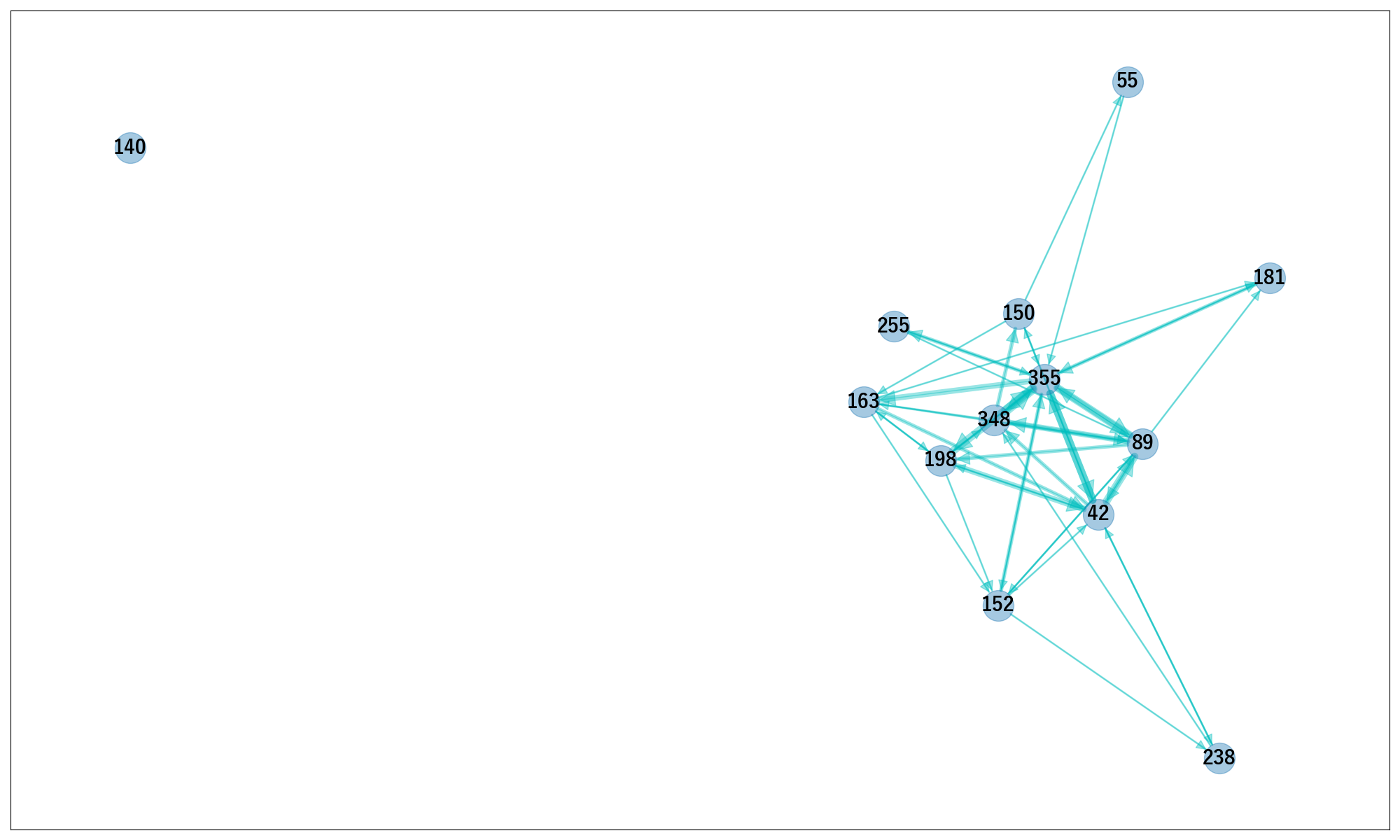} &
   \includegraphics[width=0.45\textwidth]{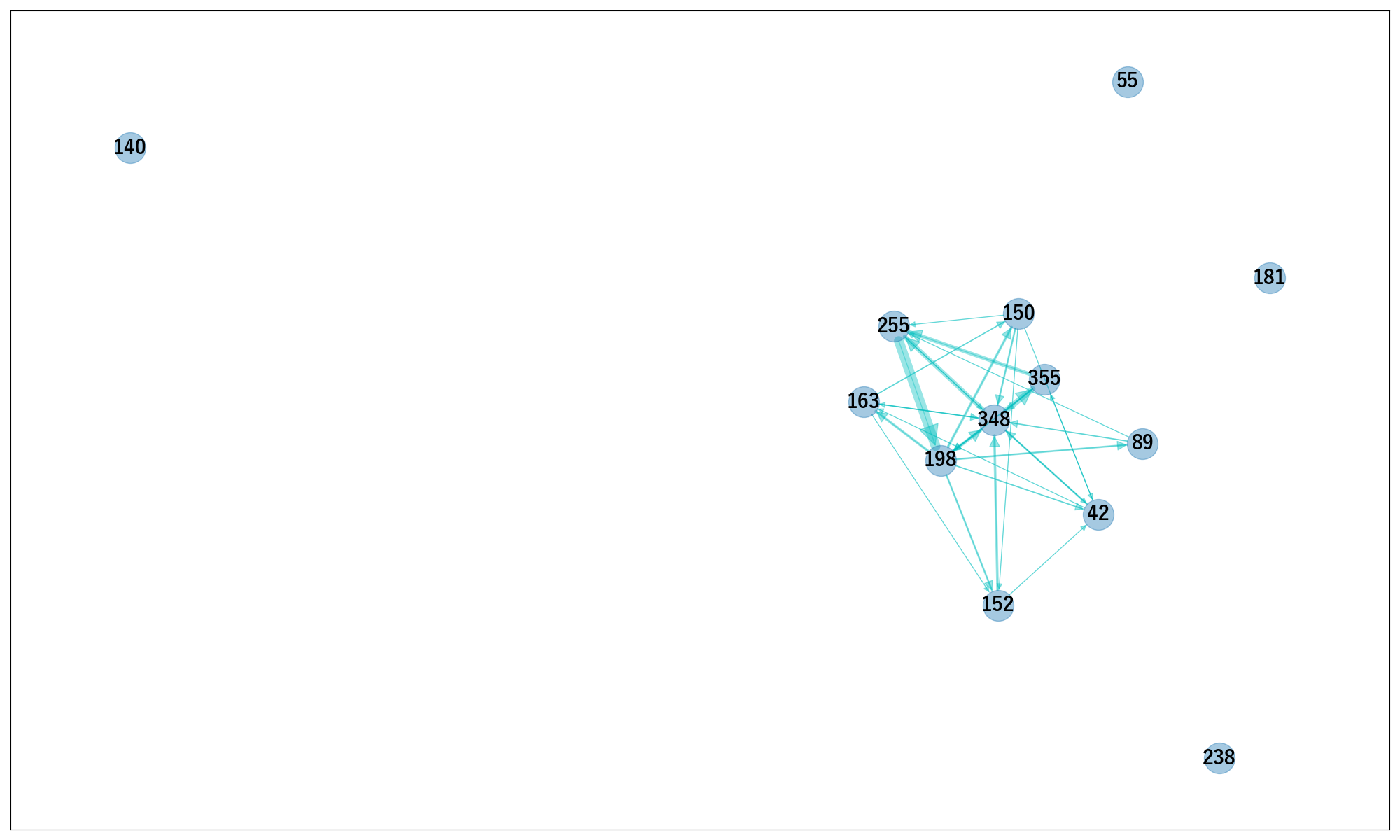} \\
   \small{(c) Subject C ($M=100$)} &  \small{(d) Subject F ($M=100$)} \\
  \end{tabular}
    \caption{Motion code graphs with $M=10$ and $M=100$}
    \label{fig:transition_JIGSAWS_dec100}
\end{figure}

Attention width $M$ of the decoder attention layers defines how many preceding
frames are referred to generate motion from motion codes. The number of reference frames is expected to affect the frame range of high attention weight
to a keyframe. In this experiment, how the sequence of extracted motion codes
changes is tested if the reference frames are wide. The attention width $M$ is
changed to 100 while $M = 10$ in Fig. \ref{fig:top1_attention}. Fig. \ref{fig:dec100_id}
shows the entire sequence of keyframe motion codes with (a) $M=10$ and with (b) $M=100$. Since
the total attention widths of six attention layers are 60 and 600, respectively,
the attention is concentrated on sparse keyframes and the number of keyframes
is reduced in the latter case. Since the boundaries between different motion
codes are still close to those of the annotated labels, the motion
codes have a relationship to the understanding of motion by humans.
Tables \ref{tab:JIGSAWS_score_gesture_dec100} and \ref{tab:JIGSAWS_score_skill_dec100} shows the results of action segmentation and skill classification by attention width of the decoder, respectively. Since the performance of action segmentation with $M=100$ is similar to the case with $M=10$, the motions can be recognized in the case of sparse keyframes. 
On the other hand, the result of skill classification is degraded with $M=100$ from the case with $M=10$. 
The reason is considered to be because, as shown in Fig. \ref{fig:transition_JIGSAWS_dec100}, in the case of $M=100$, most motion codes are shared by subjects compared to the case of $M=10$. Consequently, the
granularity of motion codes can be controlled by the parameter of the attention
width, but the side effect of sharing motion codes is an issue to be investigated in
a future study.

\begin{table}[h]
   \centering
   \caption{Results of action segmentation for JIGSAWS kinematic inputs based on the attention width of the decoder}
   \label{tab:JIGSAWS_score_gesture_dec100}
   \begin{tabular}{c|cc}
      \hline
      Attention width $M$ & Accuracy & Edit score \\
      \hline
      $10$	& 82.6 & 65.7 \\
      $100$	& 82.7 & 64.3 \\
      \hline
   \end{tabular}
\end{table}

\begin{table}[h]
   \centering
   \caption{Results of skill classification for JIGSAWS kinematic inputs based on the attention width of the decoder}
   \label{tab:JIGSAWS_score_skill_dec100}
   \begin{tabular}{c|cc}
      \hline
      Attention width $M$ 
      & \begin{tabular}{c} Micro average\\ accuracy \end{tabular}
      & \begin{tabular}{c} Macro average\\ recall \end{tabular} \\
      \hline
      $10$	& 94.9 & 94.9 \\
      $100$	& 88.5 & 88.2 \\
      \hline
   \end{tabular}
\end{table}

\section{Conclusion}\label{sec:Conclusion}

In the present paper, we proposed an encoder-decoder model that extracts
frame-wise motion codes as discrete components in order to provide intermediate
representations of human motions. The motion codes are extracted in a
self-supervised manner without using any manual annotations. We found that
generating a discrete representation contributes to extracting sparse keyframes
and visualizing the relationship between the components. We then evaluated the
effectiveness of the motion codes by applying them to multiple recognition
tasks. Since the motion codes extracted by the proposed method yield results
comparable to those that have features optimized by supervised learning, the
results of the present study show that the motion codes contain sufficient information to effectively
understand human motions.

One of the issues is optimizing the granularity of the motion codes for various
tasks. When the attention width is narrow, the granularity is smaller than that
of annotation labels, which makes it difficult to find one-to-one correspondence
with a user-defined label. Furthermore, since various levels of granularity can
be considered when attempting to explain human behavior, one of our future areas of study
will be to generate a hierarchical structure of motion codes. More specifically,
since structures can be extracted in a self-supervised manner from the dataset
used, the goal will be to construct a hierarchical motion code structure without
the need for hand-crafted level explanations of human behavior. In addition,
since the advantages of the sparse and discrete features make them easily
specified by users, another direction of future study will be to use motion
codes to generate new motions that may be difficult to explain by user-defined
labels. Such motions can be expected to be useful for robotics and computer
graphics applications.

\bmhead{Acknowledgments}

This work was supported by JSPS JP22H00545, JP22H05002 and NEDO JPNP20006 (New
Energy and Industrial Technology Development Organization) in Japan.

\section*{Declarations}

\textbf{Conflict of interest }
The authors declare that they have no conflict of interest. 

\begin{appendices}
\renewcommand{\thefigure}{\arabic{figure}}
\setcounter{figure}{12}

\section{Keyframe motion code graph}\label{sec:Keyframe motion code graph}

\begin{figure}[h]
  \centering
  \begin{tabular}{ccc}
  \includegraphics[width=0.3\textwidth]{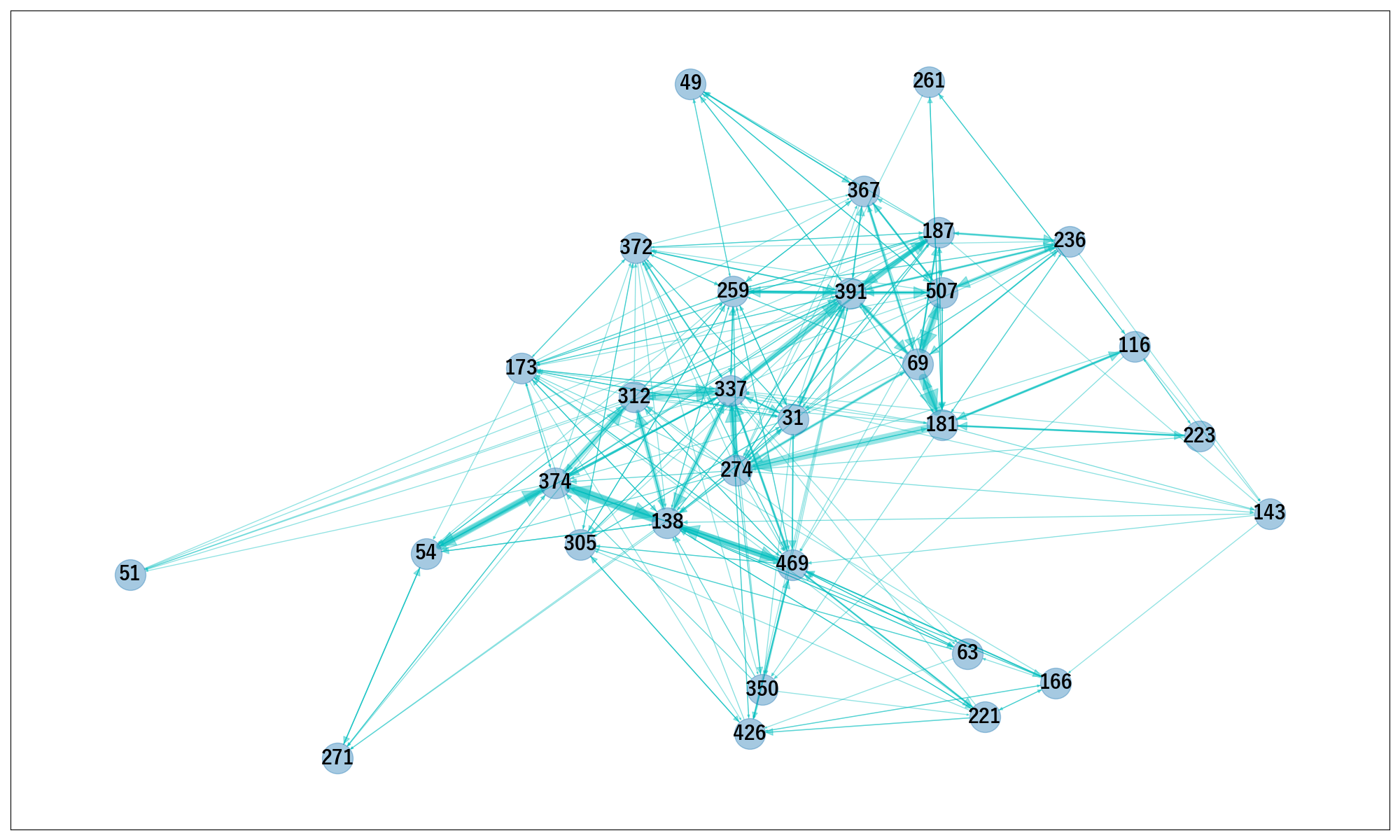} &
  \includegraphics[width=0.3\textwidth]{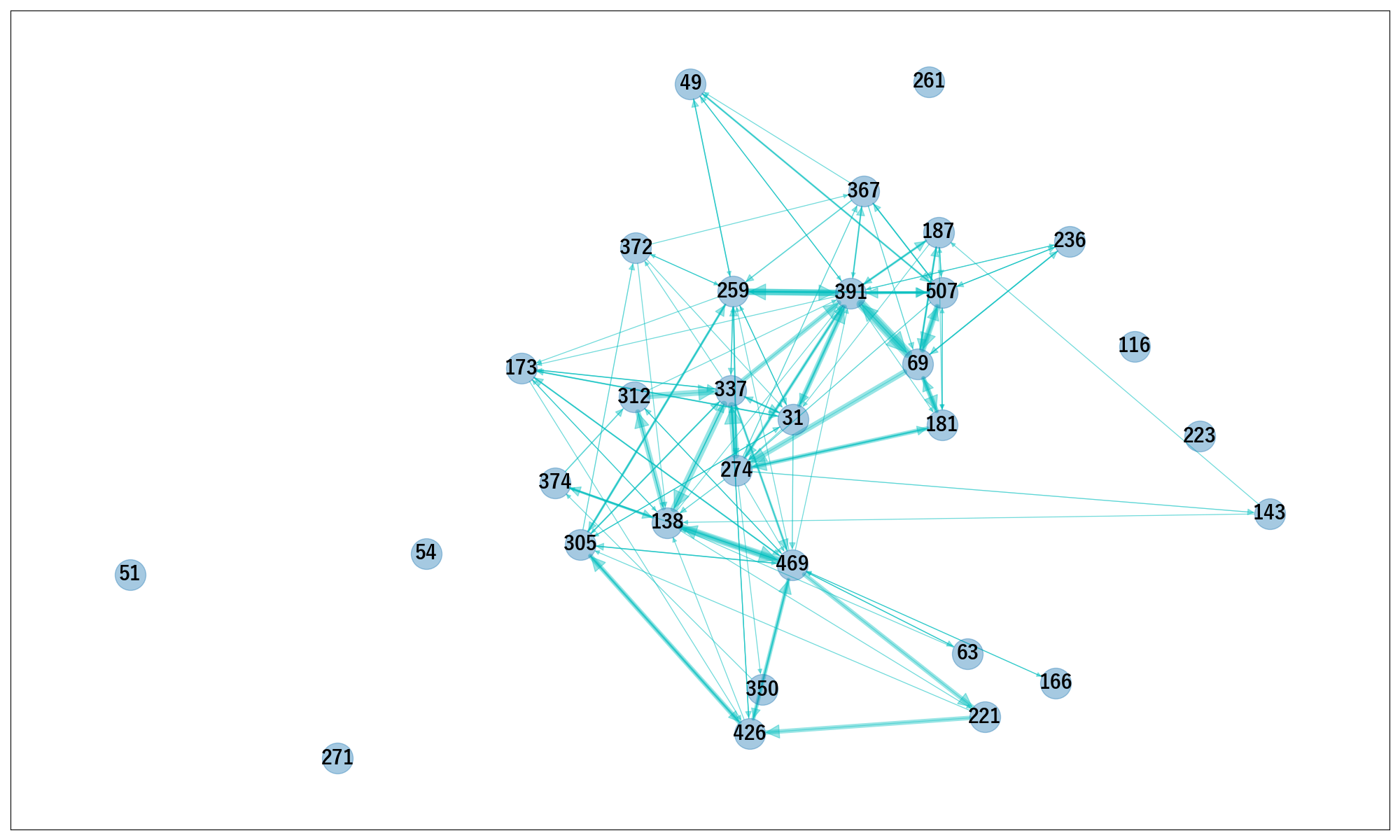} &
  \includegraphics[width=0.3\textwidth]{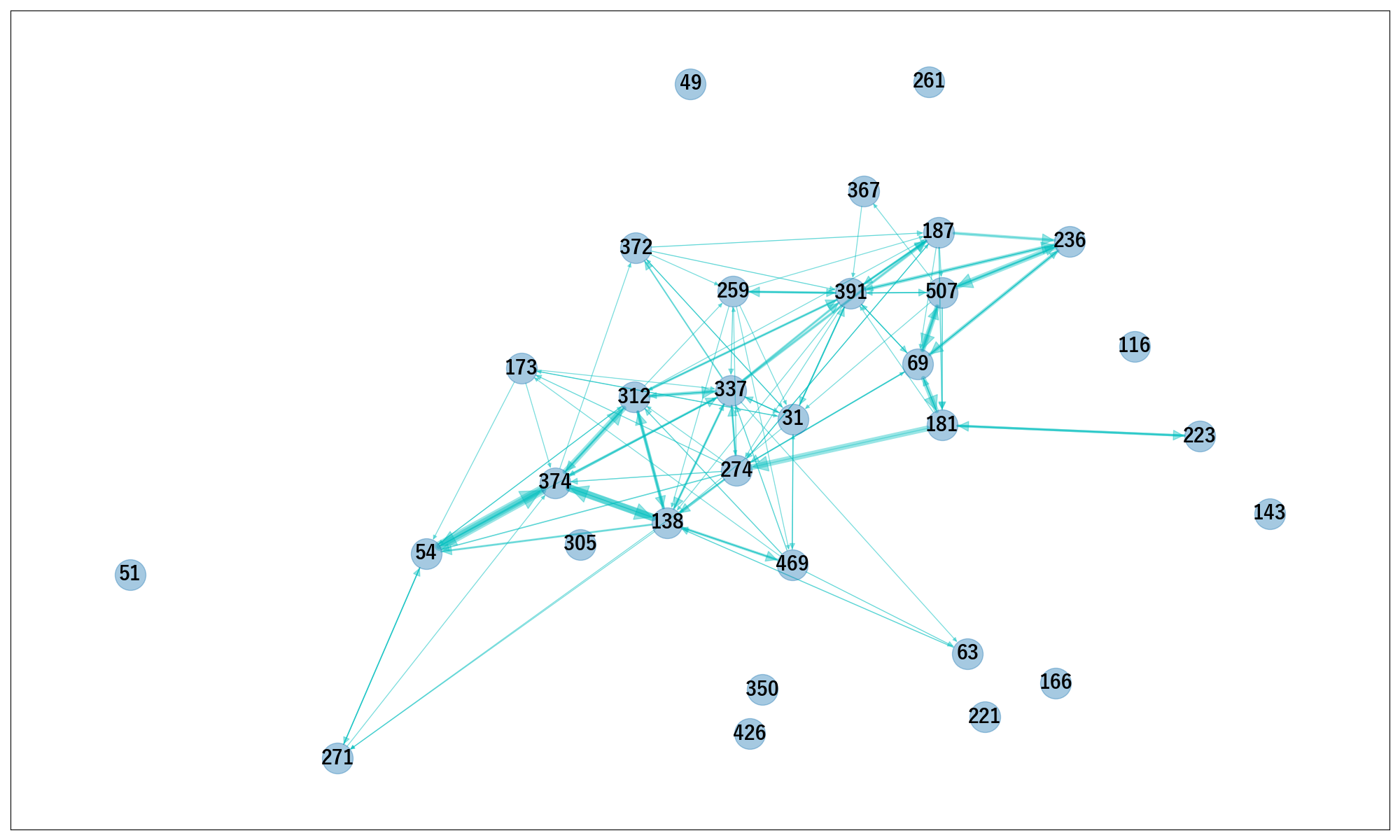} \\
   \small{(a) All subjects} &  \small{(b) Subject B} &  \small{(c) Subject C} \\
  \includegraphics[width=0.3\textwidth]{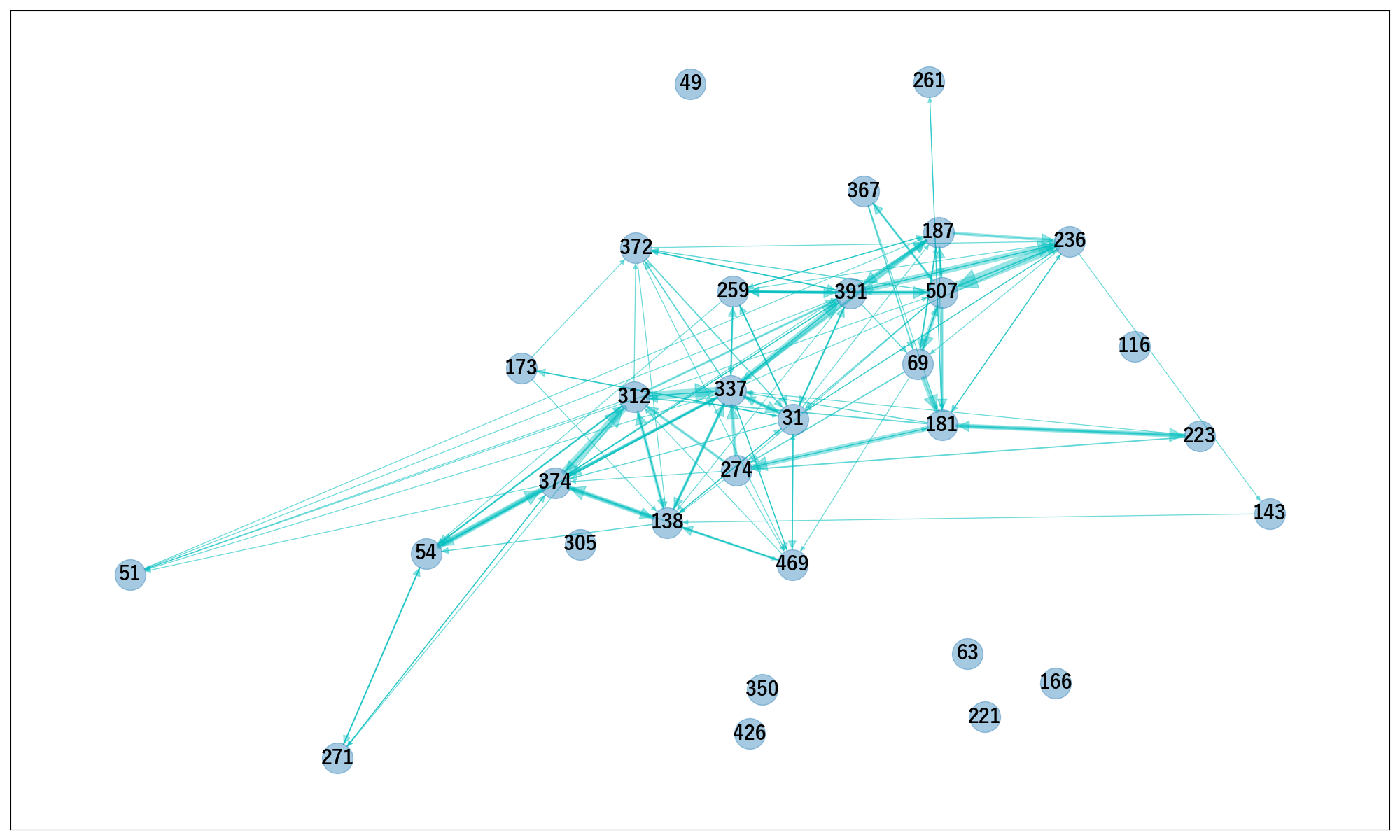} &
  \includegraphics[width=0.3\textwidth]{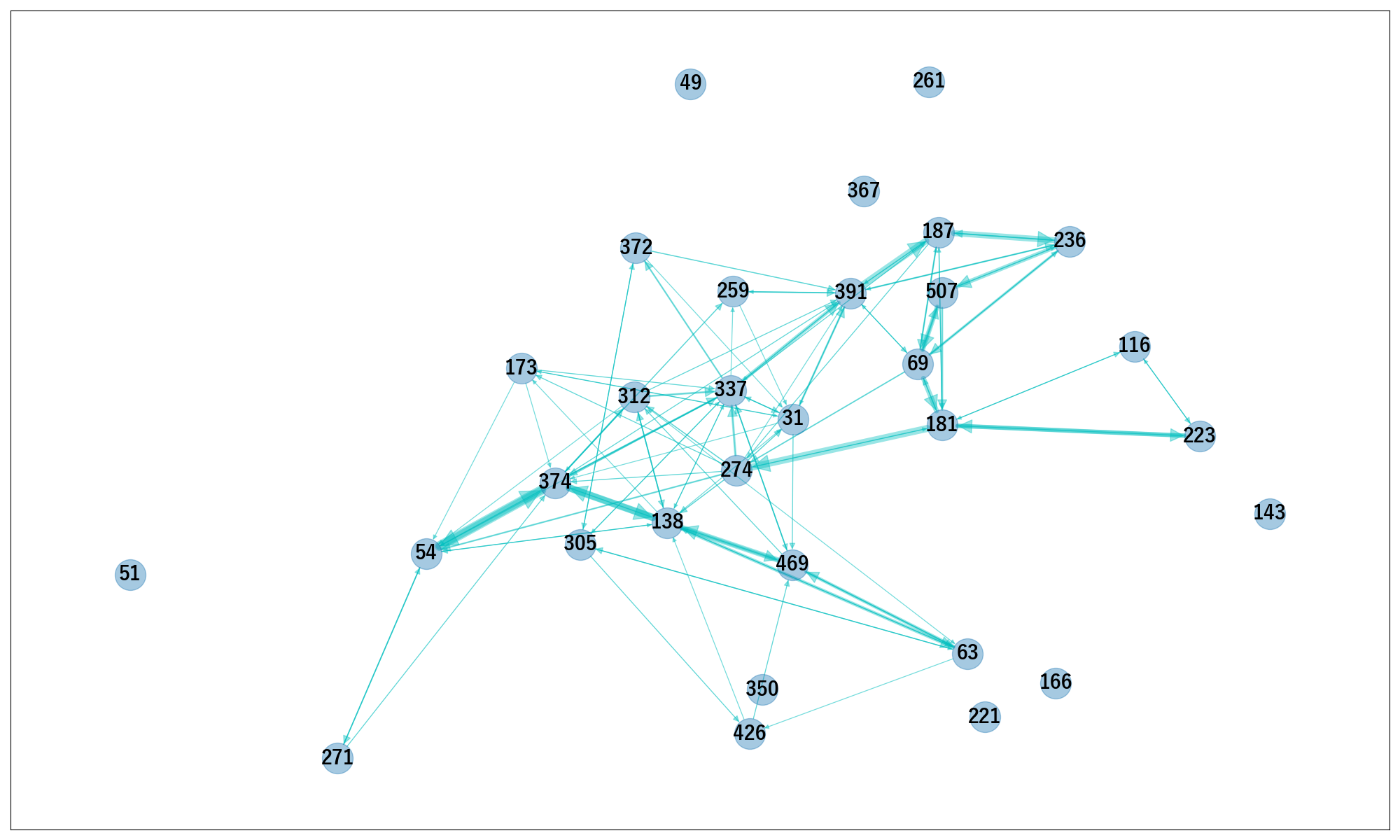} &
  \includegraphics[width=0.3\textwidth]{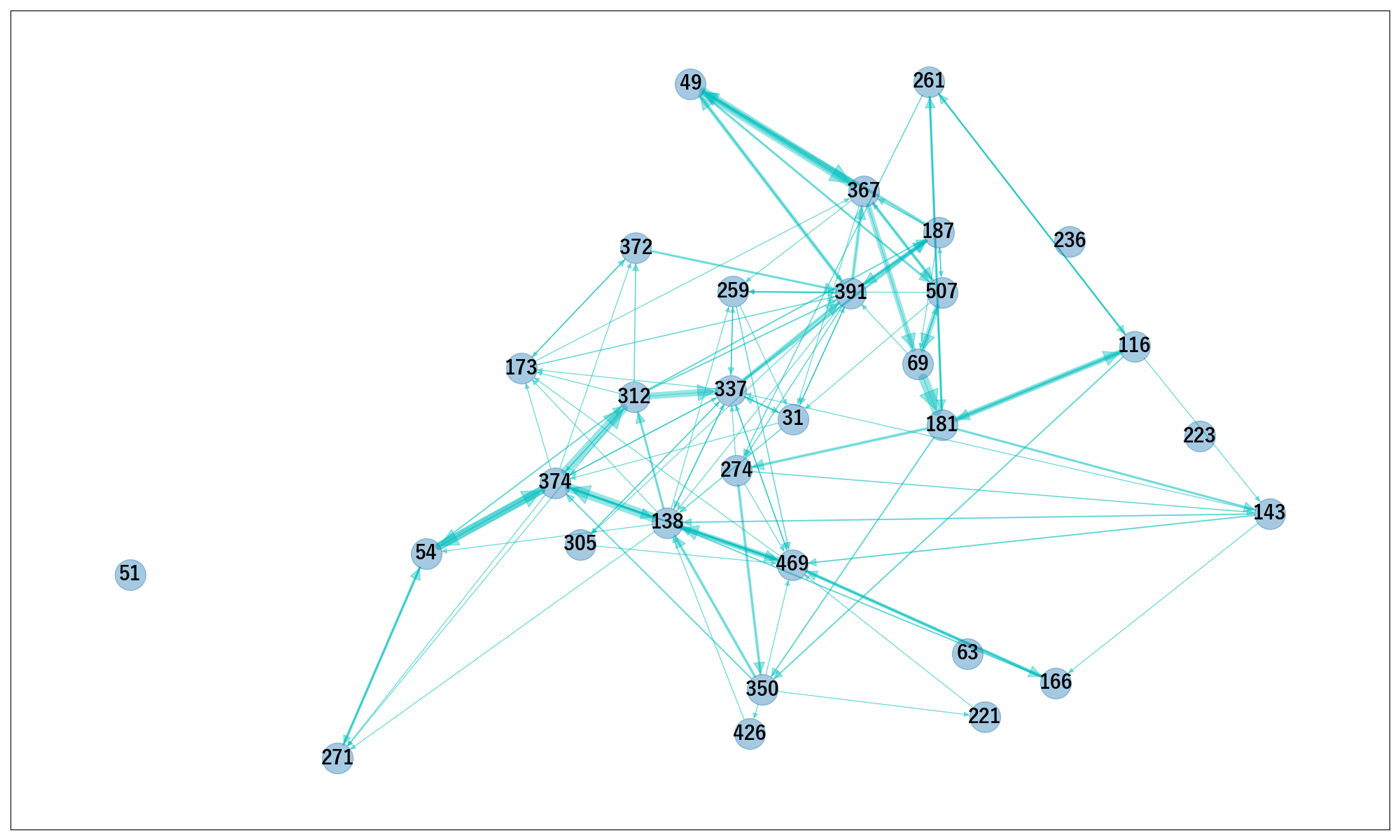} \\
   \small{(d) Subject D} &  \small{(e) Subject E} &  \small{(f) Subject F} \\
  \includegraphics[width=0.3\textwidth]{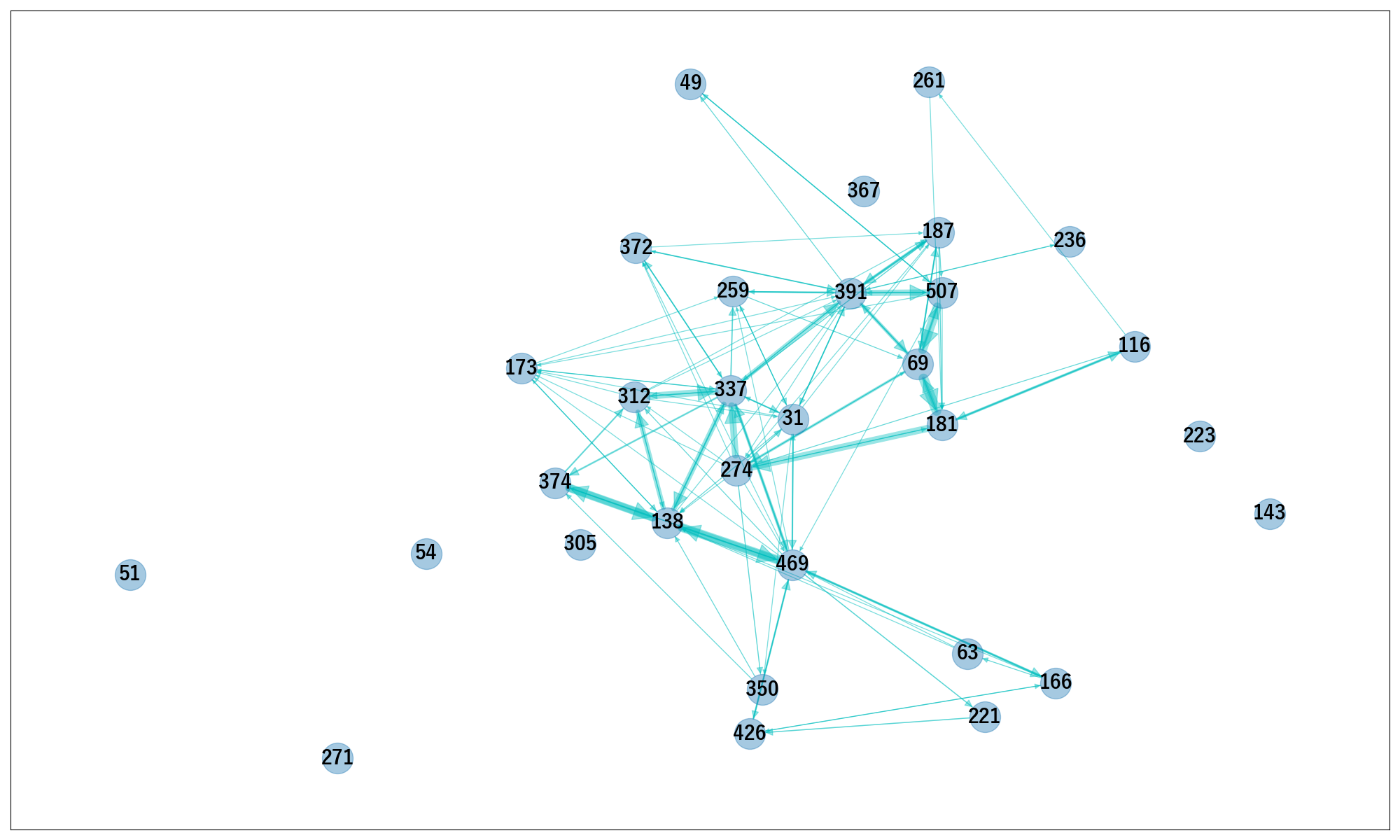} &
  \includegraphics[width=0.3\textwidth]{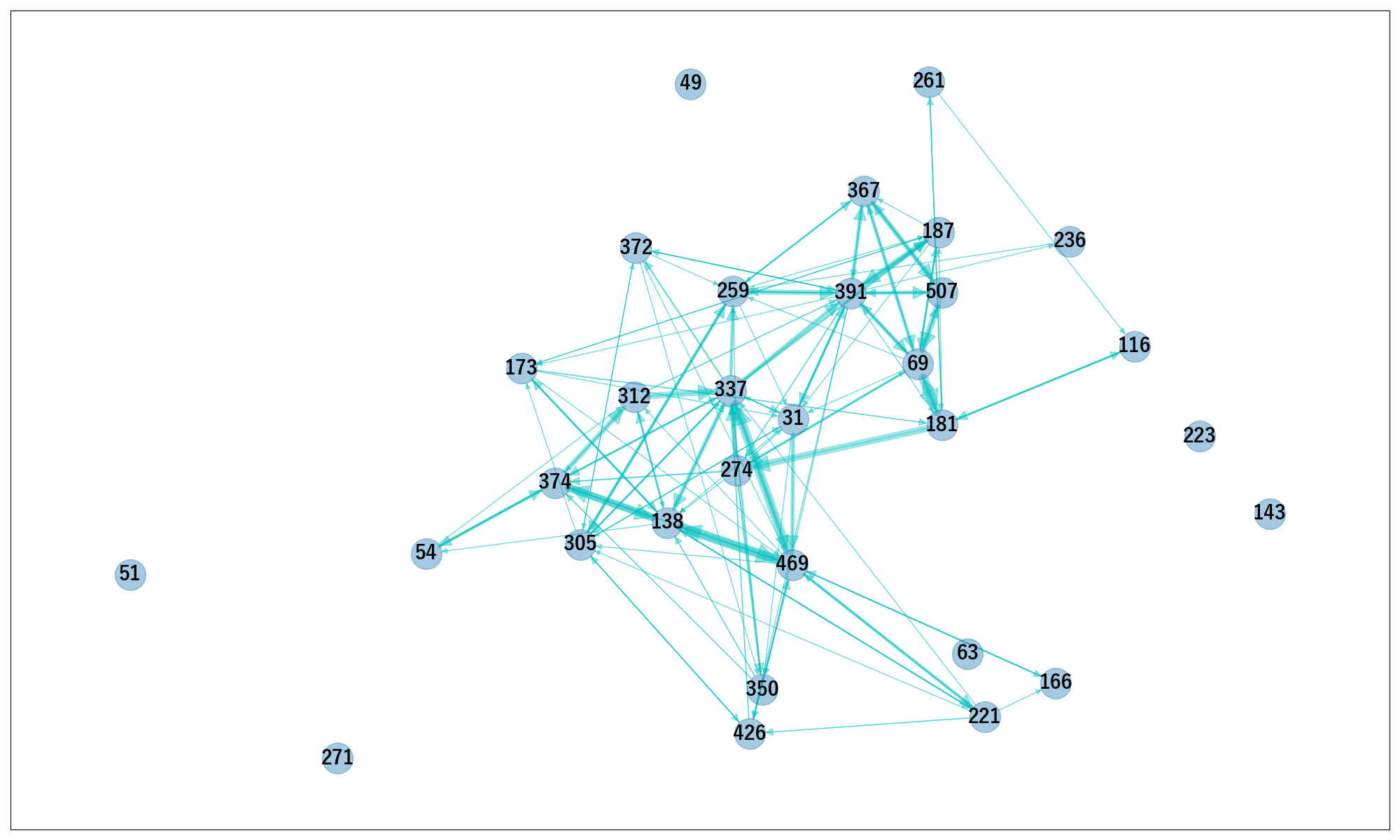} &
  \includegraphics[width=0.3\textwidth]{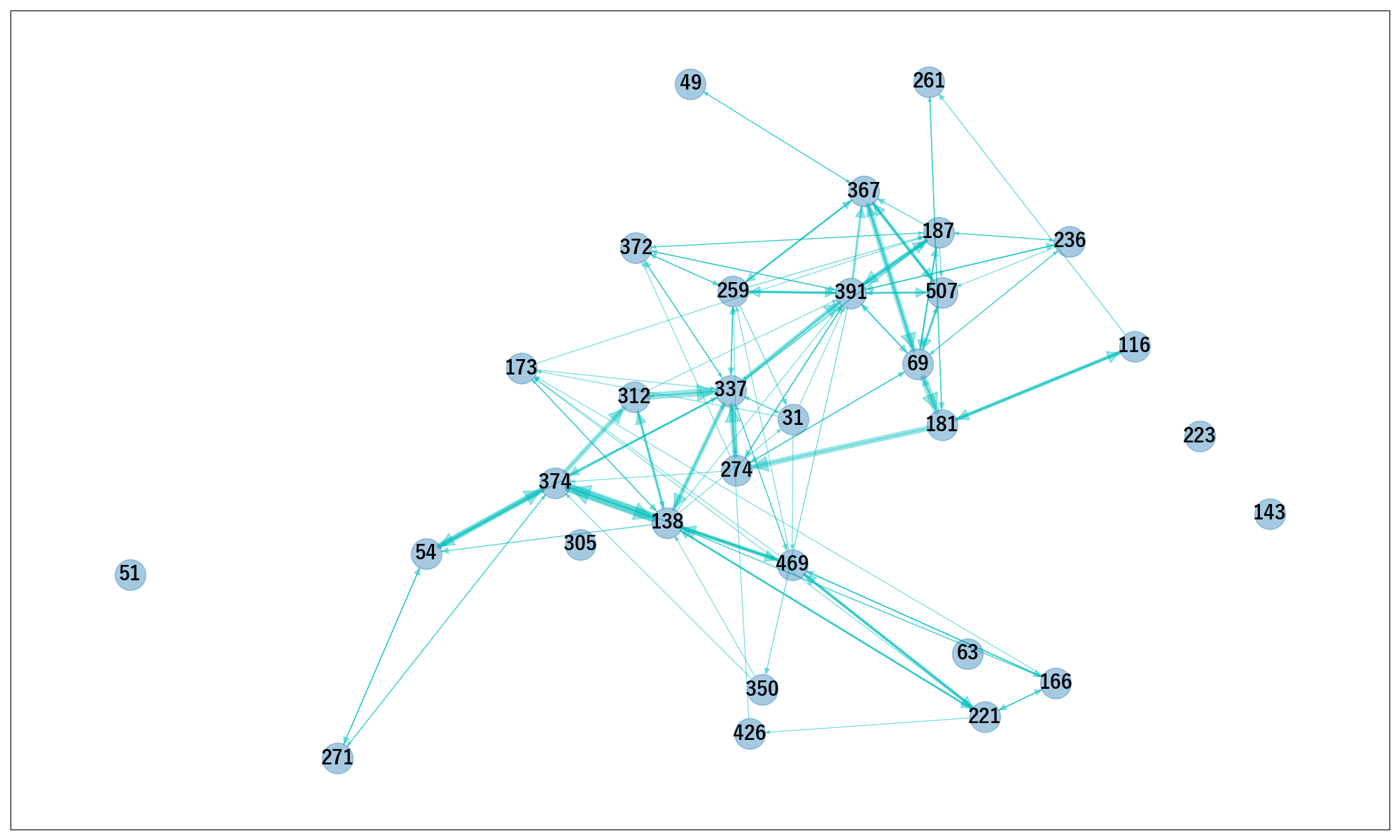} \\
   \small{(g) Subject G} &  \small{(h) Subject H} &  \small{(i) Subject I} 
  \end{tabular}
    \caption{Keyframe motion code graphs for all subjects. The kinematic data are used as input}
    \label{fig:transition_JIGSAWS_m2m}
\end{figure}

\begin{figure}[h]
    \centering
    \begin{tabular}{ccc}
    \includegraphics[width=0.3\textwidth]{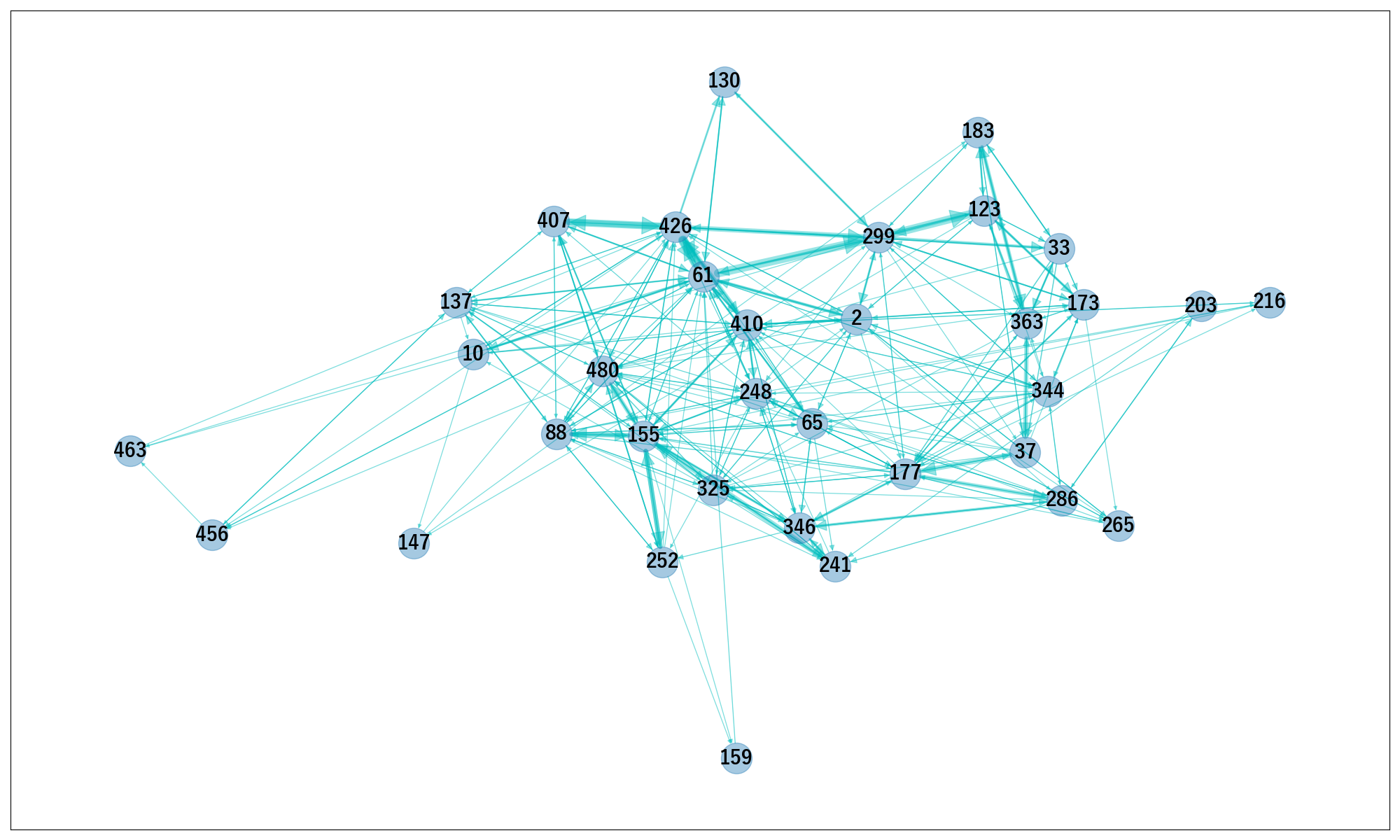} &
    \includegraphics[width=0.3\textwidth]{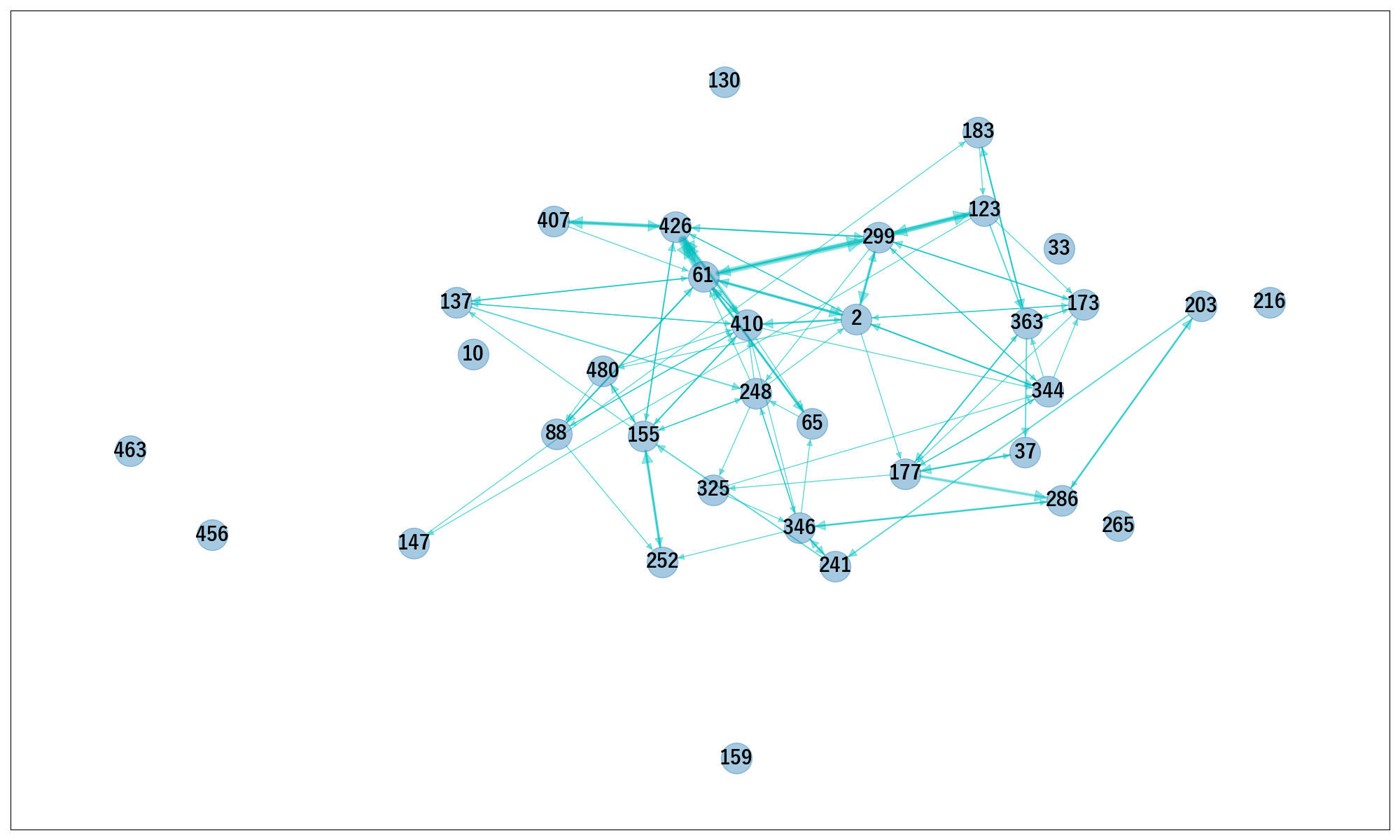} &
    \includegraphics[width=0.3\textwidth]{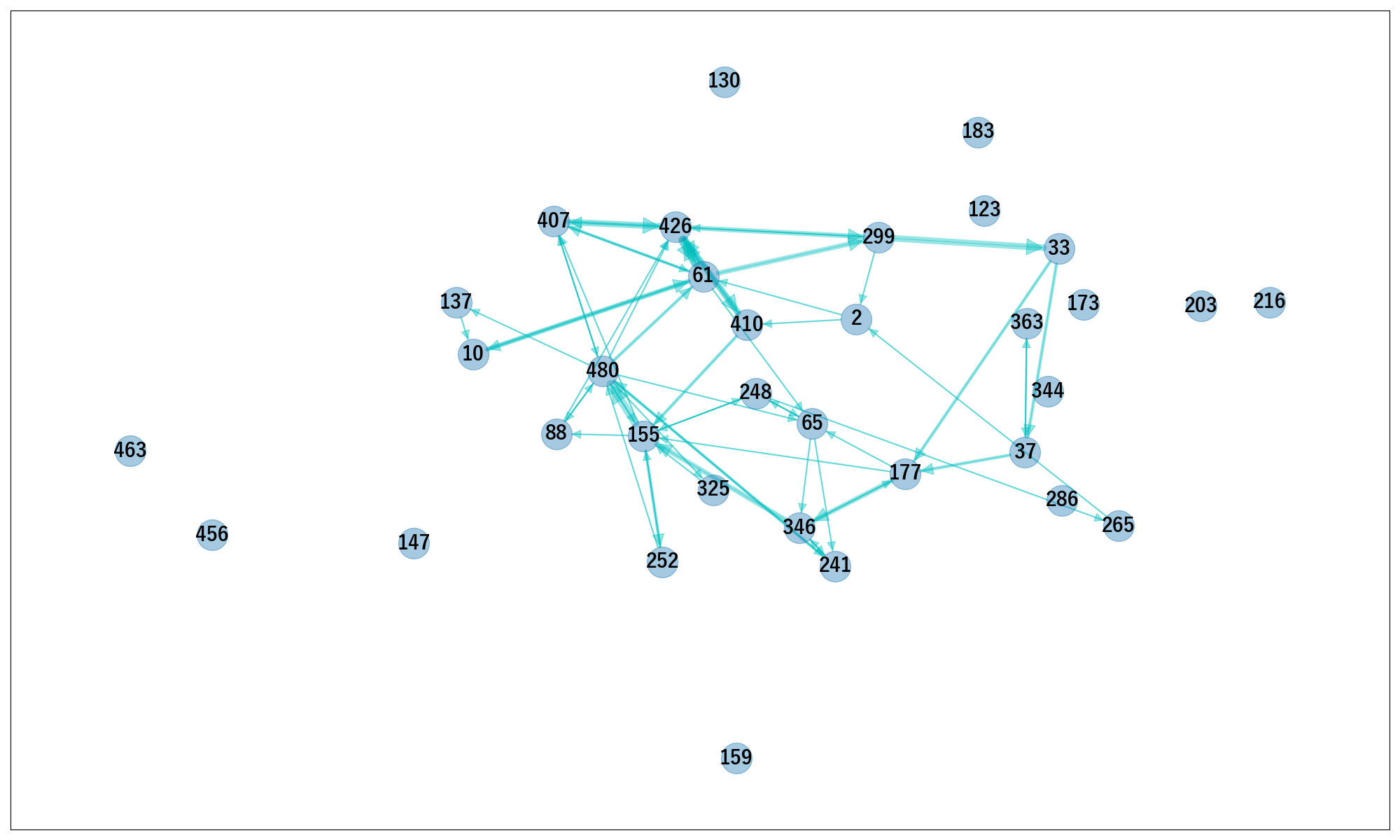} \\
     \small{(a) All subjects} &  \small{(b) Subject B} &  \small{(c) Subject C} \\
    \includegraphics[width=0.3\textwidth]{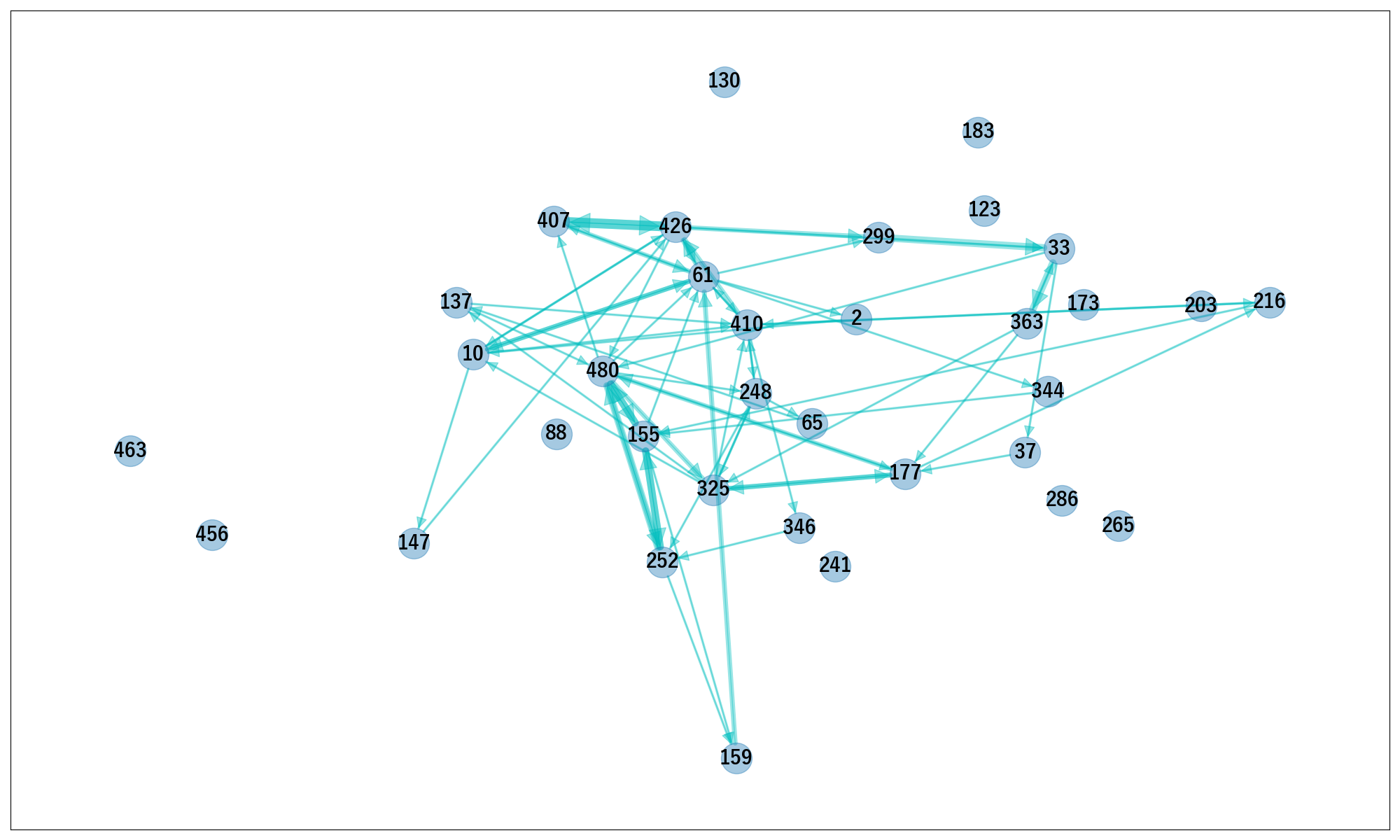} &
    \includegraphics[width=0.3\textwidth]{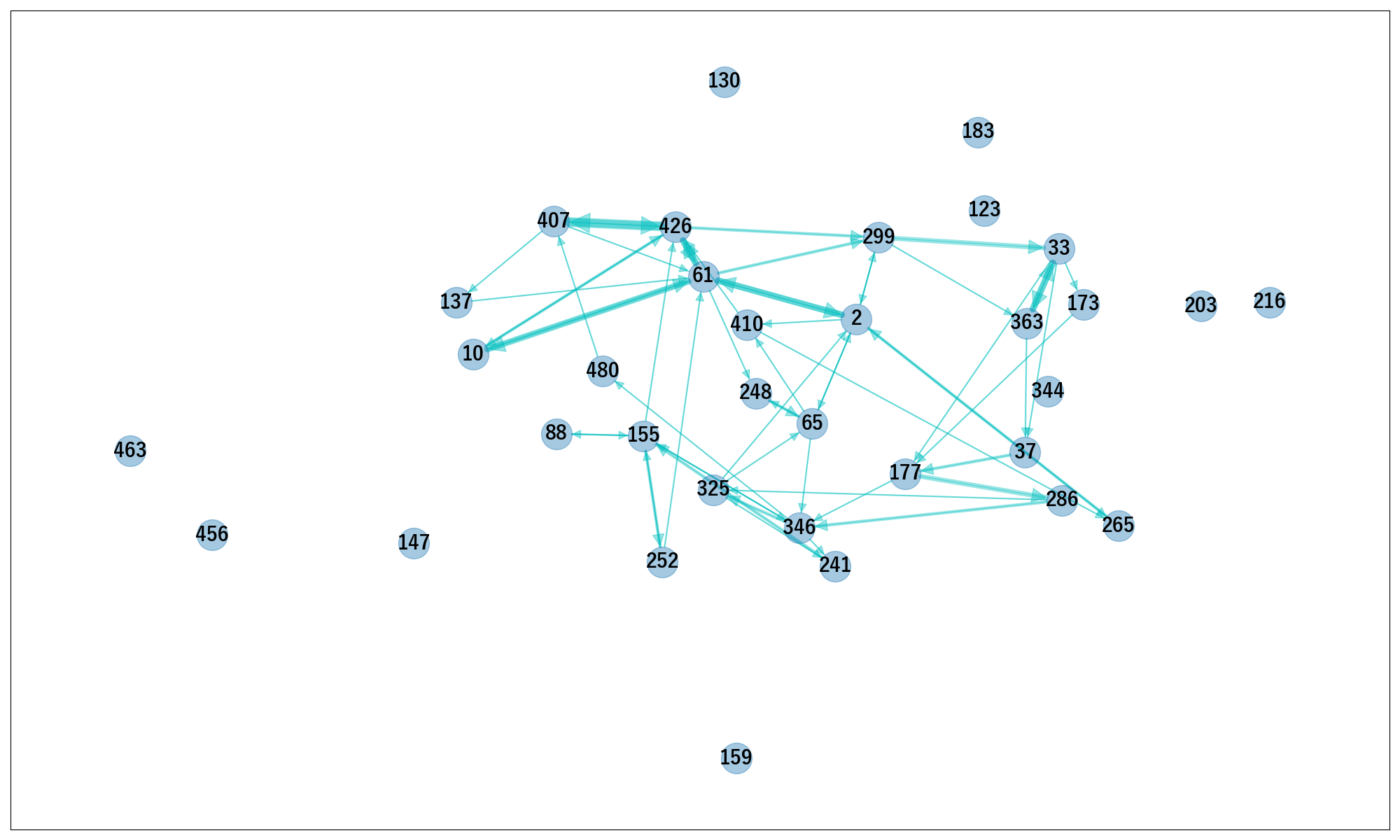} &
    \includegraphics[width=0.3\textwidth]{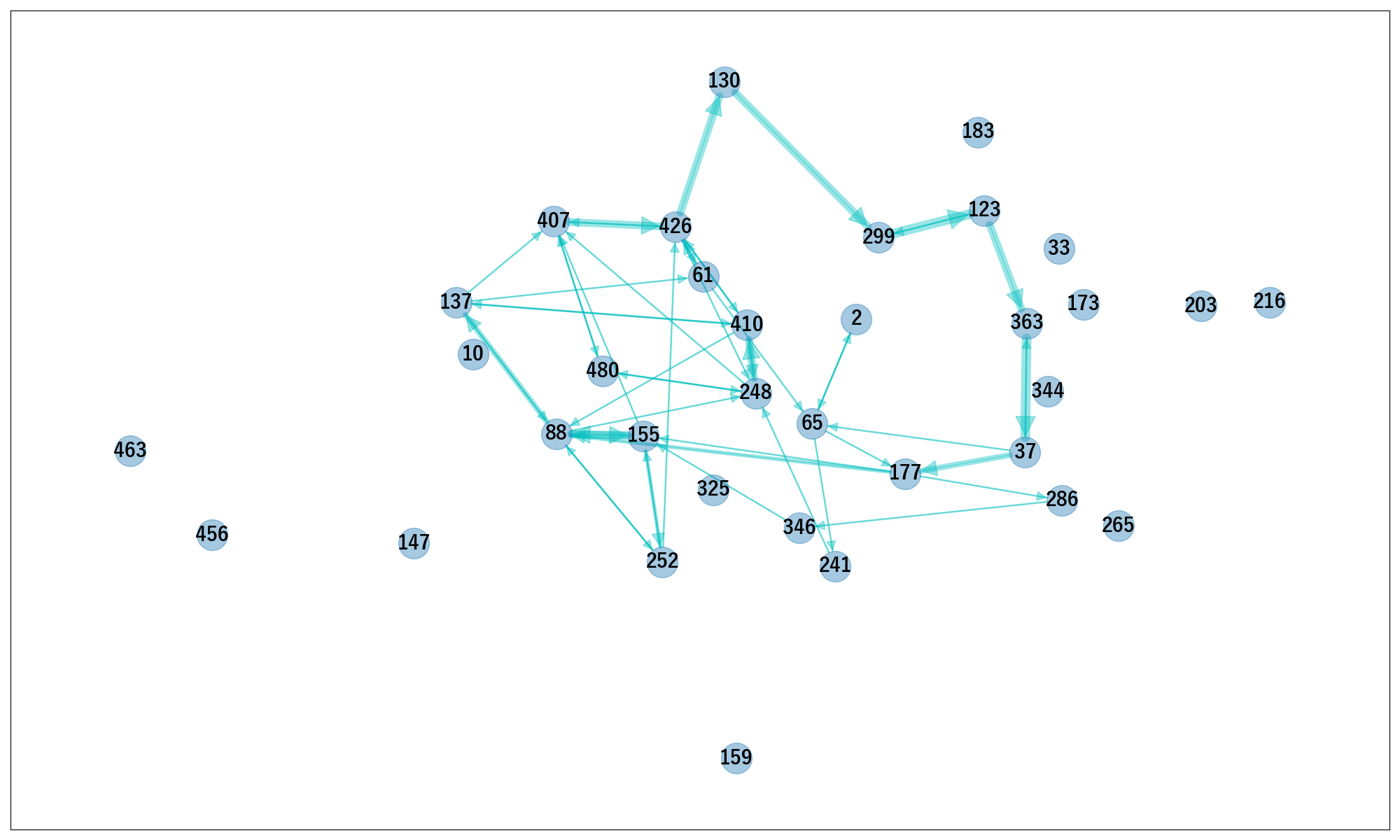} \\
     \small{(d) Subject D} &  \small{(e) Subject E} &  \small{(f) Subject F} \\
    \includegraphics[width=0.3\textwidth]{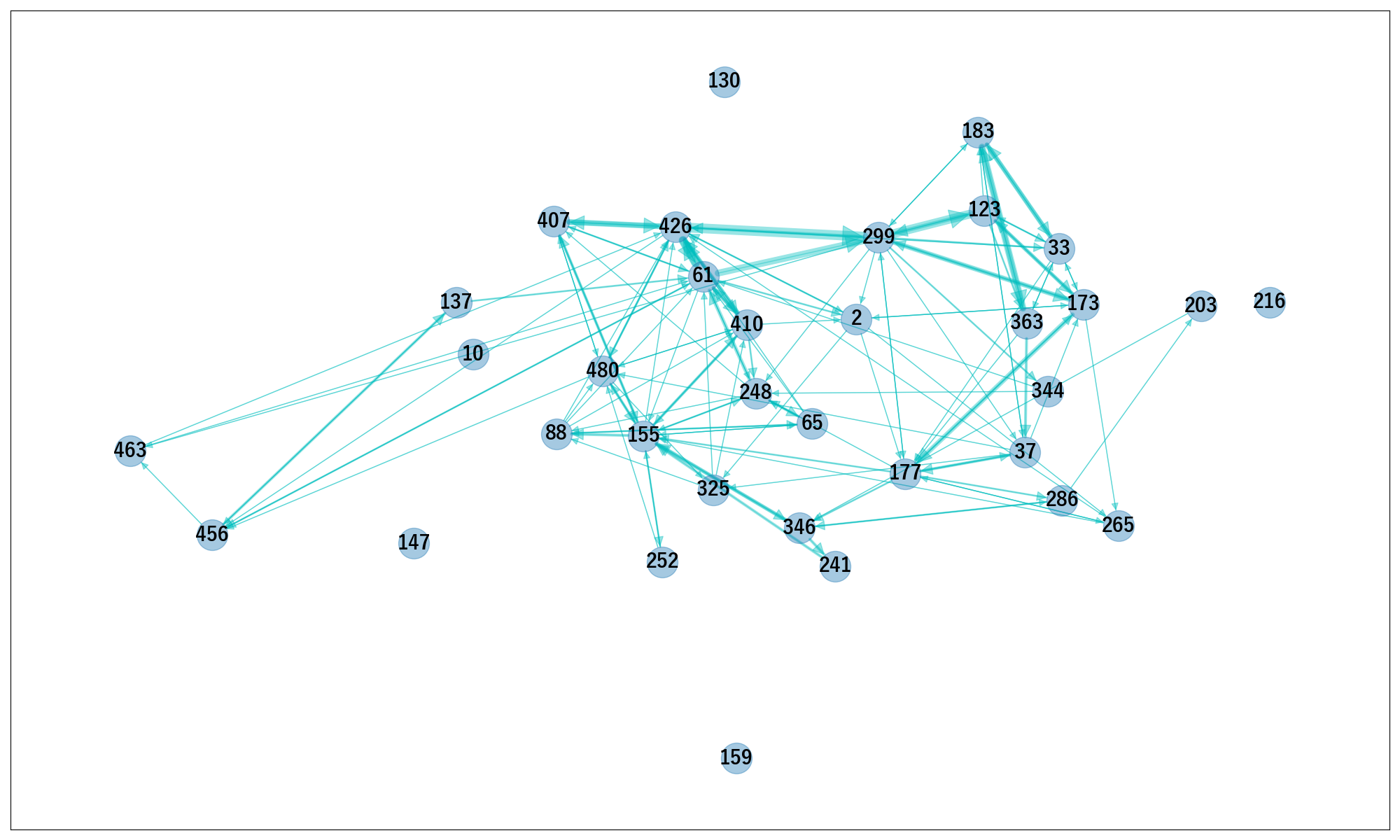} &
    \includegraphics[width=0.3\textwidth]{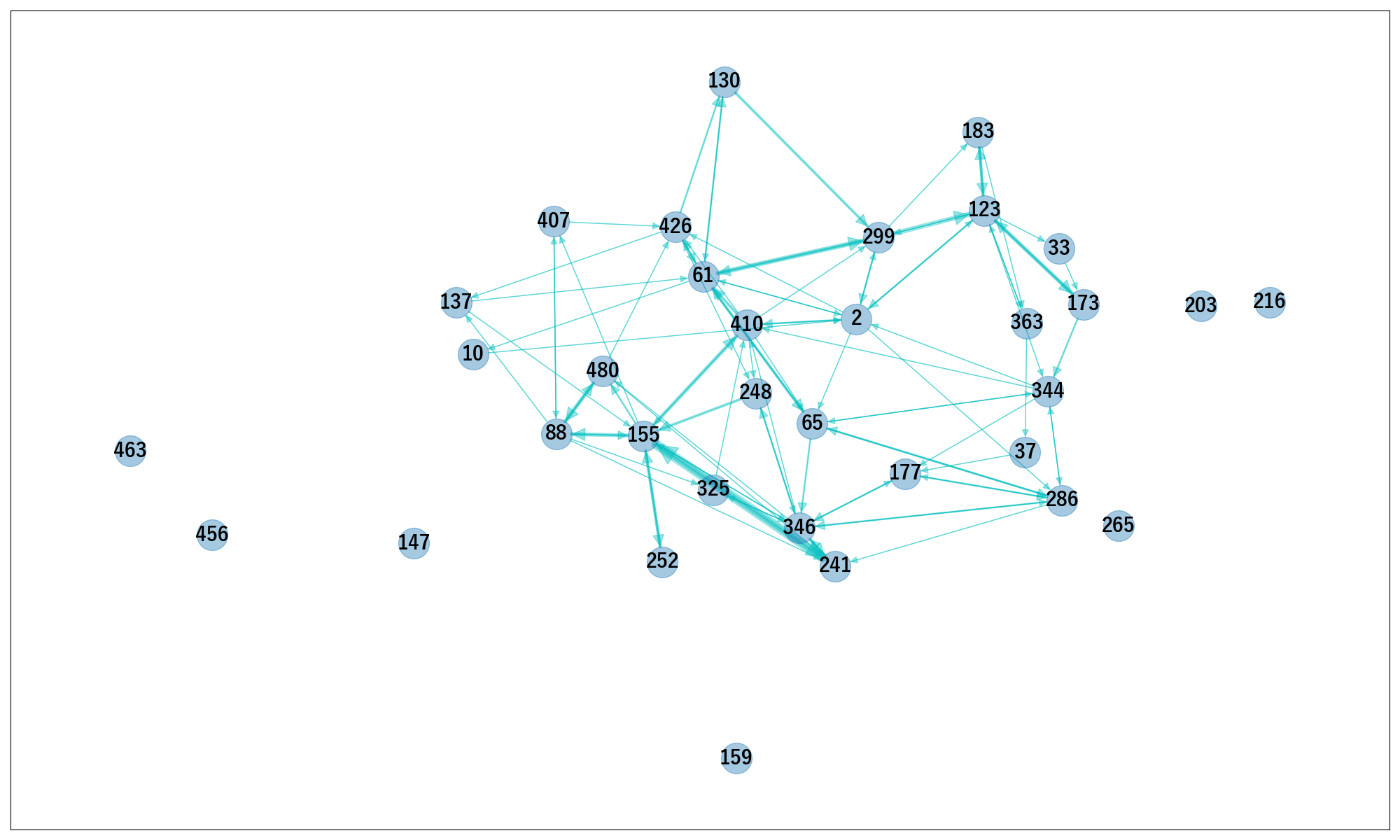} &
    \includegraphics[width=0.3\textwidth]{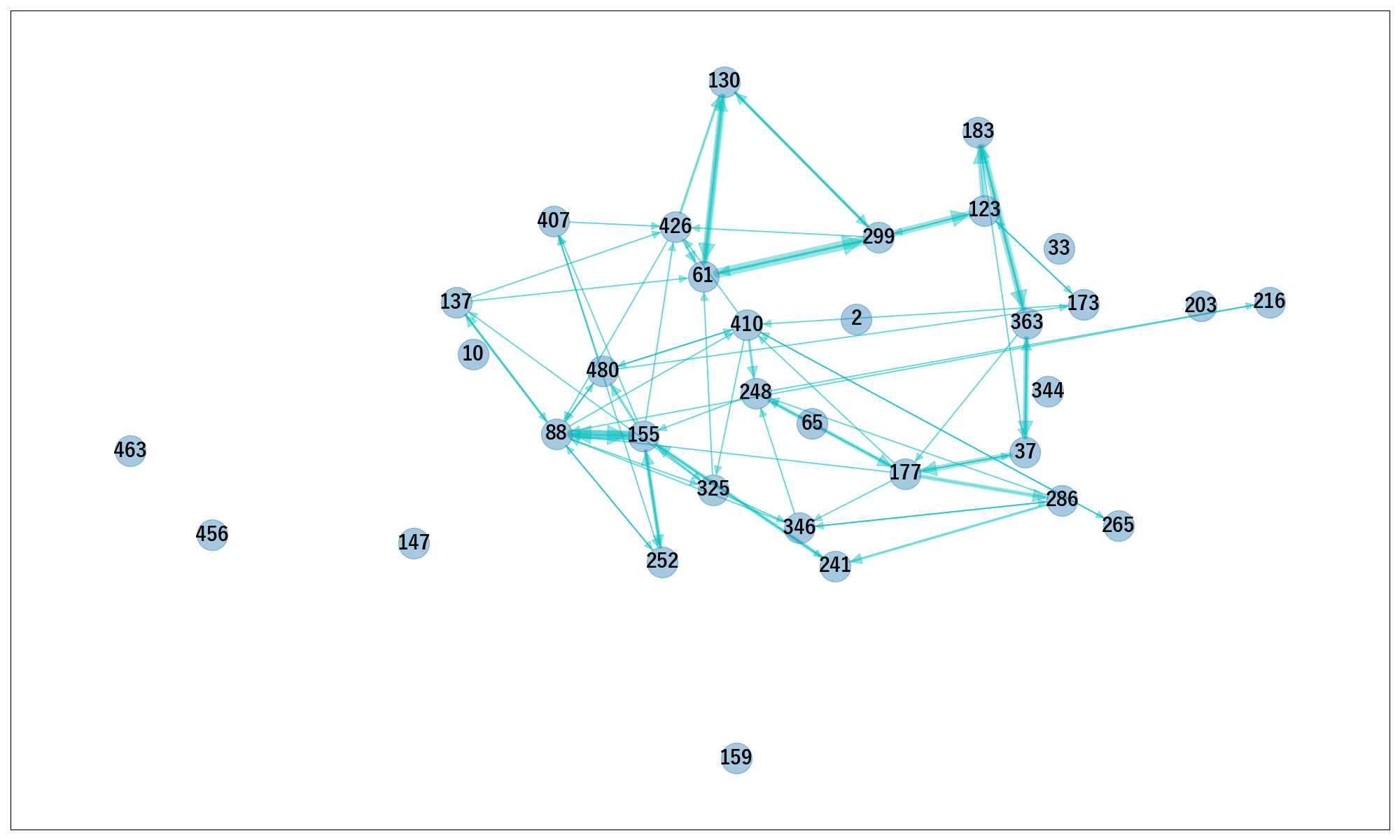} \\
     \small{(g) Subject G} &  \small{(h) Subject H} &  \small{(i) Subject I} 
    \end{tabular}
    \caption{Keyframe motion code graphs for all subjects. The video data are used as input}
    \label{fig:transition_JIGSAWS_i2m}
\end{figure}

\begin{figure}[h]
    \centering
    \begin{tabular}{ccc}
    \includegraphics[width=0.3\textwidth]{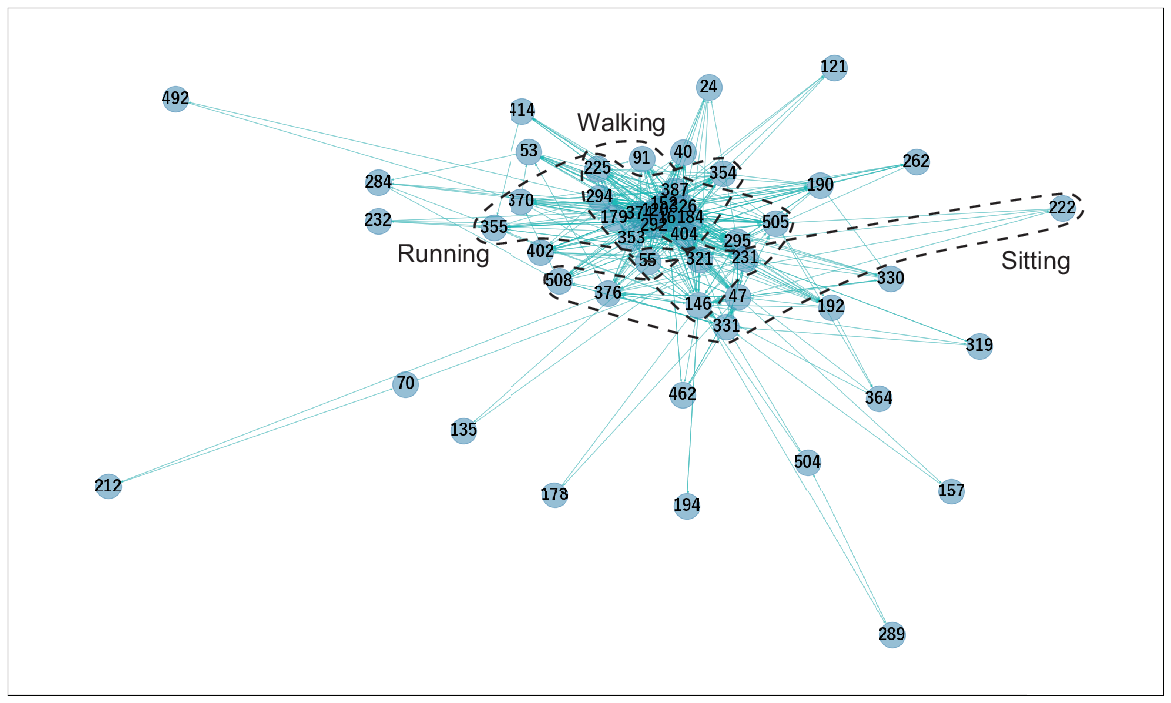} &
    \includegraphics[width=0.3\textwidth]{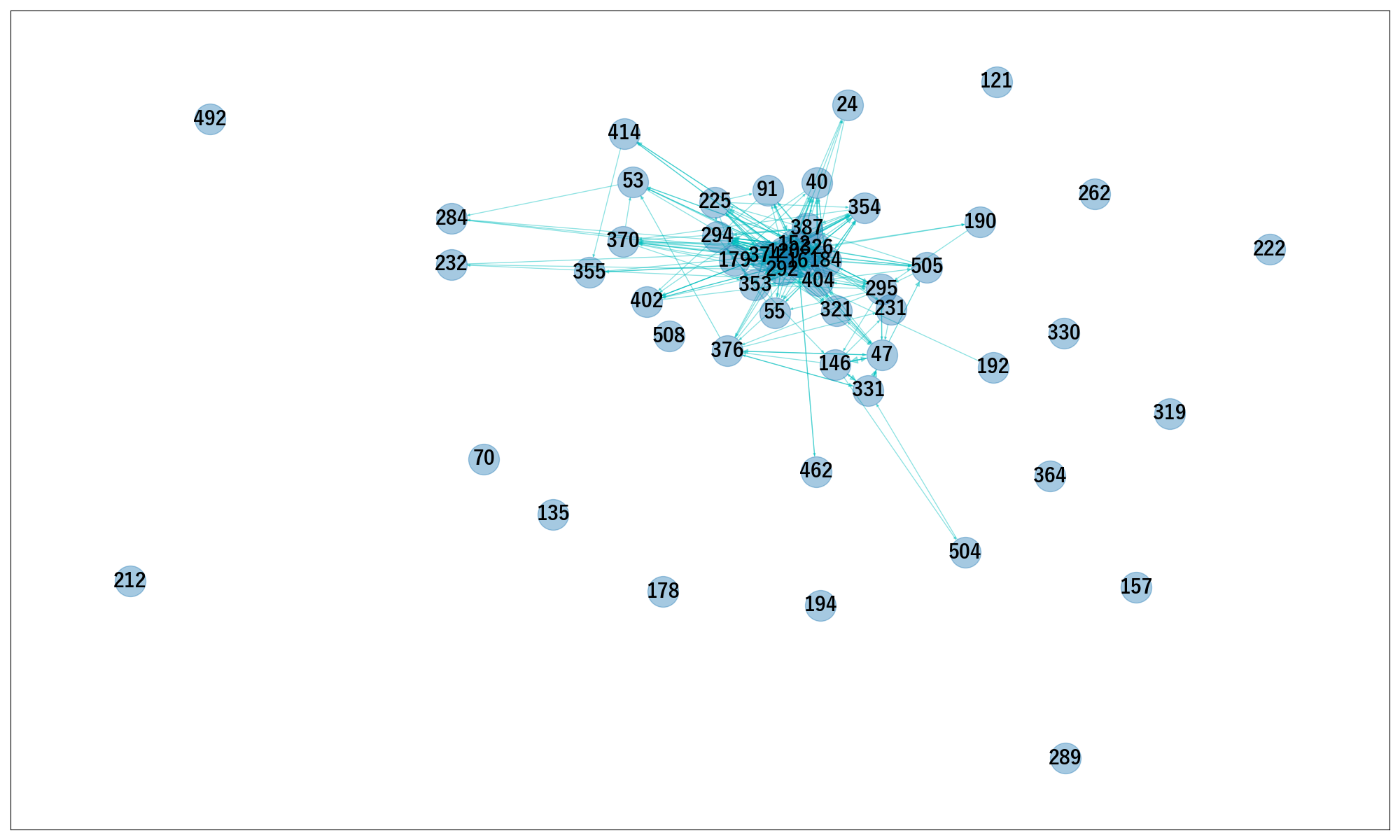} &
    \includegraphics[width=0.3\textwidth]{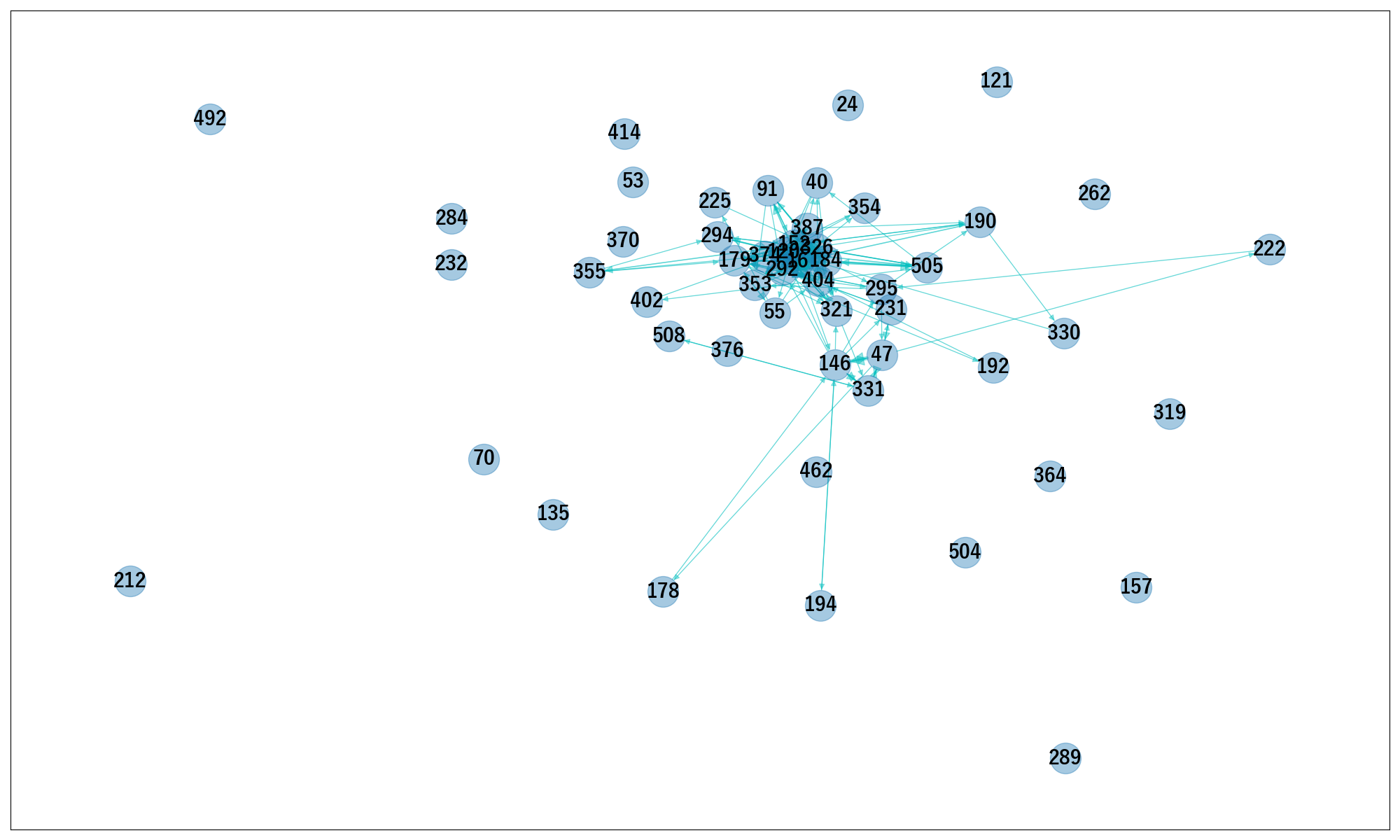} \\
     \small{(a) All subjects} &  \small{(b) Subject 1} &  \small{(c) Subject 2} \\
    \includegraphics[width=0.3\textwidth]{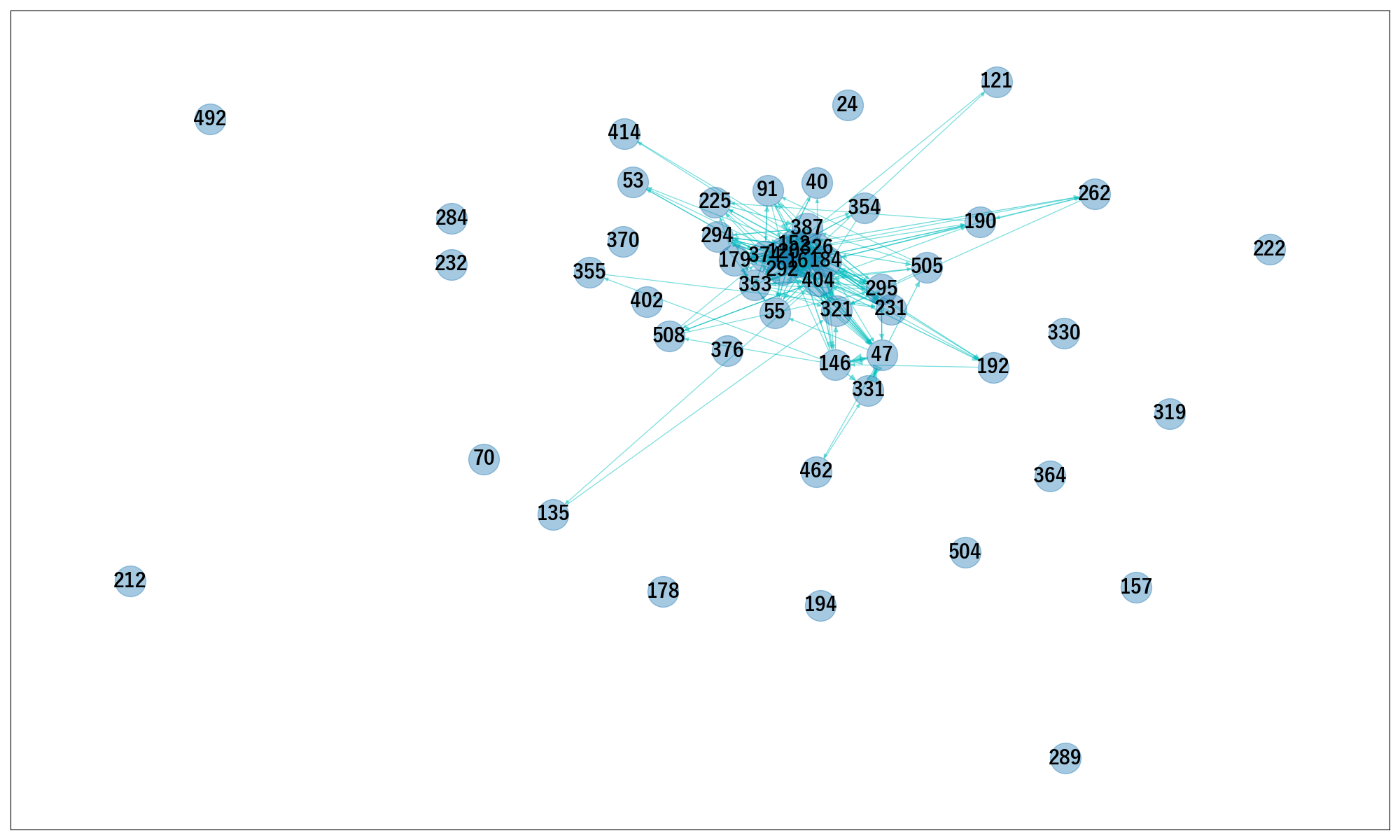} &
    \includegraphics[width=0.3\textwidth]{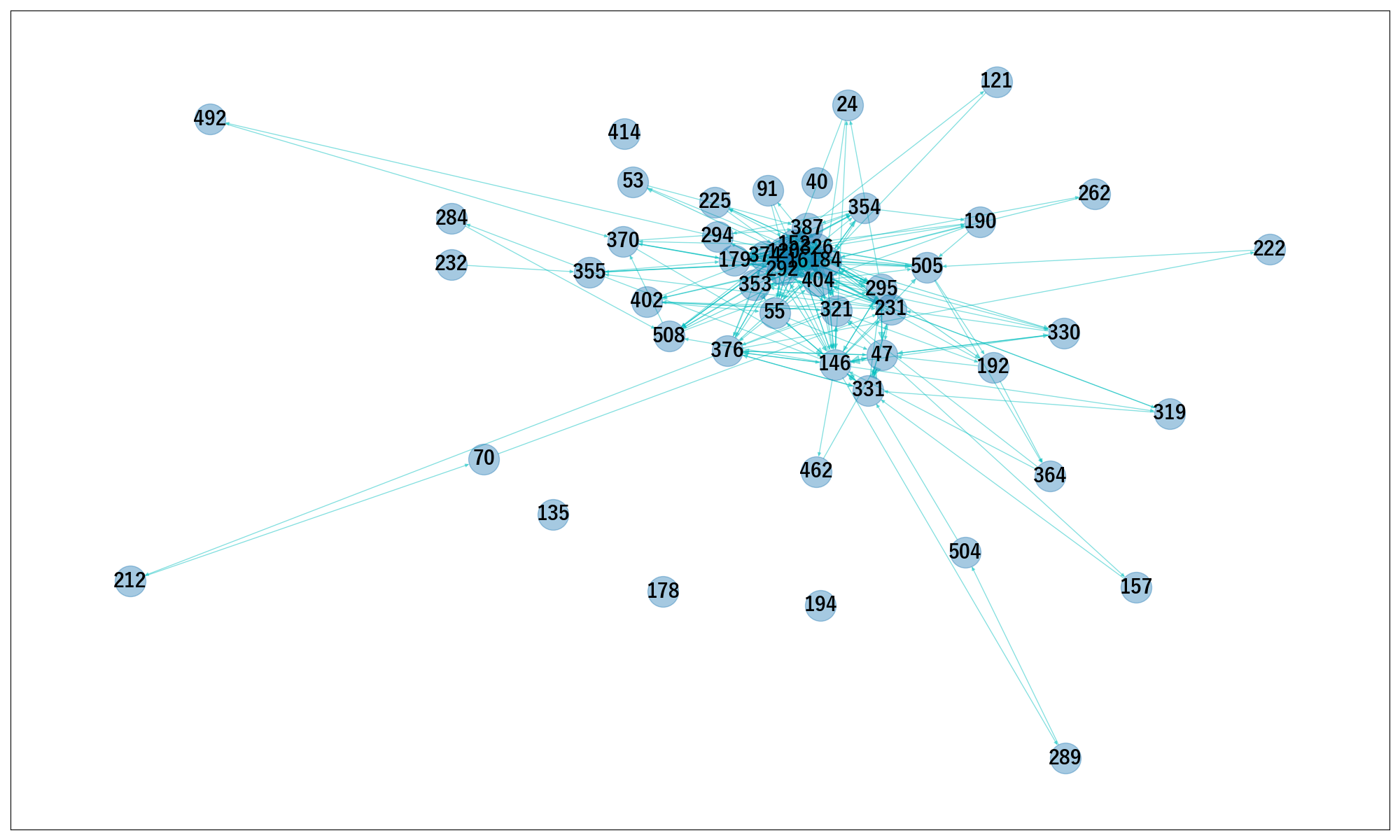} & \\
     \small{(d) Subject 3} &  \small{(e) Subject 4} & 
    \end{tabular}
    \caption{Keyframe motion code graphs for all subjects in HuGaDB. The
    dotted lines enclose the motion codes used for Walking, Running, and
    Sitting}
    \label{fig:transition_HuGaDB}
\end{figure}

In our experiments, we tested three types of input data. Figs. \ref{fig:transition_JIGSAWS_m2m}, \ref{fig:transition_JIGSAWS_i2m}, \ref{fig:transition_HuGaDB} show the keyframe motion code graphs of all individual
subjects for JHU-ISI gesture and skill assessment working set (JIGSAWS) kinematic inputs, JIGSAWS video inputs and 3D skeleton inputs (HuGaDB), respectively. As described in the main manuscript, there are differences in the codes
used between the subjects.

Online Resource 1 (ESM\_1.mp4) shows the synchronized visualization of the
input video and the keyframe motion codes for the two sequences of the JIGSAWS kinematic inputs,
\verb|Suturing_C001| of Subject C and \verb|Suturing_F001| of Subject F. One of
the clear differences is the IDs of motion codes used for the annotation,
``Pushing needle through tissue". The IDs, 49 and 236, are used by one of the two
subjects, but not by the other subject. The difference occurs repeatedly, and
the reason for the difference is considered to be caused by the difference of the
gripper pose. Since we do not have expert knowledge of robotic surgery, we
cannot tell whether it comes from the skill or habit of a subject, but the
similarities and uniqueness are visualized without the knowledge. 
Online Resource 2 (ESM\_2.mp4) shows a
comparison of changing the decoder attention width. If the attention width is
large ($M=100$), the motion is represented by a smaller number of keyframe
motion codes than in the case of small attention width ($M=10$). Since each
segment becomes long, the granularity of the motion codes is coarse. The control
of the granularity is one of our future areas of study.
Online Resource 3 (ESM\_3.mp4) shows the synchronized visualization of the
input video and the keyframe motion codes for the two sequences of the JIGSAWS video inputs,
\verb|Suturing_C001| of Subject C and \verb|Suturing_F001| of Subject F. The
difference between subjects C and F can be also observed, even if the input is
changed to videos. For example, the codes used in the annotation, ``Pushing
needle through tissue", are different.

\section{Action segmentation}\label{sec:Action segmentation}

\begin{figure}[h]
    \centering
    \includegraphics[width=1\textwidth]{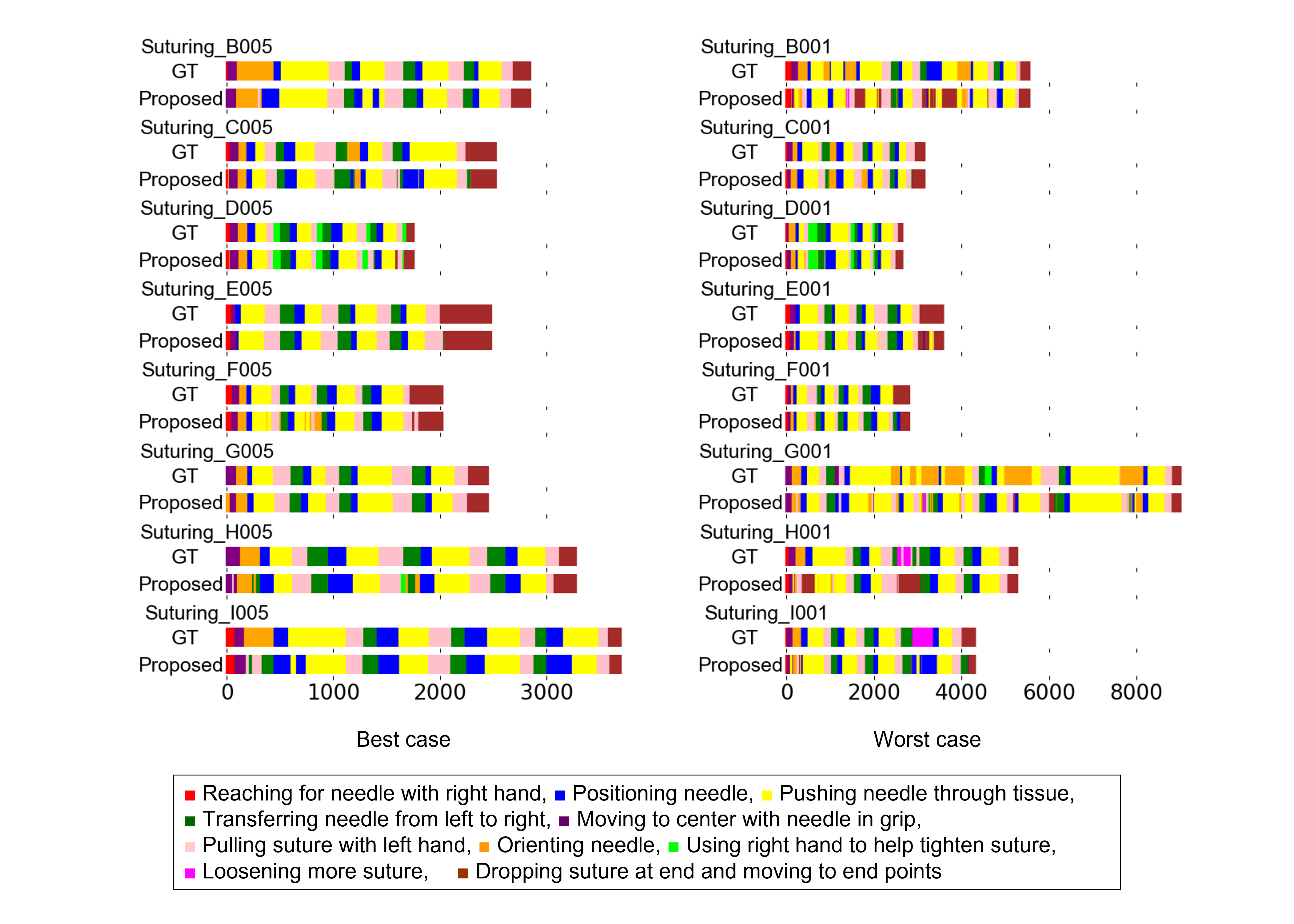}
    \caption{Action segmentation results of JIGSAWS kinematic input}
    \label{fig:action_seg_JIGm2m}
\end{figure}

Fig. \ref{fig:action_seg_JIGm2m} gives a visual overview of the action segmentation
results for JIGSAWS kinematic input. The left-hand column in the figure shows the
results of the test split with the highest accuracy in quantitative evaluation,
whereas the right-hand column shows the results of the test split of the worst case.
The action segmentation is accomplished with high quality for most of the
sequences. The sequences \verb|Suturing_H001| and \verb|Suturing_I001| are
examples of the worst case. The reason for their difference from the ground
truth is considered to be due to insufficient data because ``Loosening more suture"
appears in only one sequence in this training split (and appears in only three
sequences in the total dataset). In addition, since the subjects repeatedly try
and fail the operation in the sequences of \verb|Suturing_B001| and
\verb|Suturing_G001|, over-segmentation occurs in these cases and results in
low edit scores.

\begin{figure}[h]
    \centering
    \includegraphics[width=1\textwidth]{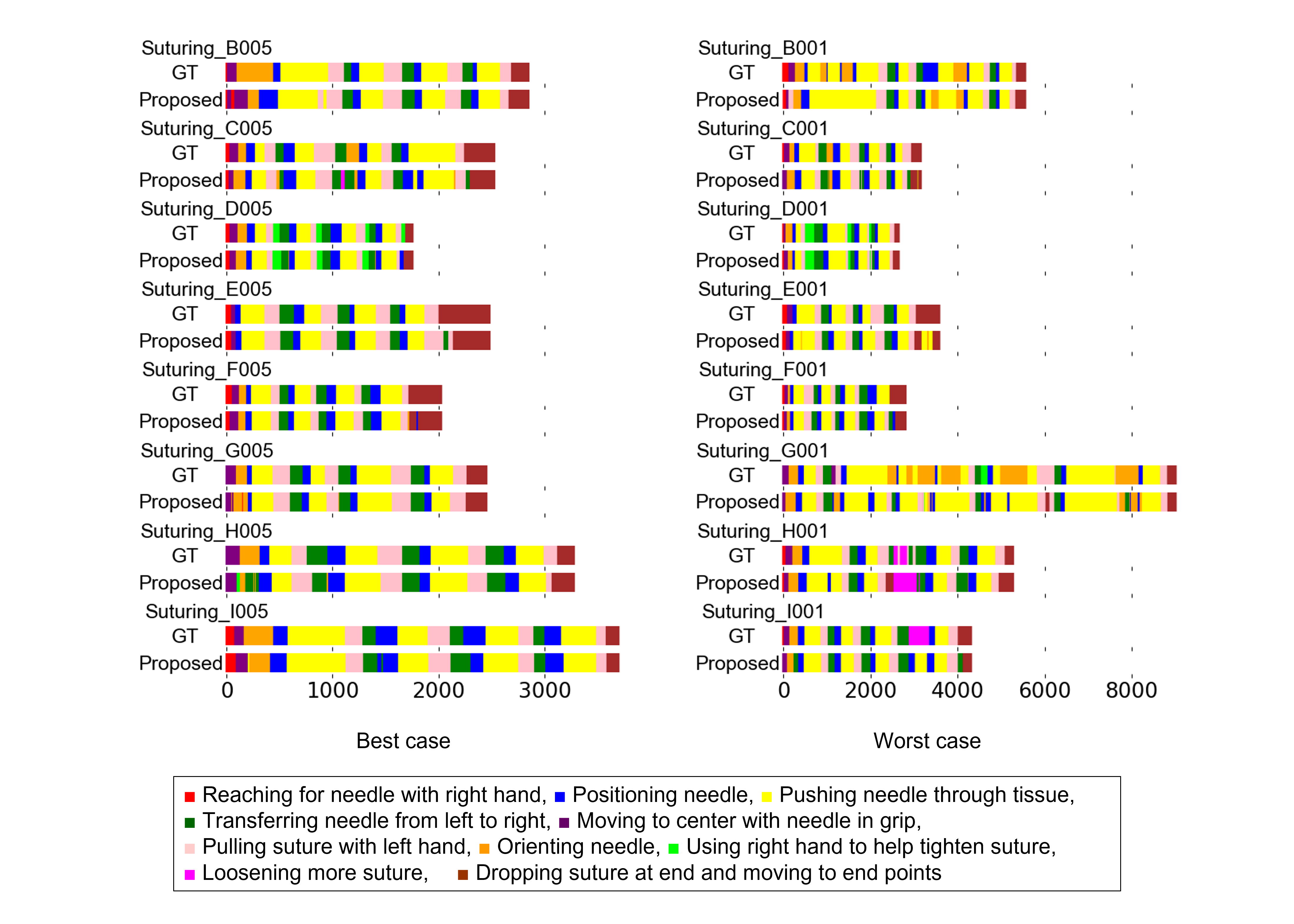}
    \caption{Action segmentation results of JIGSAWS video input}
    \label{fig:action_seg_JIGi2m}
\end{figure}

Fig. \ref{fig:action_seg_JIGi2m} gives a visual overview of the action segmentation
results for JIGSAWS video input. The left-hand column in the figure shows the results
for the test split with the highest accuracy in quantitative evaluation, whereas
the right-hand column shows the results for the test split of the worst case. The
segmentation quality of the video input is slightly more accurate than the
result for the kinematic input. The results of video input are believed to be
more accurate because the input and output are close to the condition of
annotation, which is given by watching videos. The difference from the ground
truth is considered to be due to the same reason in the case of kinematic input,
insufficient training examples and over-segmentation of repeated trials in an
operation.

\begin{figure}[h]
    \centering
    \includegraphics[width=1\textwidth]{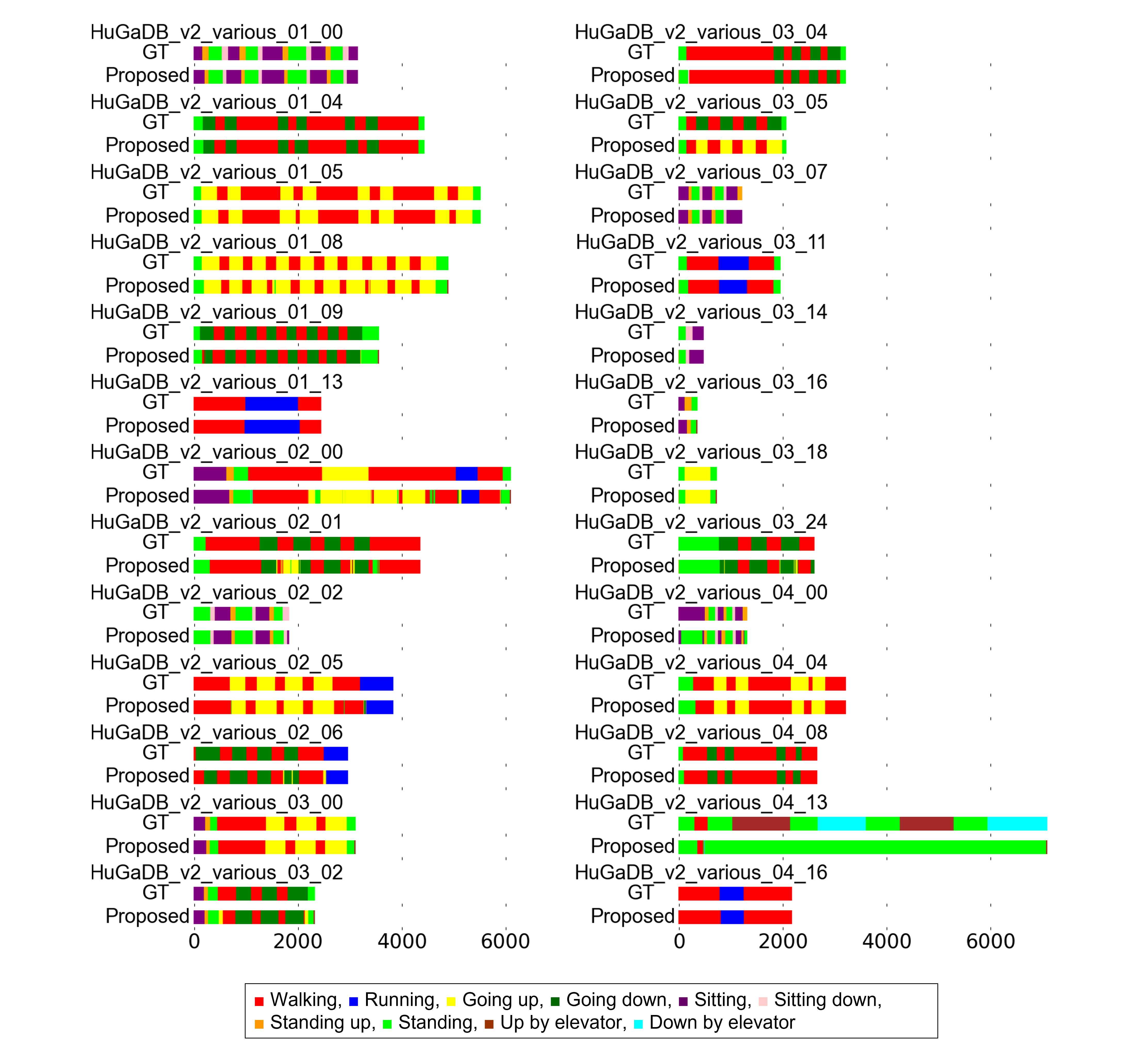}
    \caption{Action segmentation results of HuGaDB}
    \label{fig:action_seg_HuGaDB}
\end{figure}

Fig. \ref{fig:action_seg_HuGaDB} gives a visual overview of the action segmentation
results for HuGaDB. The sequences in the figure are from the test dataset,
excluding duplicate combinations of actions. The action segmentation is
accomplished with high accuracy for most of the sequences. Although the proposed
method is based on reconstructing acceleration and angular velocity, these factors do not
have significant differences between certain annotations, for example,
between ``Walking" and ``Going up/down" and between ``Standing" and ``Up/down by elevator",
which are considered to be the reason for the incorrect  classification found in
\verb|02_00|, \verb|03_05|, and \verb|04_13|.

\end{appendices}

\bibliography{article}

\end{document}